\documentclass[10pt,twocolumn,letterpaper]{article}

\usepackage{iccv}
\usepackage{times}
\usepackage{epsfig}
\usepackage{graphicx}
\usepackage{amsmath}
\usepackage{amssymb}
\usepackage{float} 
\usepackage{subfigure}
\usepackage[table]{ xcolor}

\usepackage[pagebackref=true,breaklinks=true,letterpaper=true,colorlinks,bookmarks=false]{hyperref}

\iccvfinalcopy 

\def\iccvPaperID{****} 

\ificcvfinal\fi

\begin{document}
\title{DDR-Net: Learning Multi-Stage Multi-View Stereo With Dynamic Depth Range}

\author{Puyuan Yi\thanks{equal contribution}\and Shengkun Tang\footnotemark[1]\and Jian Yao\thanks{Corresponding author} \\
School of Remote Sensing and Information Engineering, Wuhan University\\
{\tt\small yipuyuan@whu.edu.cn, shengkuntang@whu.edu.cn, jian.yao@whu.edu.cn}
}
\maketitle

\ificcvfinal\thispagestyle{empty}\fi



\begin{abstract}
    To obtain high-resolution depth maps, some previous learning-based multi-view stereo methods build a cost volume pyramid in a coarse-to-fine manner. These approaches leverage fixed depth range hypotheses to construct cascaded plane sweep volumes. However, it is inappropriate to set identical range hypotheses for each pixel since the uncertainties of previous per-pixel depth predictions are spatially varying. Distinct from these approaches, we propose a Dynamic Depth Range Network (DDR-Net) to determine the depth range hypotheses dynamically by applying a range estimation module (REM) to learn the uncertainties of range hypotheses in the former stages. Specifically, in our DDR-Net, we first build an initial depth map at the coarsest resolution of an image across the entire depth range. Then the range estimation module (REM) leverages the probability distribution information of the initial depth to estimate the depth range hypotheses dynamically for the following stages. Moreover, we develop a novel loss strategy, which utilizes learned dynamic depth ranges to generate refined depth maps, to keep the ground truth value of each pixel covered in the range hypotheses of the next stage. Extensive experimental results show that our method achieves superior performance over other state-of-the-art methods on the DTU benchmark and obtains comparable results on the Tanks and Temples benchmark. The code is available at https://github.com/Tangshengku/DDR-Net.
\end{abstract}

\begin{figure*}[t]
\begin{center}
\subfigure[CasMVSNet~\cite{cascade}]{
\begin{minipage}[t]{0.333\linewidth}
\centering
\includegraphics[height=1.8in,width=2.2in]{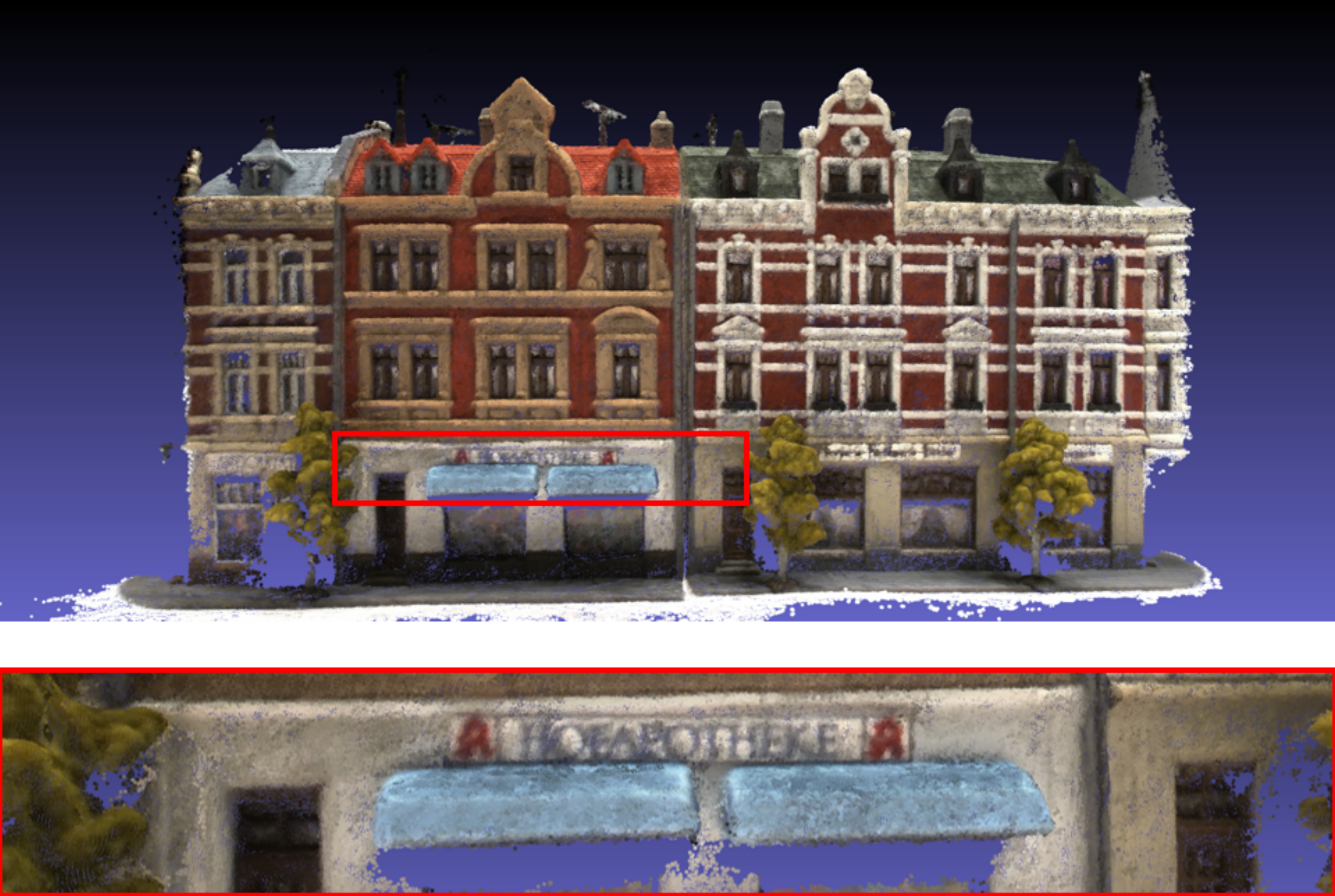}
\end{minipage}%
}%
\subfigure[UCSNet~\cite{ucs}]{
\begin{minipage}[t]{0.333\linewidth}
\centering
\includegraphics[height=1.8in,width=2.2in]{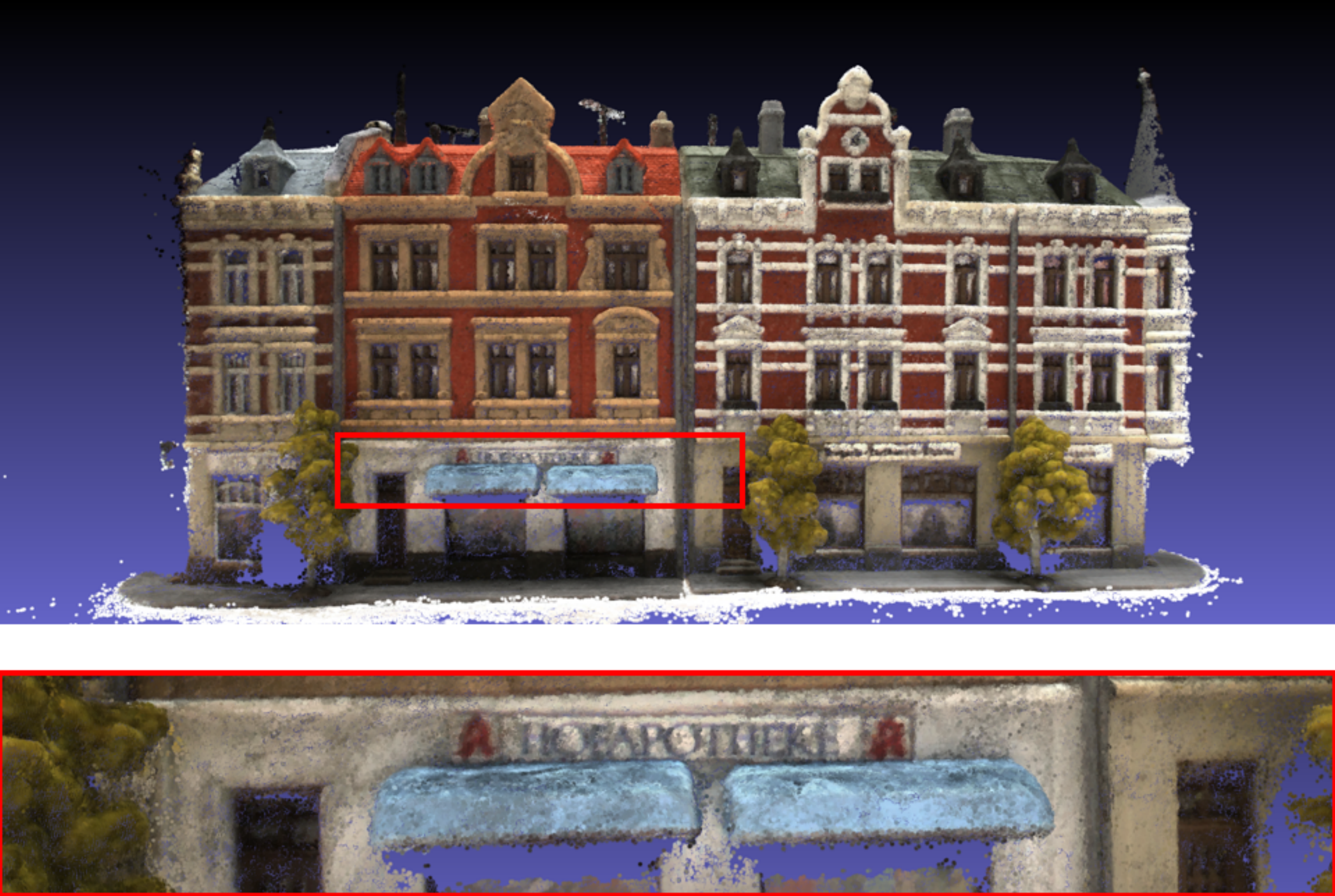}
\end{minipage}%
}%
\subfigure[DDR-Net (Ours)]{
\begin{minipage}[t]{0.333\linewidth}
\centering
\includegraphics[height=1.8in,width=2.2in]{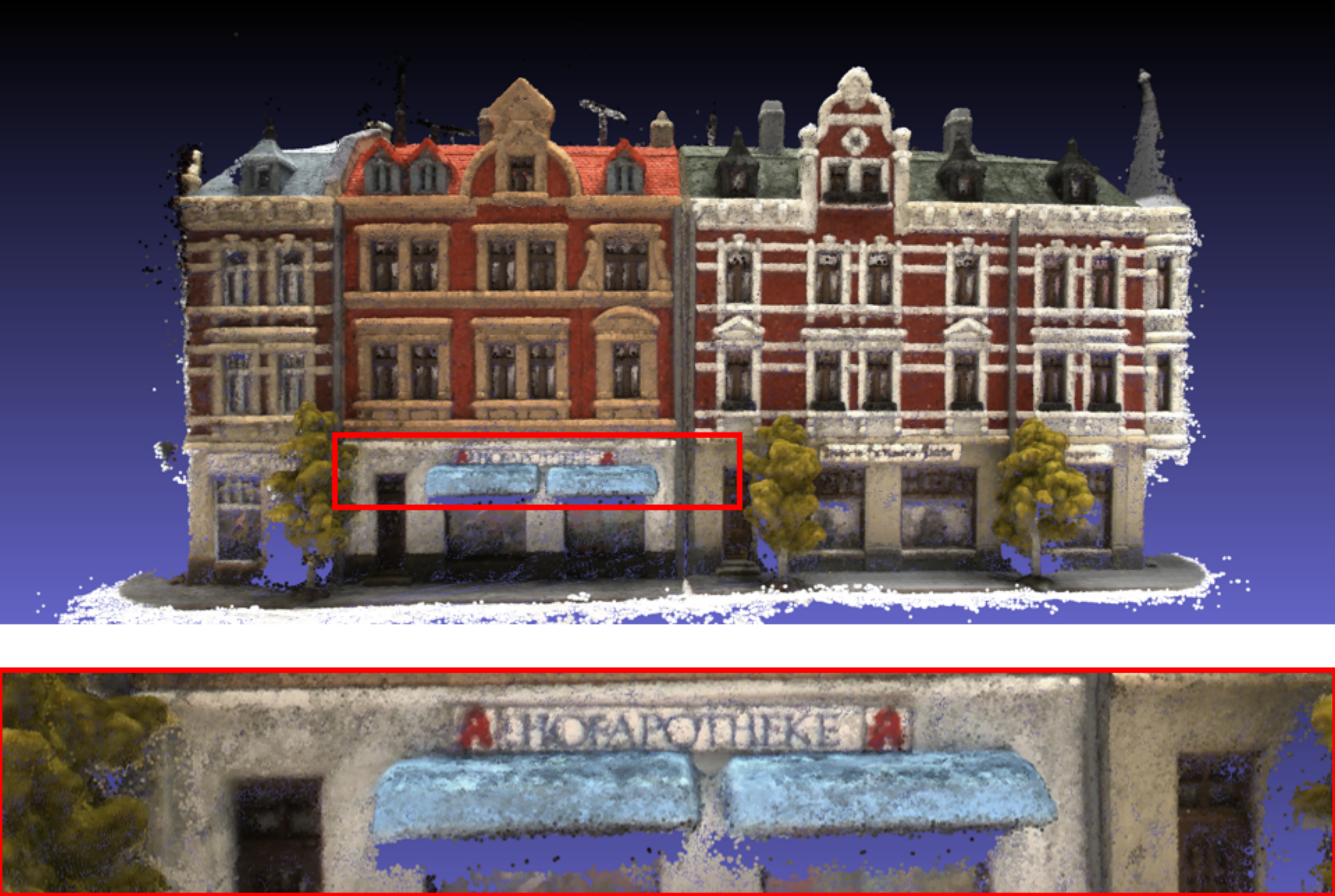}
\end{minipage}
}%
\end{center}
\vspace{-1.25em}
\caption{Point clouds reconstructed by state-of-the-art methods~\cite{ucs,cascade} and our proposed DDR-Net.}
\vspace{-1.25em}
\end{figure*}


\vspace{-1em}
\section{Introduction}
\label{Sec:Introduction}
\vspace{-0.5em}

Multi-view stereo (MVS) is a classic and fundamental problem in computer vision, which utilizes a set of images captured by calibrated cameras from multi viewpoints to reconstruct the 3D model of a certain scene. It has various applications in 3D visualization, autonomous driving and robotics~\cite{furukawa2015multi,seitz2006comparison}, etc.

Traditional methods~\cite{gipuma,surfacenet,pointcloudmethod2,patch1} formulate the MVS task as an optimization problem and utilize hand-crafted metrics to refine this task. Although traditional methods show great performance on the reconstruction of scenes with Lambertian surfaces, they still suffer from unreliable matching correspondences under the condition of illumination changes, low-texture regions and optical reflections. 

With the success of convolutional neural networks (CNNs), many learning-based methods have been developed to figure out these problems and achieve better results than traditional methods. Learning-based MVS methods~\cite{cascade,softargmin,pmvsnet,attention,cvp,mvsnet,recurrent} introduce deep CNNs to predict the depth map of each view followed by a multi-view post-processing for point cloud fusion. These methods infer the depth value of each pixel by leveraging probability distribution with a reasonable depth range. Specifically, Gu~\etal~\cite{cascade} built a cascaded cost volume to progressively subpartition the local space and refine the depth prediction with increasing resolution and accuracy. An essential step in CasMVSNet is to determine the depth range hypotheses for each stage. Generally, most methods in a coarse-to-fine manner~\cite{cascade,cvp} utilize static depth range hypotheses through the entire process which is inappropriate for depth estimation since static hypotheses are unable to adapt to the uncertainties of previous per-pixel depth predictions. The pixels with high uncertainties require wider range hypotheses to cover their ground truth depth values while the pixels with high confidences demand narrower depth ranges to achieve higher depth sampling rates. Some methods utilize numerical characteristics of probability distribution to estimate range hypotheses. Cheng~\etal~\cite{ucs} estimated depth ranges by variances and achieved better results. However, variances are lack of information from nearby pixels since their reception fields are confined. Moreover, using variances to compute uncertainties is not suitable for non-Gaussian probability distributions. To this end, we propose a Dynamic Depth Range Network (DDR-Net) to estimate depth range hypotheses dynamically, which achieves a significant improvement on accuracy and completeness for the final point cloud reconstruction.

In this work, we make efforts to estimate more accurate depth range hypotheses dynamically. We propose a novel cascaded structure with a refined loss strategy. Range estimation module (REM) is developed to estimate the uncertainty of a depth range from the previous probability distribution. Range estimation module (REM) is a shallow 2D CNNs structure. Since the network has a large reception field, REM is able to gather uncertainty information from nearby pixels to determine a more accurate depth range. Our proposed DDR-Net consists of three stages and the range estimation module (REM) is available in the 1st and 2nd stages because the depth range hypotheses in the 3rd stage are the final sampling and are directly utilized to produce the final high-resolution depth map.

To keep corresponding ground truth depth value of each pixel covered in the hypothesis range of the next stage, we propose a novel loss strategy that utilizes new sampled depth range hypotheses and a probability distribution of the former stage to produce a refined depth map. We clamp depth range hypotheses in the former stage by a new sampled depth range. And the corresponding probability distribution will also be clamped to obtain a refined probability distribution. After normalizing the probability distribution, the refined depth map can be regressed in the $soft$ $argmin$ manner. Through leveraging the refined depth map, the range estimation module (REM) is able to learn better depth range hypotheses of higher confidences.

Our main contributions are summarized as follows:\\[-2em]
\begin{itemize}
\item We propose a novel multi-stage MVS framework (DDR-Net) with range estimation module (REM) to estimate more accurate depth ranges, which dynamically determines the depth range of every pixel in depth maps and gains more accurate depth values.\\[-2em]
\item We design a novel loss strategy to refine our range estimation module (REM), which further refines our output depth range estimation.\\[-2em]
\item Extensive experimental results show that our method achieves state-of-the-art (SOTA) performance on the DTU benchmark and comparable reconstruction results with other state-of-the-art methods on the Tanks and Temples benchmark.
\end{itemize}

\begin{figure*}[t]
\begin{center}
\includegraphics[width=1.05\textwidth,height=0.4\textwidth]{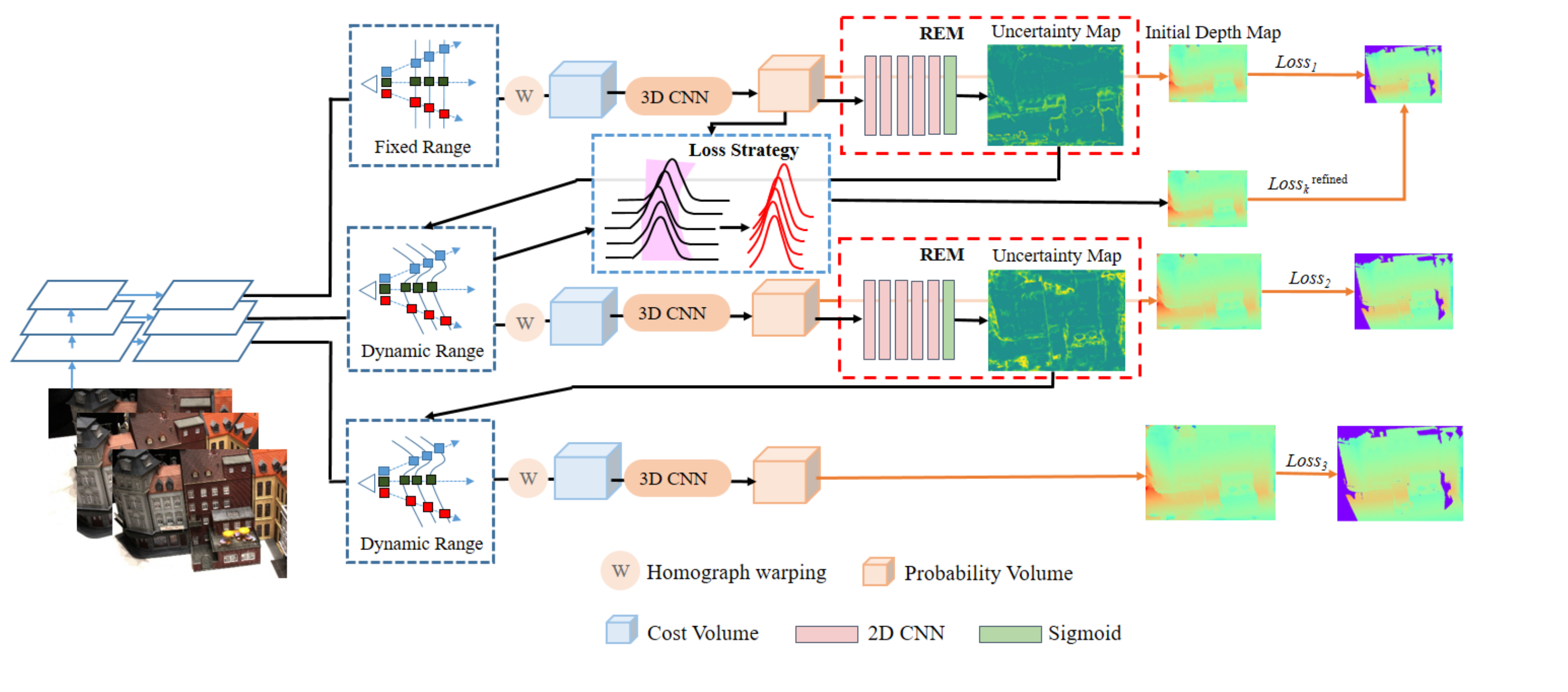}
\end{center}
\vspace{-2.0em}
\caption{The full network architecture of our proposed DDR-Net.}
\label{Fig:Architecture}
\vspace{-1.25em}
\end{figure*}

\vspace{-1.0em}
\section{Related Work}
\label{Sec:RelatedWork}
\vspace{-0.5em}

\noindent {\bf Traditional methods.} Multi-view stereo has been well studied for decades and a lot of great methods had been proposed. Traditional MVS methods can be generally categorized into four types: volumetric methods, patch-based methods, point-cloud based methods and depth map reconstruction methods. Volumetric methods~\cite{surfacenet,kar2017learning,richter2018matryoshka,wu2018learning,zhang2018learning} construct a bounding volume containing the scene and divide the volume into small voxel grids. Then these methods choose the voxels belonging to surface of the scene and filter the unattached voxels. Patch-based methods~\cite{henderson2019learning,kato2018neural,patch1,wang2018pixel2mesh} consider the surface as a collection of patches. These approaches first reconstruct the surface in some areas and then propagate to some low-texture areas. Point-cloud based methods~\cite{mvspointcloudmethod1,pointcloudmethod2,lin2018learning,wang2019mvpnet} directly utilize the 3D points to produce results. Depth-map based methods~\cite{depth1,gipuma,depth2,depth3} aim to estimate the depth of the scene and reconstruct point clouds through various fusion methods. Among these methods, depth-map based MVS methods have shown more flexibility with low time and GPU memory consuming which became the major methods utilized in learning-based frameworks. Although traditional methods achieve remarkable results, they still suffer from reconstructing illumination-changes, low-texture and intense-reflection regions.

\noindent {\bf Learning-based methods.} Yao~\etal~\cite{mvsnet} first introduced a differentiable homography warping to build a variance-based cost volume followed by multi-scale 3D CNNs for regularization. It achieved incredible results compared with conventional methods. However, 3D CNNs are generally time and GPU memory consuming. To allow handling high-resolution images, Yao~\etal~\cite{recurrent} utilized a gated recurrent unit (GRU)~\cite{GRU} to replace 3D CNNs in MVSNet to address the memory-intensive issue. Luo~\etal~\cite{attention} utilized an attention mechanism to construct and regularize cost volumes. Both perceptual information and contextual information of the local scene are fused to improve the matching robustness. Zhang~\etal~\cite{visibility} proposed a formulation to calculate the uncertainties of a cost volume and fuse cost volumes according to these uncertainties by integrating occluded information in the MVS network. Xu~\etal~\cite{xu2020pvsnet} learned to predict the visibility of every image and present an anti-noise strategy to refine the visibility network. Recently, a novel cascade formulation of 3D cost volumes was presented in~\cite{ucs,cascade,cvp} to reduce the memory consumption. The fine-grained depth map are refined in a coarse-to-fine manner. Our methods follow the same coarse-to-fine strategy as these approaches and propose a novel range estimation module (REM) to estimate a more accurate depth range.

\noindent {\bf Depth range hypotheses.} Multi-stage multi-view stereo frameworks have been proposed in~\cite{ucs,cascade,cvp}, which build a cascade cost volume to estimate the depth map in a coarse-to-fine manner. In these methods, the  depth range hypotheses of the first stage covers the entire depth range of the input scene. In the following stages, the hypotheses range is narrowed based on the predicted depth map from the previous stage. CasMVSNet~\cite{cascade} set a reducing factor of the depth range hypotheses which narrows the depth range manually and depth ranges of all pixels are identical. CVP-MVSNet~\cite{cvp} found 3D points that are two pixels away from the its projection along the epipolar line in both directions on the source view as the depth residual range for depth refinement. UCSNet~\cite{ucs} leveraged the variance of the depth distribution to determine the depth range hypotheses in the next stage. Our method is particularly inspired by this line of work where we propose a range estimation module (REM) to measure the depth range hypotheses from a probability distribution combined with a novel designed loss strategy. 

\vspace{-1em}
\section{Our Method}
\label{Sec:OurMethod}
\vspace{-0.5em}

In this paper, we propose an effective and efficient multi-view stereo framework (DDR-Net) that utilizes a novel range estimation module (REM) to estimate the depth range hypotheses in a coarse-to-fine strategy. Given a reference image $\mathbf{I}_1$ and $N-1$ source images $\left\{ \mathbf{I}_i \right\}_{i=2}^N$, our proposed DDR-Net progressively regresses a fine-grained depth map using a learned dynamic depth range. The overall of our proposed network is illustrated in Figure~\ref{Fig:Architecture}. We first leverage a feature pyramid network (FPN)~\cite{fpn} to extract multi-scale image features (see Section~\ref{Sec:MultiScaleFeatureExtraction}) under different image resolutions. Then we construct multi-scale cost volumes using image features through a homography warping process and apply 3D CNNs to regularize cost volumes to form a probability volume (see Section~\ref{Sec:CostVolumeConstruction}). A novel range estimation module (REM) is designed to learn the dynamic depth range by utilizing probability distributions (see Section~\ref{Sec:DynamicDepthRangeEstimation}). Finally, we apply supervision to depth output of all stages and design a loss strategy to constrain our novel REM (see Section~\ref{Sec:LossStrategy}).

\begin{figure*}
\centering
\begin{minipage}[b]{0.2\linewidth}
\centering
\subfigure[Original image]{
\includegraphics[width=1in,height=1in]{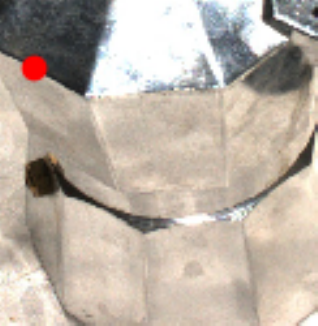}
}%

\subfigure[Ground truth]{\label{Genelecs:Genelec 8030AP}\includegraphics[width=1in,height=1in]{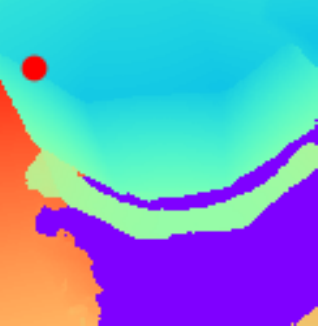}}
\end{minipage} 
\hspace{-0.6cm}
\begin{minipage}[b]{0.2\linewidth}
    \centering
    \subfigure[UCSNet~\cite{ucs}]{\label{Genelecs:Genelec 8020 AP}\includegraphics[width=1in,height=1in]{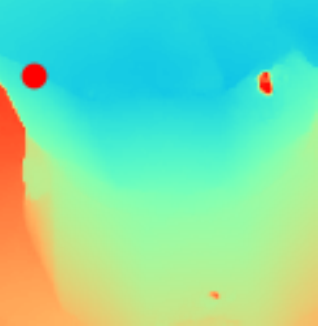}}
    
    \subfigure[DDR-Net (Ours)]{\label{Genelecs:Genelec 8030 AP}\includegraphics[width=1in,height=1in]{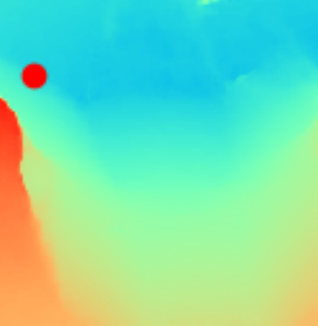}}
\end{minipage} 
\medskip
\hspace{-0.2cm}
\begin{minipage}[b]{0.2\linewidth}
\centering
\subfigure{\includegraphics[width=1.4in,height=1.1in]{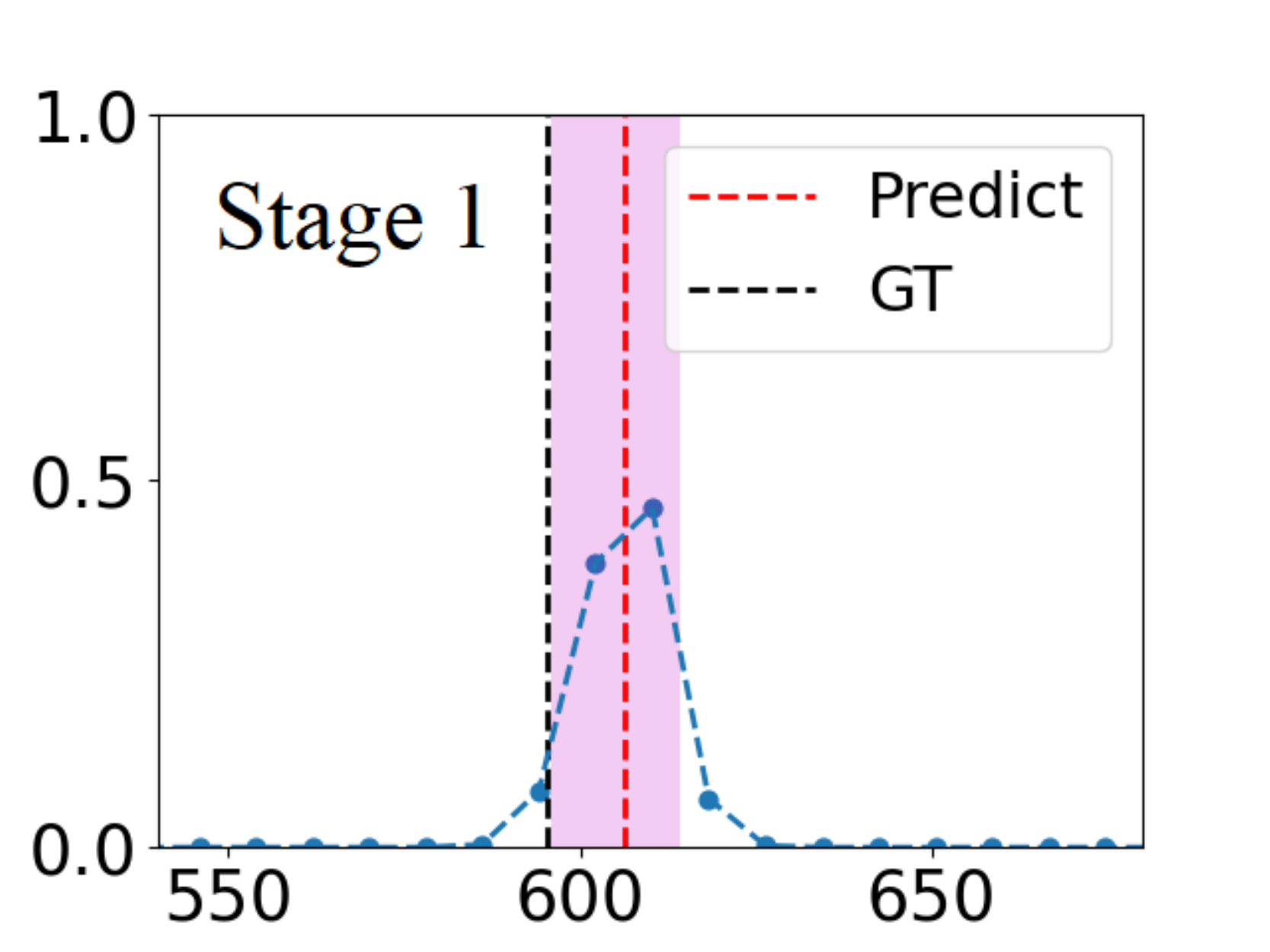}}\vskip 8pt
\subfigure{\includegraphics[width=1.4in,height=1.1in]{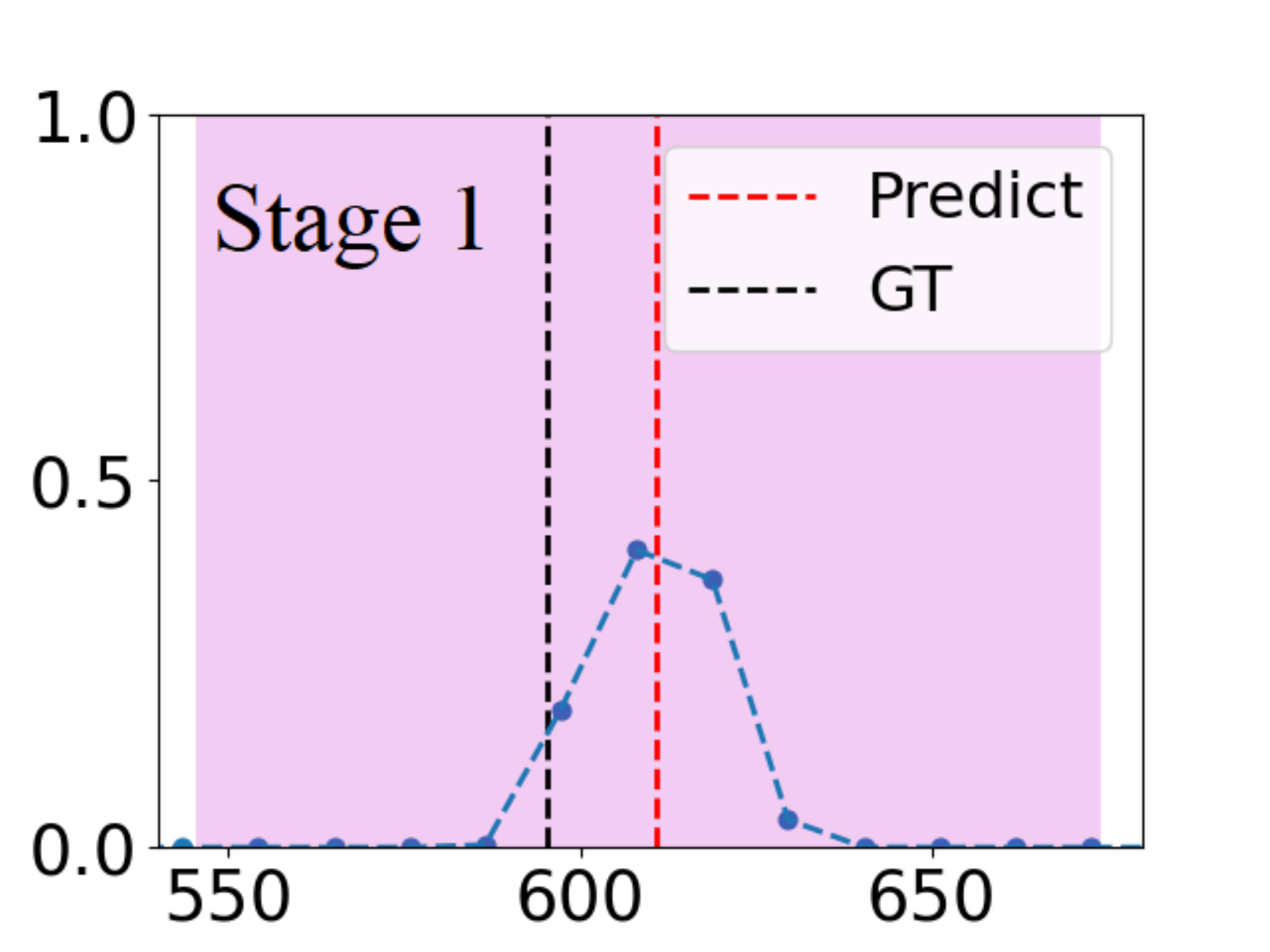}}\addtocounter{subfigure}{-2}

\end{minipage}
\hspace{-0.4cm}
\begin{minipage}[b]{0.1\linewidth}
\centering
\subfigure[Results from UCSNet~\cite{ucs}]{\includegraphics[width=1.4in,height=1.1in]{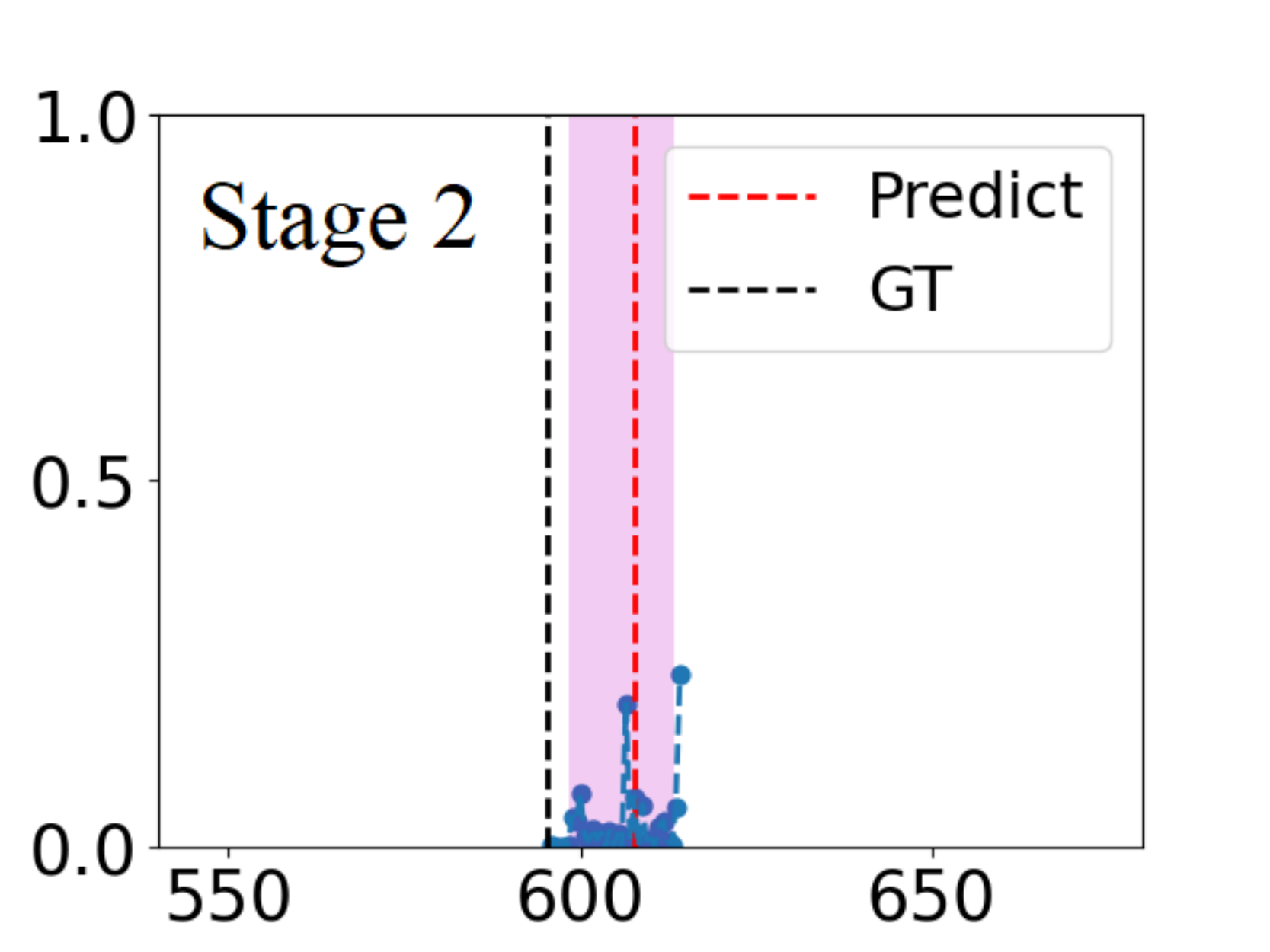}}\vskip -6pt
\subfigure[Results from DDR-Net]{\includegraphics[width=1.4in,height=1.1in]{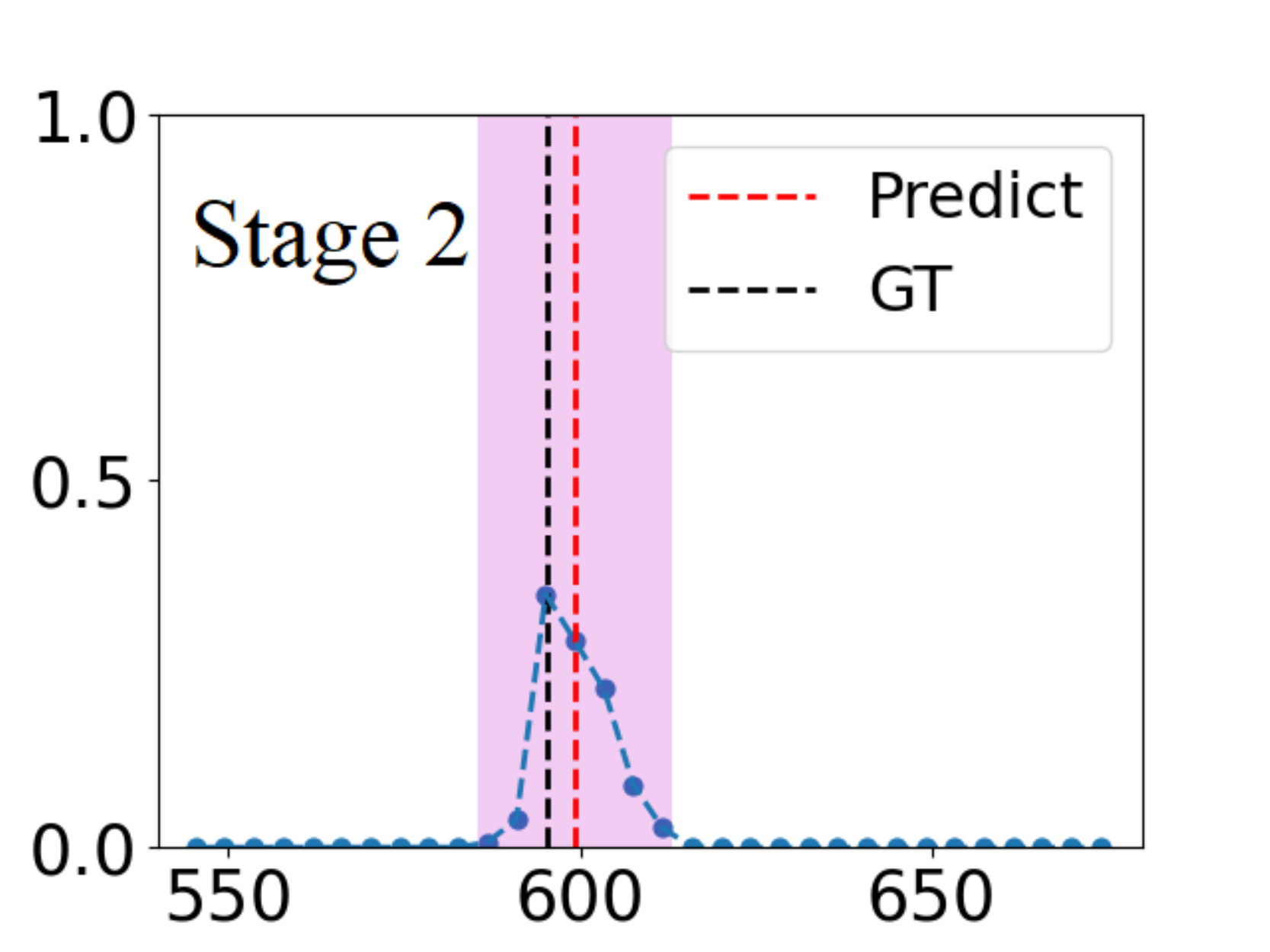}}
\end{minipage}
\hspace{1.4cm}
\begin{minipage}[b]{0.2\linewidth}
\centering
\subfigure{\includegraphics[width=1.4in,height=1.1in]{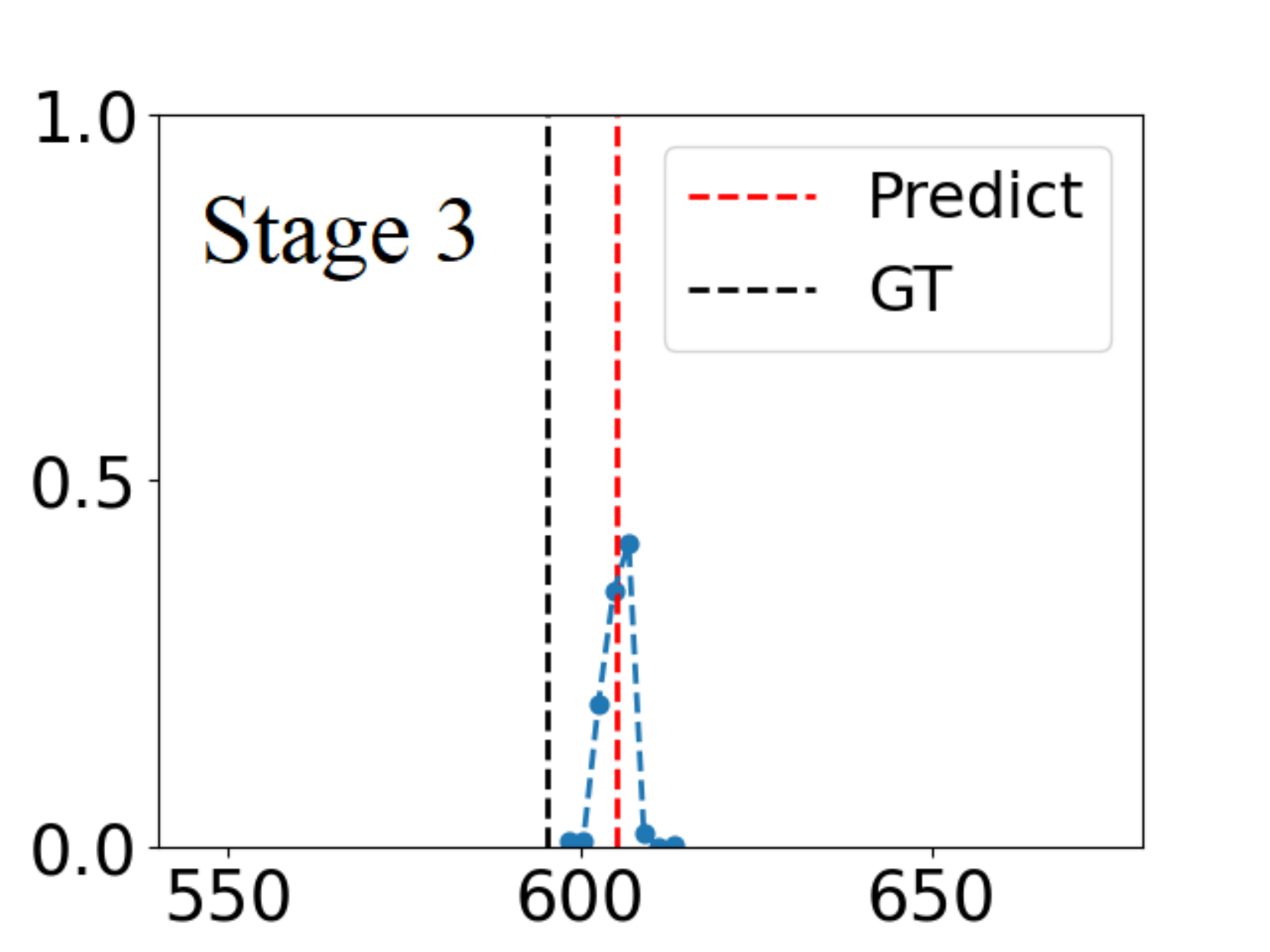}}\vskip 8pt
\subfigure{\includegraphics[width=1.4in,height=1.1in]{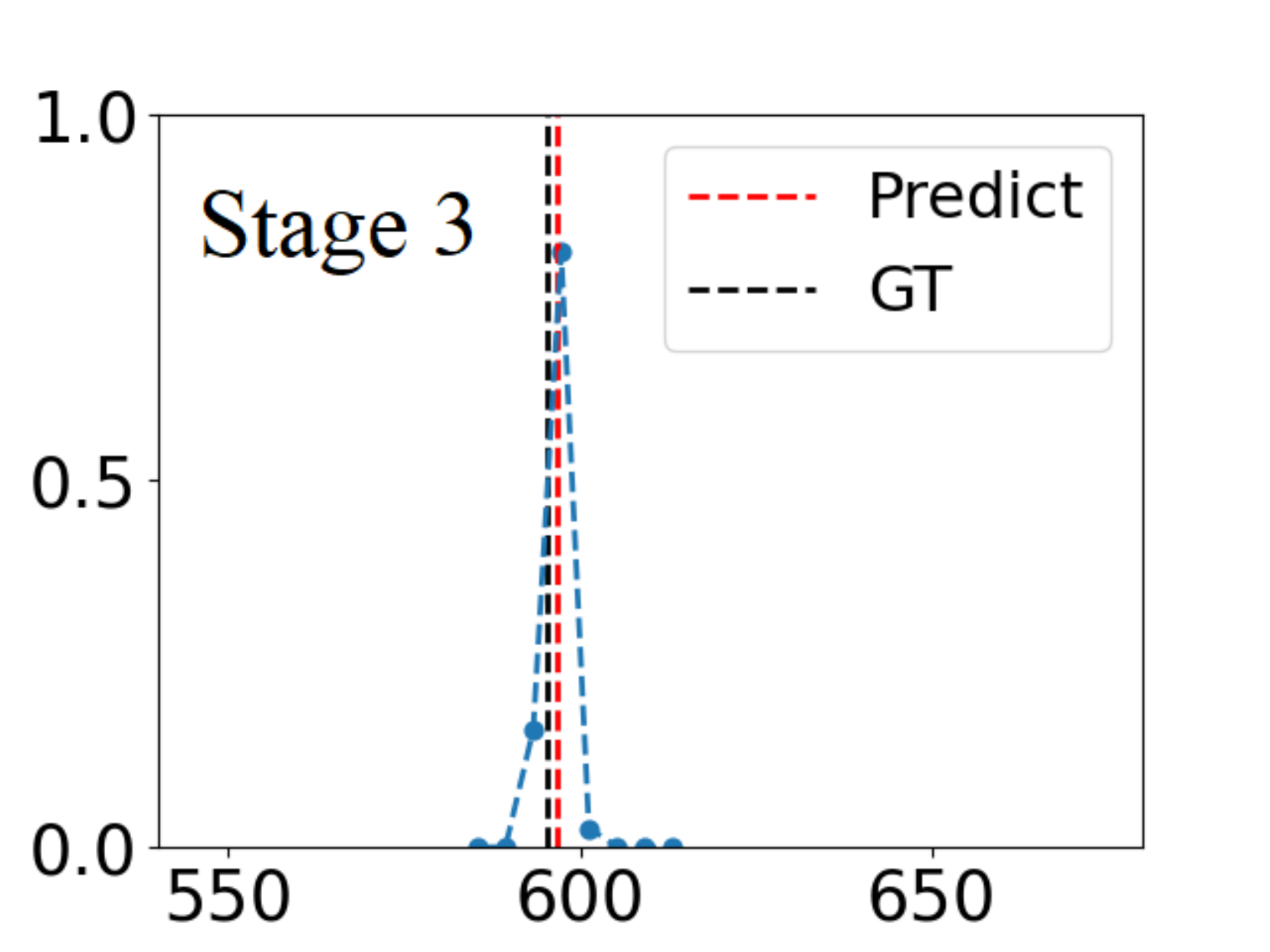}}
\end{minipage}
\vspace{-0.75em}
\caption{Illustration of detailed depth prediction and dynamic
depth range estimation of a specific example. On the left, we show
the RGB image crops (a), the ground truth depth (b), the predicted depth map (c) from UCSNet~\cite{ucs} and the predicted depth map (d) from our proposed DDR-Net. On the right, we show the details of a pixel (red points in the images) with predicted probability distribution (connected blue dots), the depth value prediction (red dash line), the ground truth depth (black dash line) and uncertainty intervals (pink) in the three stages. The top results (e) are from UCSNet~\cite{ucs} and the bottom results (f) are from our proposed DDR-Net.}
\label{fig:Genelecs}
\vspace{-1em}
\end{figure*}

\vspace{-0.5em}
\subsection{Multi-Scale Feature Extraction}
\label{Sec:MultiScaleFeatureExtraction}
\vspace{-0.25em}

Several previous works~\cite{pmvsnet,mvsnet,recurrent} utilize a shallow network to downsample original images and extract high-level features with semantic information. Due to the limit of GPU memory, the cost volume is built upon the final high-level features and produce the depth map with certain low resolution. In order to obtain high-resolution depth maps, we utilize a feature pyramid network (FPN) to extract multi-scale features. As shown in Figure~\ref{Fig:Architecture}, the network only consists of three stages with several convolutional layers. Given the input image $\mathbf{I}$ , the final outputs will be $\mathbf{O}_{1}$, $\mathbf{O}_{2}$, and $\mathbf{O}_{3}$ for building cost volumes. Let the resolution of input image be $W\times H\times C$, the corresponding resolutions of output features {$\mathbf{O}_{1}$, $\mathbf{O}_{2}$, and $\mathbf{O}_{3}$} will be {$W\times H\times32$, $\frac{W}{2}\times\ \frac{H}{2}\times32$, $\frac{W}{4}\times \frac{H}{4}\times32$}, respectively, where $W$, $H$ and $C$ denote width, height and channel, respectively. Our multi-scale feature extraction strategy utilizes both high-level semantic information and low-level structural information to produce high-resolution depth maps from coarse to fine.

\vspace{-0.5em}
\subsection{Cost Volume Construction}
\label{Sec:CostVolumeConstruction}
\vspace{-0.25em}

Similar to previous methods~\cite{ucs,cascade}, we construct cascade cost volumes in multi-stages by using the homography warping process in~\cite{mvsnet} to warp the features of source views to the reference view.  The warping function of the first stage is expressed as: 
\begin{equation}
    \mathbf{H}_{i}(d)=\mathbf{K}_{i}\mathbf{R}_{i} \bigg(\mathbf{I}-\frac{(\mathbf{t}_{1}-\mathbf{t}_{i}) \mathbf{n}_{1}^\top}{d}\bigg) \mathbf{R}_{1}^\top  \mathbf{K}_{1}^{-1},
\end{equation}
where $\mathbf{H}_{i}(d)$ refers to the homography between the feature maps of the $i$-th view and the reference feature map at the depth $d$. Moreover, $\mathbf{K}_{i}$, $\mathbf{R}_{i}$, and $\mathbf{t}_{i}$ denote the camera intrinsics, rotations and translations of the $i$-th view, respectively. And $\mathbf{n}_1$ denotes the principle axis of the reference camera. A variance based metric is then operated to aggregate arbitrary cost volumes of source views and the reference cost volume to a single one. 

In the following stages (the 2nd and the 3rd stages), we follow~\cite{cascade} to set the homography functions of $(k+1)$-th stage as:
\begin{equation}
    \mathbf{H}_{i}(d_{k}^m\!+\!\Delta_{k+1}^m)=\mathbf{K}_{i} \mathbf{R}_{i}\ \!\bigg(\mathbf{I}-\frac{(\mathbf{t}_{1}\!-\!\mathbf{t}_{i}) \mathbf{n}_{1}^\top}{d_{k}^m\!+\!\Delta_{k+1}^m}\!\bigg) \mathbf{R}_{1}^\top \mathbf{K}_{1}^{-1},
\end{equation}
%
where $d_{k}^m$ denotes the predicted depth of the $m$-th pixel at the $k$-th stage, and $\Delta_{k+1}^m$ is the residual depth of the $m$-th pixel to be learned at the $(k+1)$-th stage.

After constructing the cost volume, we apply 3D CNNs to regularize the cost volume to infer the probability volume and get the probability distribution of the depth. Following previous works, we utilize a 3D U-Net~\cite{ronneberger2015u} architecture proposed in~\cite{mvsnet}, which is able to aggregate context information in multi-scales. 

The acquired depth probability volume consists of $D_{k}$ depth probability maps $\{\mathbf{P}_{k,j}\}_{j=1}^{D_k}$, where $D_{k}$ is the number of the depth hypotheses (depth plane) of this stage. Each $\mathbf{P}_{k,j}$ is associated with a certain depth map $\mathbf{L}_{k,j}$. Then we use a differentiable $soft$ $argmin$ process to compute the depth value at the pixel $\mathbf{x}$ as:
\begin{equation}
    \hat{\mathbf{L}}_{k}(\mathbf{x})=\sum\nolimits_{j=1}^{D_{k}} \left(\mathbf{L}_{k,j}(\mathbf{x})\cdot \mathbf{P}_{k,j}(\mathbf{x})\right).
\end{equation}

\vspace{-0.5em}
\subsection{Dynamic Depth Range Estimation}
\label{Sec:DynamicDepthRangeEstimation}
\vspace{-0.25em}

Accurate and appropriate depth range estimation for each pixel on the 2nd and 3rd stages is essential for generating high quality point clouds. To this end, we propose a novel range estimation module (REM) to estimate the 2nd and the 3rd depth ranges dynamically for each pixel by using the information of a probability volume in the former stage.

Given a set of probability maps, previous methods only pay attention to the probability distribution of a single pixel to design the depth range in the next stage. However, we argue that both the context information and the probability distribution of the nearby pixels are also significant to determine the depth ranges of pixels. The illustration of this assumption is presented in the supplementary details. Also, the information of a probability distribution is not fully utilized in previous methods which leverage numerical characteristics to measure the uncertainties. To solve this problem, we design a range estimation module (REM) to estimate the dynamic depth range adaptively. In particular, we utilize the previous probability volume and apply a single network to learn the uncertainties from it. As shown in Figure \ref{Fig:Architecture}, our REM is composed of several simple 2D CNNs and a sigmoid activation function. The sigmoid function restricts the values of final output from 0 to 1. Note that the REM is light-weighted but quite effective for the following depth range hypotheses estimation. With the convolutional network, the reception field is enlarged and the information of nearby pixels is fully considered for the following depth estimation. The final output of the network is an uncertainty map $\mathbf{C} = \{\mathbf{C}(\mathbf{x})\}_{\mathbf{x}\in \mathbf{I}}$ where $\mathbf{C}(\mathbf{x})$ refers to the uncertainty of the depth distribution of the pixel $\mathbf{x}$ in the image $\mathbf{I}$. The full network architecture of our proposed range estimation module (REM) is presented in the supplementary details.

Given an uncertainty value $\mathbf{C}(\mathbf{x})$ obtained by REM and the previous depth prediction $\mathbf{L}(\mathbf{x})$ at the pixel $\mathbf{x}$, we use the following equation to determine the depth range $D(\mathbf{x})$ in the next stage:
\begin{equation}
    D(\mathbf{x})=[\mathbf{L}(\mathbf{x})-\lambda \mathbf{C}(\mathbf{x}), \mathbf{L}(\mathbf{x})+\lambda \mathbf{C}(\mathbf{x})],\label{Eq:DepthRange}
\end{equation}
where $\lambda$ is a hyperparameter that determines how large the confidence interval is. For a pixel $\mathbf{x}$, we uniformly sample the depth values in the depth range $D(\mathbf{x})$. In this way, we construct dynamic depth hypotheses for each pixel in the next stage. It is noteworthy that our whole range estimation module (REM) is light-weighted and time-saving by only adding a few parameters.

To demonstrate the effectiveness of our module and compare with UCSNet~\cite{ucs}, we show a specific example with a pixel and its estimated uncertainty intervals around the prediction in Figure~\ref{fig:Genelecs}. Both methods are not able to obtain satisfactory depth prediction on the first stage. Note that UCSNet fails to obtain accurate depth prediction because of its inappropriate uncertainty estimation determined by variance. Its depth range estimation fails to cover ground truth on the second stage. However, our proposed REM which leverages probability volume of the former stage and take nearby pixels into consideration, learns depth range hypotheses with a higher confidence and adjust to the probability prediction at each stage to achieve better optimized intervals. In general, our DDR-Net has a better spatial partitioning and enables to reconstruct the final point clouds with higher completeness and accuracy. 
\vspace{-0.5em}
\subsection{Loss Strategy}
\label{Sec:LossStrategy}
\vspace{-0.25em}

Range estimation module (REM) is developed to dynamically estimate the depth range of a certain scene for the second and the third stages in our proposed DDR-Net. However, the estimated depth range may still not accurate enough to cover ground truth depth values of all pixels. To this end, we design a novel loss strategy to refine our range estimation module (REM) and produce better depth ranges with higher confidences.

\begin{figure}[t]
\centering
\subfigure[Original]{
\begin{minipage}[t]{0.32\linewidth}
\centering
\includegraphics[width=1.2in,height=1in]{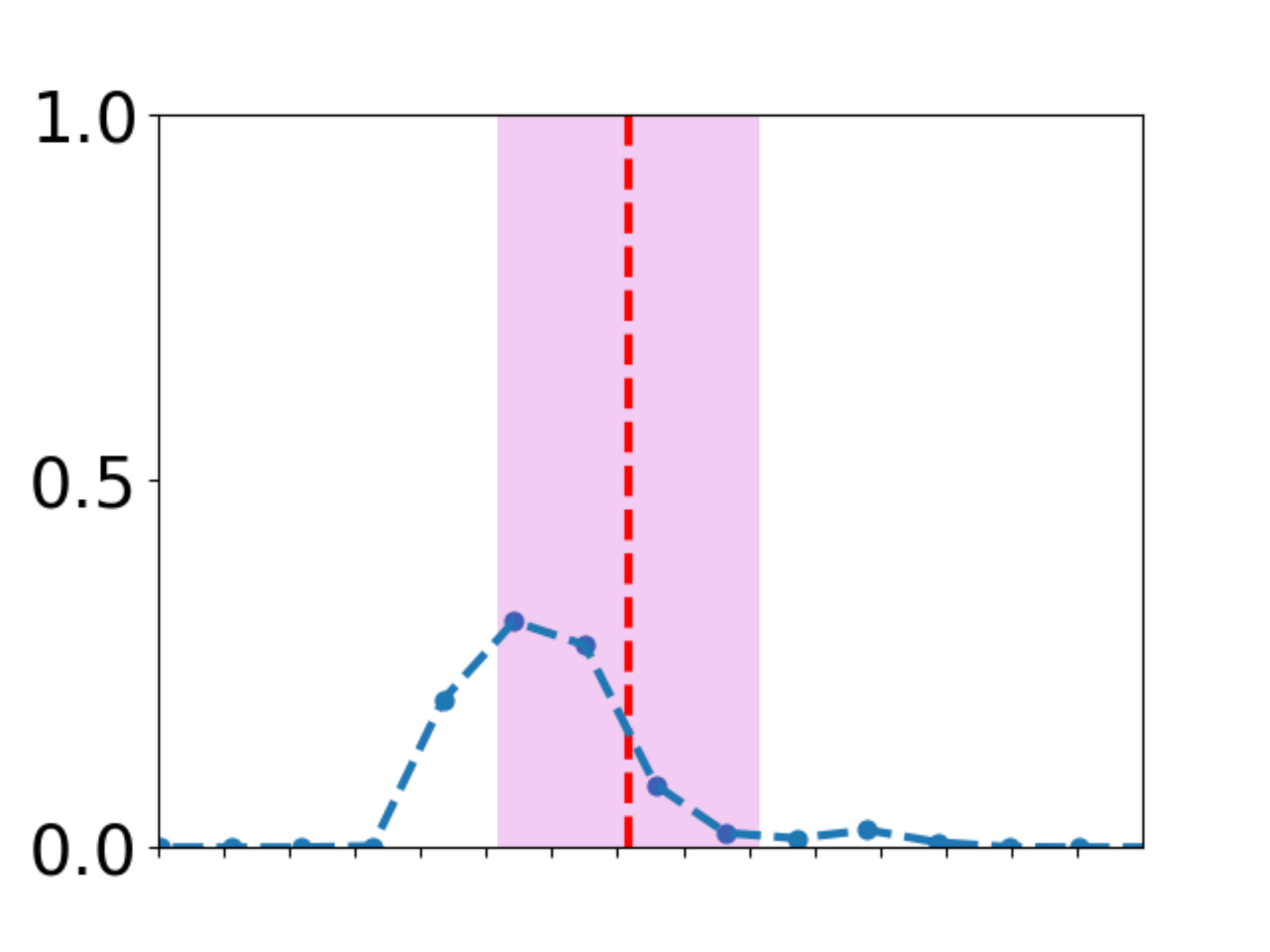}
\end{minipage}
}%
\subfigure[Clamp]{
\begin{minipage}[t]{0.32\linewidth}
\centering
\includegraphics[width=1.2in,height=1in]{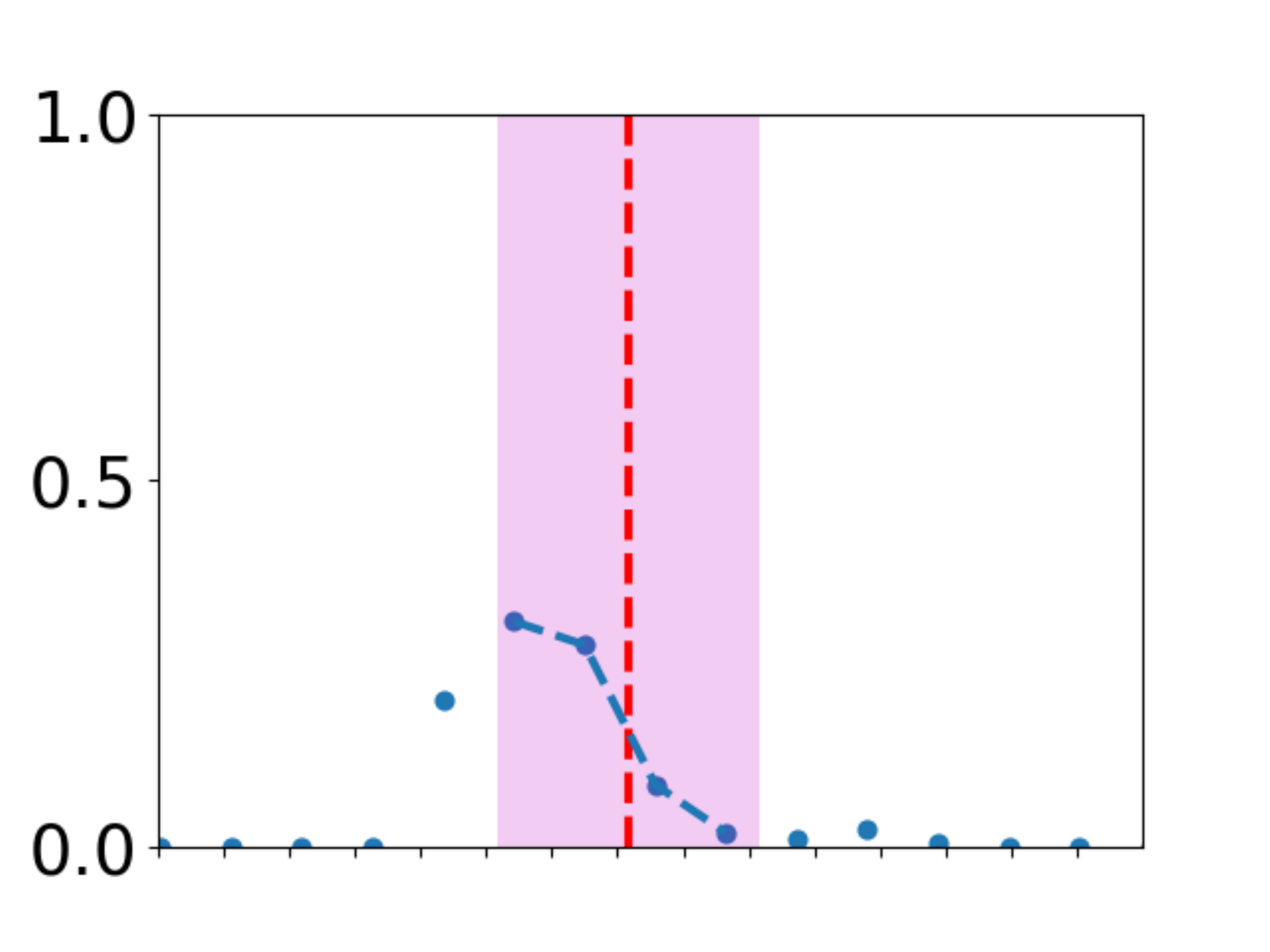}
\end{minipage}
}%
\subfigure[Normalization]{
\begin{minipage}[t]{0.32\linewidth}
\centering
\includegraphics[width=1.2in,height=1in]{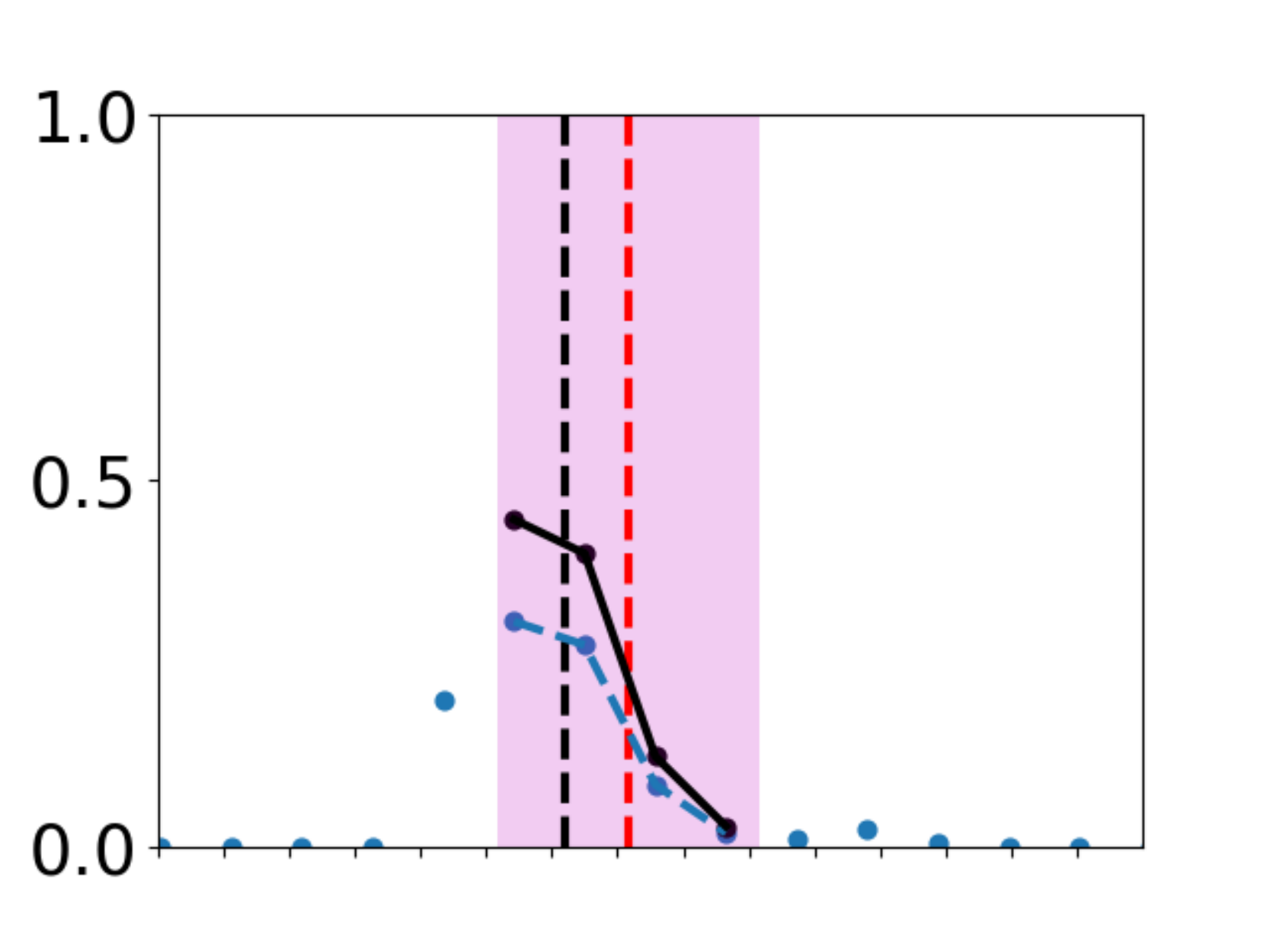}
\end{minipage}
}%
\centering
\caption{Illustration of our novel loss strategy. In (a), we show the probability distribution of a specific pixel (connected blue dots), depth prediction (red dash line) and uncertainty intervals (pink areas). Then we utilize clamp strategy to cut the original probability distribution as illustrated in (b). Then the refined probability distribution will be normalized and output the refined depth value (black dash line).}
\label{Fig:ClampStrategy}
\vspace{-0.5cm}
\end{figure}

As shown in Figure~\ref{Fig:ClampStrategy}, given the maximum and minimum depth values of a new dynamic depth range $D_{\max}^{k+1}$ and $D_{\min}^{k+1}$ in the 2nd stage (or 3rd stage), the probability volume $\mathbf{P}_{k,j}$ and the depth range hypotheses $\mathbf{L}_{k,j}$ in the former stage, the refined depth range hypotheses and the probability volume are obtained through the following clamp operations as:
\begin{equation}
    \mathbf{L}_{k,j}^{\rm{refined}}(\mathbf{x})\!=\!\begin{cases} \mathbf{L}_{k,j}(\mathbf{x}),&D_{\rm{min}}^{k+1}\!<\!\mathbf{L}_{k,j}(\mathbf{x})\!<\!D_{\rm{max}}^{k+1}\\
    \\0,&\rm{else}
    \end{cases}
\end{equation}
\begin{equation}
    \mathbf{P}_{k,j}^{\rm{refined}}(\mathbf{x})=\begin{cases} \mathbf{P}_{k,j}(\mathbf{x}), &\mathbf{L}_{k,j}^{\rm{refined}}(\mathbf{x})\neq 0\\
    \\0. &\rm{else}\\
    \end{cases}
\end{equation}
After clamp operations, the refined probability volumes will be normalized to ensure that the sum probability in a single pixel is one. With refined depth range hypotheses and refined probability volumes, the final refined depth map is calculated as:
\begin{equation}
    \hat{\mathbf{L}}_{k}^{\rm{refined}}(\mathbf{x})=\sum\nolimits_{j=1}^{D_k}(\mathbf{L}_{k,j}^{\rm{refined}}(\mathbf{x})\cdot \mathbf{P}_{k,j}^{\rm{refined}}(\mathbf{x})).
\end{equation}
The resolutions of refined depth maps are the same as the output depth maps in the 1st and 2nd stages, which are  $\frac{W}{4}\times \frac{H}{4} $ and $\frac{W}{2}\times \frac{H}{2} $, respectively. The ground truth will be downsampled to the same resolutions.

We utilize the smooth L1 loss as our loss function for both initial depth maps and refined maps. The final loss function is:
\begin{equation}
    Loss=\sum\nolimits_{k=1}^{3}\alpha_{k} Loss_{k}+ \sum\nolimits_{k=1}^{2}\beta_k Loss_{k}^{\rm{refined}},\label{Eq:Loss}
\end{equation}
where $Loss_{k}$, $Loss_{k}^{\rm{refined}}$, $\alpha_{k}$, and $\beta_k$ denote the loss and refined loss of the $k$-th stage and their corresponding weights, respectively.

We assume that the depth prediction of each stage should have similar probability distribution even their depth sampling rates progressively increase. Following this assumption, our loss strategy produces a rough depth prediction of the new stage in advance by leveraging new estimated depth range hypotheses. If the estimated depth range does not contain the ground truth depth value of each pixel, the output refined depth map will have a deviation compared to the ground truth depth map. In this way, our range estimation module (REM) can learn an appropriate depth range hypotheses with high confidences to adjust to the probability distribution. Note that the whole operation is naturally differentiable and can be implemented as a refine module in any multi-view stereo framework in a coarse-to-fine manner without adding any additional parameters. 

\vspace{-1.0em}
\section{Experiments}
\label{Sec:Experiments}
\vspace{-0.5em}

In this section, we evaluated our proposed DDR-Net in the standard benchmark of multi-view stereo with a thorough set of experiments to demonstrate  effectiveness and efficiency of our model.

\vspace{-0.5em}
\subsection{Datasets}
\label{Sec:Datasets}
\vspace{-0.25em}

{\bf DTU dataset}~\cite{dtu} is a large-scale MVS dataset with 124 scenes scanned from 49 or 64 views under 7 different lighting conditions. Each scene provides a reference point cloud and camera parameters of the captured images. Following CasMVSNet~\cite{cascade}, we generated the depth maps for each view in three different image resolutions as our training ground truth. We used the same training, validation and evaluation sets as defined in ~\cite{mvsnet}.

{\bf Tanks and Temples}~\cite{tanks} contains realistic scenes in different scales. To compare with other state-of-the-art methods, we evaluated our model on its intermediate set which consists of 8 scenes.

\begin{table}[t]  
\centering
\small
\begin{tabular*}{6.5cm}{l|lll}  
\hline  
Methods &Acc. &Comp. &Overall\\ 
\hline  
Camp~\cite{camp}  & 0.835 & 0.554 & 0.695\\  
Furu~\cite{mvspointcloudmethod1}  & 0.613 & 0.941 & 0.777\\
Tola~\cite{tola2012efficient}  & 0.342 & 1.190 & 0.766\\ 
Gipuma~\cite{gipuma}  & {\bf 0.283} & 0.873 & 0.578\\
MVSNet~\cite{mvsnet}  & 0.396 & 0.527 & 0.462\\
R-MVSNet~\cite{recurrent}  & 0.383 & 0.452 & 0.417\\
Point-MVSNet~\cite{point-based}  & 0.342 & 0.411 & 0.376\\
CasMVSNet~\cite{cascade} & 0.346 & 0.351 & 0.348\\
CVP-MVSNet~\cite{cvp} & 0.296 & 0.406 & 0.351\\
AttMVS~\cite{attention} & 0.383 & 0.329 & 0.356\\
Vis-MVSNet~\cite{visibility} & 0.369 & 0.361 & 0.365\\
UCSNet~\cite{ucs} & 0.338 & 0.349 & 0.344\\
\hline  
DDR-Net (Ours) &0.339 &{\bf 0.320} &{\bf 0.329}\\
\hline
\end{tabular*} 
\caption{Quantitative results of point cloud reconstruction on the DTU dataset~\cite{dtu}. DDR-Net achieves best completeness and overall score compared with other state-of-the-art methods}  
\label{Tab:DTUScores}
\vspace{-0.5cm}
\end{table}

\begin{figure*}[htbp]
\begin{center}
\subfigure[CasMVSNet~\cite{cascade}]{
\begin{minipage}[t]{0.25\linewidth}
\centering
\includegraphics[height=2.2in,width=1.65in]{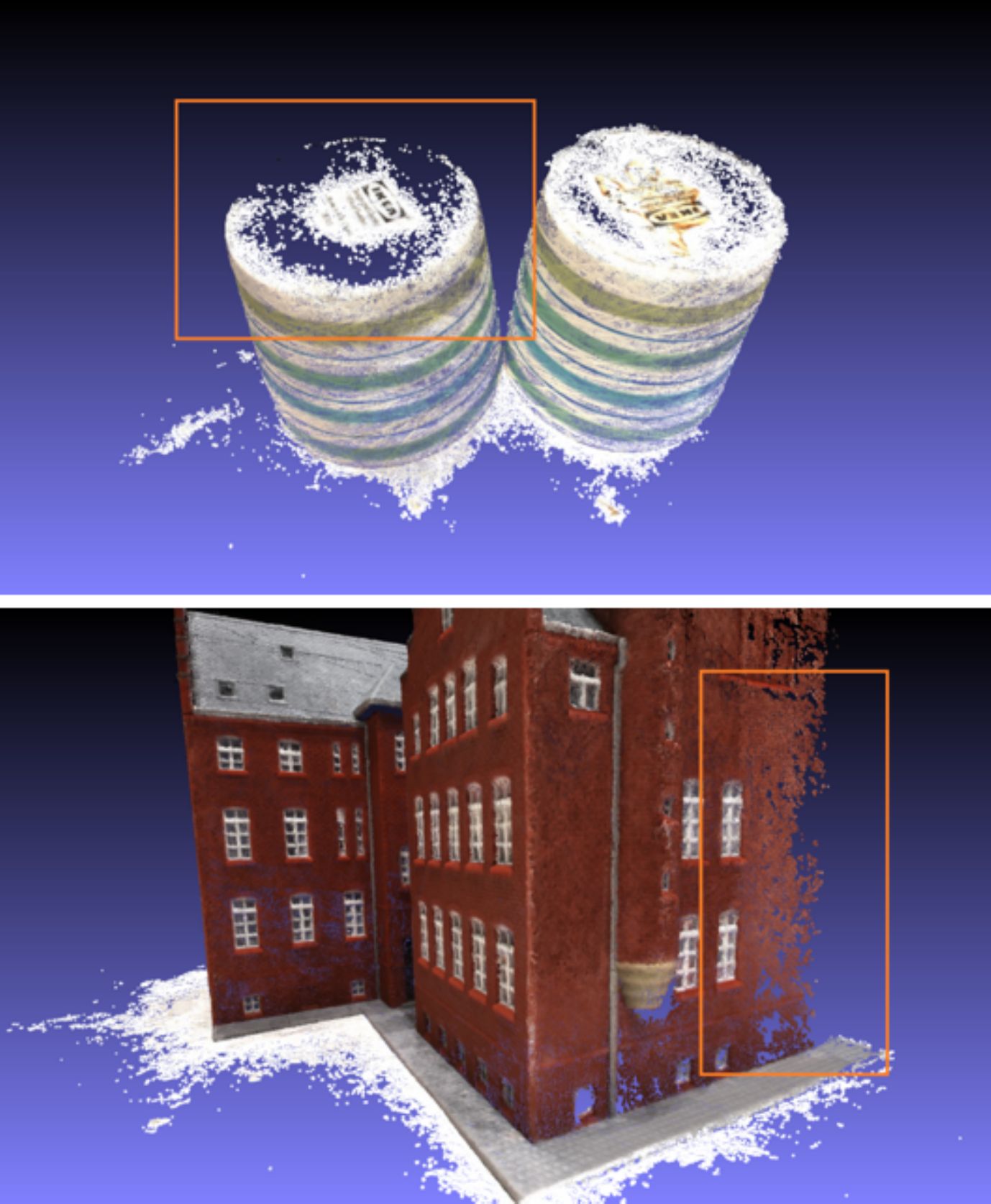}
\end{minipage}%
}%
\subfigure[DDR-Net (Ours)]{
\begin{minipage}[t]{0.25\linewidth}
\centering
\includegraphics[height=2.2in,width=1.65in]{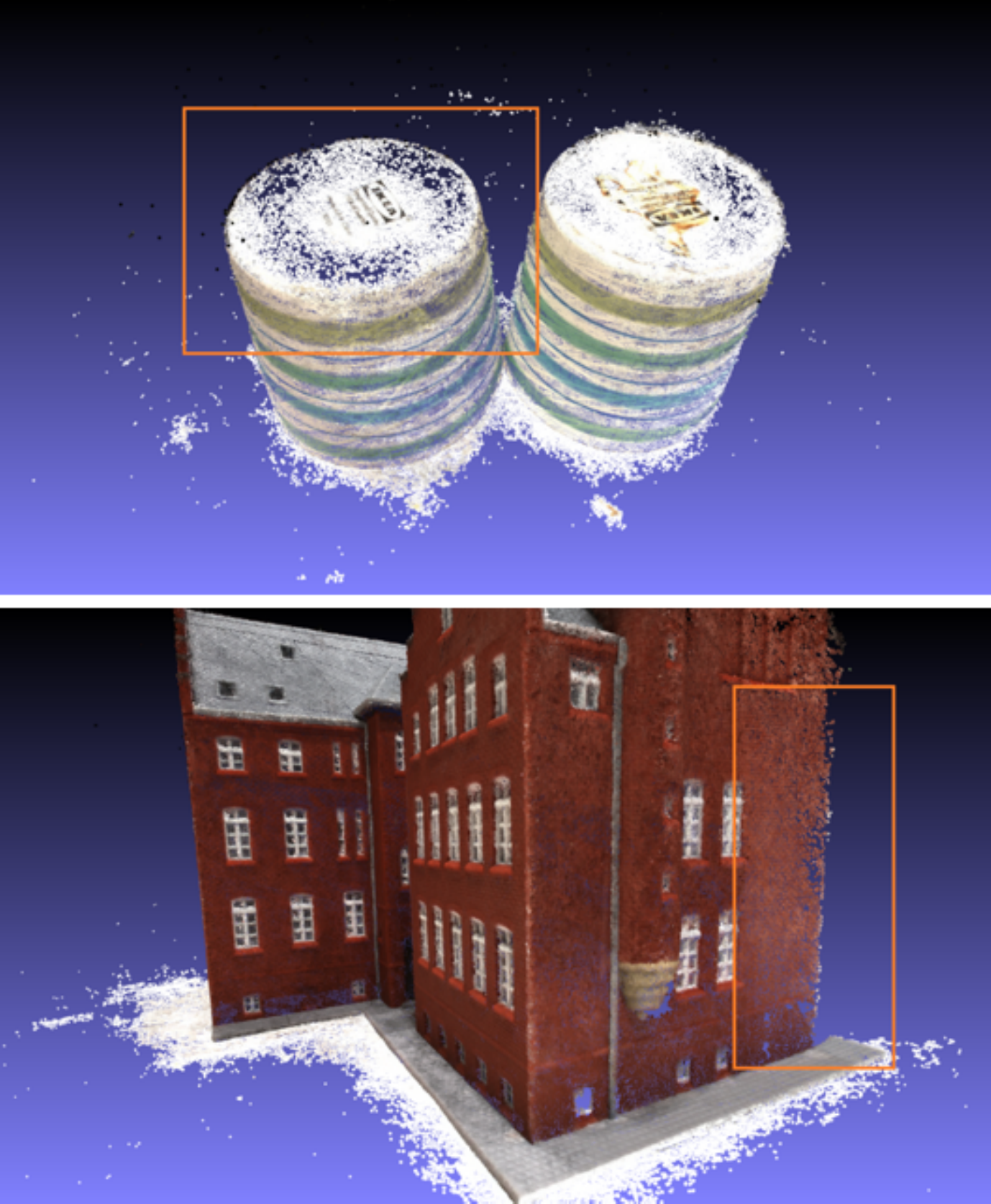}
\end{minipage}%
}%
\subfigure[Ground Truth]{
\begin{minipage}[t]{0.25\linewidth}
\centering
\includegraphics[height=2.2in,width=1.65in]{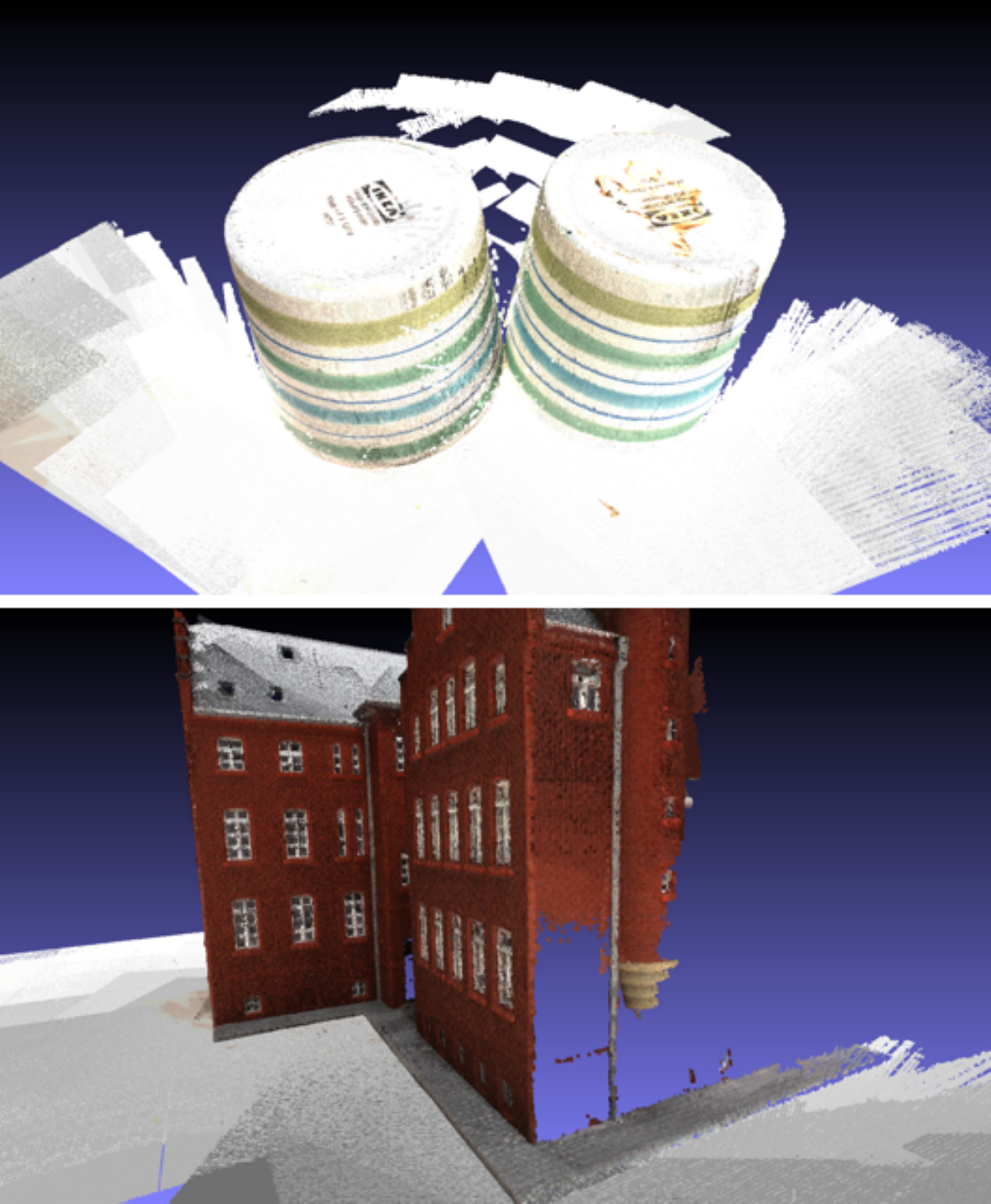}
\end{minipage}
}%
\end{center}
\vspace{-1em}
\caption{Different results of point cloud reconstruction on the DTU dataset~\cite{dtu}. As shown in orange boxes, our reconstruction generates more complete point cloud with better details, which demonstrates the effectiveness of our method. }
\vspace{-0.5em}
\label{Fig:DTUPointCloud}
\end{figure*}

\begin{table*}[t]
\footnotesize
\centering
\begin{tabular*}{17.2cm}{l|l|l|l|l|l|l|l}
\hline  
Methods &Input Sizes &Depth Map Sizes &Acc. (mm) &Comp. (mm) &Overall (mm) &GPU Memory (MB) &Runtime (s) \\ 
\hline  
MVSNet~\cite{mvsnet}	&1600$\times$1184	&400$\times$288	&\hspace{0.25cm}0.396	&\hspace{0.35cm}0.527	&\hspace{0.4cm}0.462	&\hspace{0.7cm}22511	&\hspace{0.25cm}2.76\\
Point-MVSNet~\cite{point-based}	&1600$\times$1184	&800$\times$576	&\hspace{0.25cm}0.342	&\hspace{0.35cm}0.411	&\hspace{0.4cm}0.376	&\hspace{0.7cm}13081	&\hspace{0.25cm}3.04\\
CVP-MVSNet~\cite{cvp}	&1600$\times$1184	&1600$\times$1184	&\hspace{0.25cm}\bf{0.296}	&\hspace{0.35cm}0.406	&\hspace{0.4cm}0.351	&\hspace{0.7cm}8795	&\hspace{0.25cm}1.72\\
CasMVSNet~\cite{cascade}	&1600$\times$1184	&1600$\times$1184	&\hspace{0.25cm}0.346	&\hspace{0.35cm}0.351	&\hspace{0.4cm}0.348	&\hspace{0.7cm}10153	&\hspace{0.25cm}0.89\\
UCSNet~\cite{ucs}	&1600$\times$1184	&1600$\times$1184	&\hspace{0.25cm}0.338	&\hspace{0.35cm}0.349	&\hspace{0.4cm}0.344	&\hspace{0.7cm}7525	&\hspace{0.25cm}\bf{0.87}\\
\hline
DDR-Net (Ours)	&1600$\times$1184	&1600$\times$1184	&\hspace{0.25cm}0.339	&\hspace{0.35cm}\bf{0.320}	&\hspace{0.4cm}\bf{0.329} &\hspace{0.7cm}\bf{7345} &\hspace{0.25cm}0.88	\\
\hline
\end{tabular*} 
\caption{Comparison of reconstruction quality, GPU memory usage and runtime on the DTU dataset for different input sizes. For the same size of input images, our proposed DDR-Net achieves high quality point cloud reconstruction with smallest GPU memory consumption and reasonable running time.}
\label{Tab:DTUMemoryTime}
\end{table*}

\begin{table*}[h!]
\footnotesize
\centering
\begin{tabular*}{14.2cm}{l|lllllllll}
\hline  
Methods &Mean &Family &Francis &Horse &Lighthouse &M60 &Panther&Playground&Train\\ 
\hline  
MVSNet~\cite{mvsnet}  &43.48 &55.99 &28.55 &25.07 &50.79 &53.96 &50.86 &47.90 &34.69\\ 
R-MVSNet~\cite{recurrent} &48.40 &69.96 &46.65 &32.59 &42.95 &51.88 &48.80 &52.00 &42.38\\
CVP-MVSNet~\cite{cvp}  &54.03 &{\bf 76.50} &47.74 &36.34 &55.12 &{\bf57.28} &{\bf54.28} &\underline{57.43} &\underline{47.54}\\ 
CasMVSNet~\cite{cascade}  &{\bf 56.42} &\underline{76.36} &{\bf58.45} &{\bf46.20} &{\bf55.53} &56.11 &\underline{54.02} &{\bf58.17} &46.56\\
UCSNet~\cite{ucs}  &54.83 &76.09 &53.16 &43.03 &54.00 &55.60 &51.49 &57.38 &{\bf47.89}\\
\hline  
DDR-Net (Ours) & \underline{54.91} & 76.18 & \underline{53.36} & \underline{43.43} & \underline{55.20} & 55.57 & 52.28 & 56.04 & 47.17\\
\hline
\end{tabular*} 
\caption{Results of point cloud reconstruction on the Tanks and Temples dataset~\cite{tanks}.}  
\label{Tab:TanksTemplesScores}
\vspace{-0.5cm}
\end{table*}

\begin{figure*}[t]
\begin{center}
\label{tankpoint}
\includegraphics[width=1\textwidth]{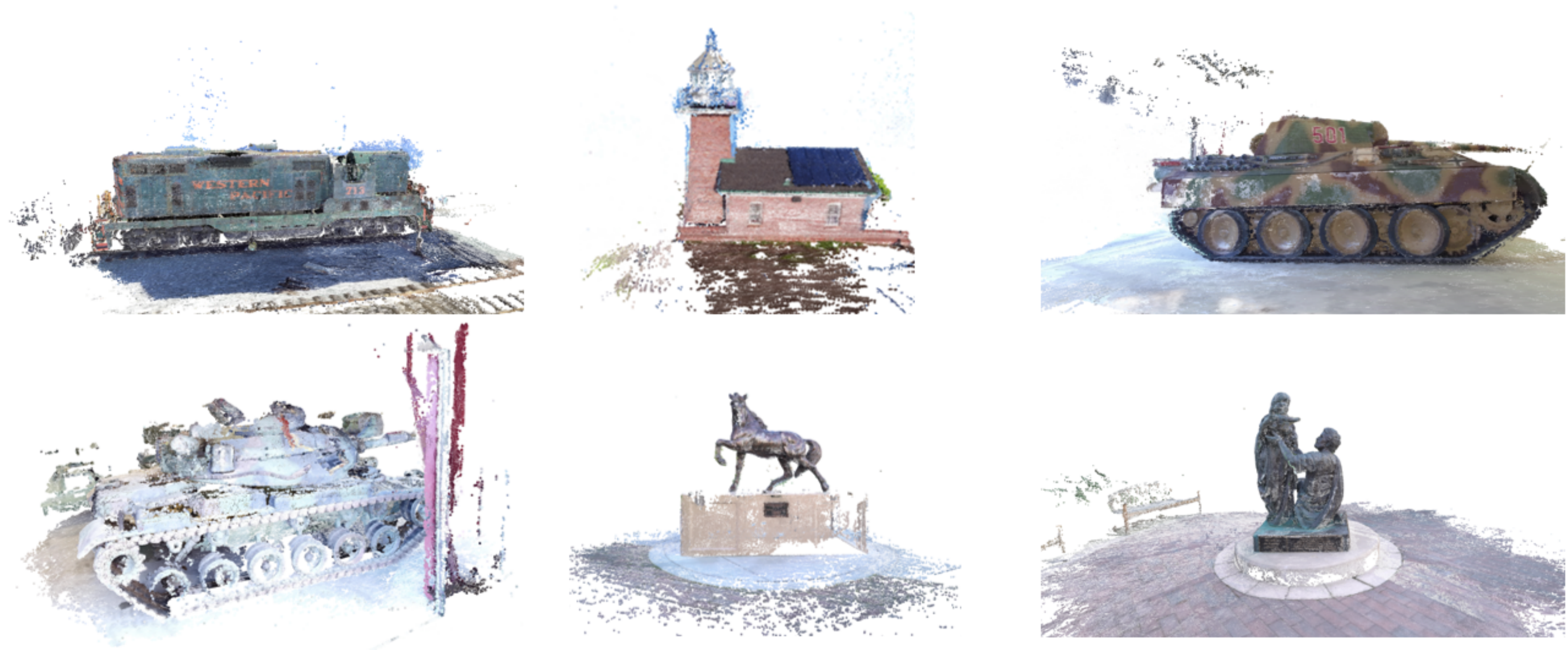}
\end{center}
\vspace{-1em}
\caption{The point cloud reconstruction on the Tanks and Temples dataset~\cite{tanks}.}
\label{Fig:TanksTemplesPointCloud}
\vspace{-0.5em}
\end{figure*}

\vspace{-0.5em}
\subsection{Implementation}
\label{Sec:Implementation}
\vspace{-0.25em}

\noindent {\bf Training.}  We trained our DDR-Net on DTU dataset for all following experiments. The ground truth depth maps were generated in different image resolutions ($\frac{W}{4} \times \frac{H}{4}$, $\frac{W}{2} \times \frac{H}{2}$ and $W \times H$) for three stages by down-sampling the original images. We set the resolution of input images to $640\times512$ and the number of neighbouring views to 3. The the same source view selection strategy as in~\cite{mvsnet} was used in both training and evaluation. The numbers of depth planes were set to be 48, 32, and 8 for three stages, respectively. The loss weights for three stages were set as $\alpha_{1} = 0.5$, $\alpha_{2} = 1.0$, and $\alpha_{3} = 2.0$, and the weights of refined losses were set as $\beta_1 = 3.0$ and $\beta_2 = 0.0$. The experiments of setting the weights of refined losses are presented in the supplementary details. The parameter $\lambda$ in Eq.~(\ref{Eq:DepthRange}) which determines the confidence interval was set to 1.5 and 0.75 for the 1st and the 2nd stages. We implemented our work using PyTorch and used the ADAM optimizer to train our model. The whole network was trained for 16 epochs with a batch size of 2 on four Nvidia GTX 1080Ti cards. The initial learning rate was set to 0.001 and was halved iteratively at the 10th, 12th and 14th epochs.

\noindent {\bf Evaluation.}  Similar to previous methods, we used 5 neighbouring views for our testing and set the input image size to be $1600 \times 1184$. For the DTU benchmark, we calculated the accuracy and completeness to demonstrate the effectiveness of our model. We also calculated the averages of accuracy and completeness as the overall scores to measure the overall quality of the reconstruction point clouds. For Tanks and Temples, we used F-score as the evaluation metric to test our the generalization ability of our model.

\begin{table*}[t]
\centering
\begin{tabular*}{16cm}{l|l|l|l|l|l|l}  
\hline  
Methods &1st Range&2nd Range&2nd Stage Ratio&3rd Range&3rd Stage Ratio&Depth Plane\\ 
\hline  
CasMVSNet~\cite{cascade}  & 508.8mm	&169.72mm	&0.9532	&21.09mm	&0.8441	&48,32,8\\
UCSNet~\cite{ucs}  &508.8mm	&29.46mm	&0.8507	&10.10mm	&0.7310	&64,32,8
\\
\hline
Ours (REM) &508.8mm	&145.96mm	&0.9490	&20.91mm	&0.8391	&48,32,8\\
Ours (REM+Loss) &508.8mm	&139.46mm	&0.9317	&19.24mm	&0.8435	&48,32,8
\\
\hline
\end{tabular*} 
\caption{Evaluation of the dynamic depth range estimation in the 1st, 2nd and 3rd stages for our proposed DDR-Net with REM and REM+Loss models compared with CasMVSNet~\cite{cascade} and UCSNet~\cite{ucs}.}  
\vspace{-0.5cm}
\label{Tab:DDRRatios}
\end{table*} 

\vspace{-0.5em}
\subsection{Results on the DTU Dataset}
\label{Sec:DTUResults}
\vspace{-0.25em}

In this section, we compared our proposed DDR-Net with both traditional methods and recent learning-based methods. Qualitative comparisons are shown in Table~\ref{Tab:DTUScores}. While Gipuma~\cite{gipuma} achieves the best performance in terms of accuracy, our proposed DDR-Net outperforms other state-of-the-art methods in both completeness and overscore and ranks the 1st place of the quality of reconstruction point cloud on the DTU datasets. As shown in  Figure~\ref{Fig:DTUPointCloud}, our DDR-Net generates more complete point clouds in the areas with less textures, for example the boundaries of the church and the white cup covers, which demonstrate the effectiveness of our DDR-Net.

To demonstrate the efficiency of our method, we further compare our method with other multi-view stereo frameworks in terms of both GPU memory usage and runtime. As shown in Table~\ref{Tab:DTUMemoryTime}, our DDR-Net has a smaller memory usage and similar testing runtime compared with UCSNet~\cite{ucs} but achieves better point cloud reconstruction results. In addition, our approach can produce better results with smaller memory consuming (27.6$\%$ smaller) and lower running time compared with CasMVSNet~\cite{cascade}. These sufficiently demonstrate the benefit of our novel range estimation module (REM), in terms of system resource usage. Our DDR-Net combined with REM obtains high quality point cloud results with high computation and memory efficiency.
More visualized results of the reconstruction point cloud are presented in the supplementary material.

\vspace{-0.5em}
\subsection{Results on Tanks and Temples}
\label{Sec:TanksTemplesResults}
\vspace{-0.25em}

Here, we evaluated our method on the intermediate set of Tanks and Temples to test the generalization ability of our model. We used the DTU dataset to train our network and reconstructed the point cloud in the Tanks and Temples dataset without any fine-tuning. The source image number is set to $N = 5$ for network inference.	Quantitative results are shown in Table~\ref{Tab:TanksTemplesScores}. Comparing with other state-of-the art methods that utilize the similar coarse-to-fine strategy, our DDR-Net achieves a comparable F-score (54.91). In particular, our method obtains a higher F-score than UCSNet~\cite{ucs}, which is the best baseline on the DTU dataset and is only lower than CasMVSNet~\cite{cascade}. Note that CasMVSNet leverage more depth planes (160 planes for the first stage) to increase the depth sampling rate while our method only focuses on improving the estimation of depth range and utilizes the original depth plane setting (48,32,8) to construct the point cloud. Qualitative results of our point cloud reconstruction are shown in Figure~\ref{Fig:TanksTemplesPointCloud}.

\vspace{-0.5em}
\subsection{Ablation Study}
\label{Sec:AblationStudy}
\vspace{-0.25em}

In this section, we show extensive experimental results to verify the performance of range estimation module (REM) and the novel loss strategy in our proposed DDR-Net.

\noindent {\bf Effectiveness of REM.} Our designed range estimation module (REM) aims to learn the dynamic depth range for each pixel more precisely. We evaluated our network on the DTU testing dataset. Table~\ref{Tab:DDRRatios} shows the average lengths of dynamic depth ranges and the ratios of the pixels whose estimated depth ranges cover the ground truth depth values in our proposed DDR-Net. The corresponding values of both CasMVSNet~\cite{cascade} and UCSNet~\cite{ucs} are also shown in Table~\ref{Tab:DDRRatios}. To have a fair comparison, the input image resolution was set to $1152\times864$ and their original testing parameters were utilized to complete this experiment.
In addition, we observe that our full model (REM+Loss) is able to learn the dynamic depth range more reasonably. While UCSNet~\cite{ucs} covers short depth ranges (only 29.46mm in the second stage), its ratios (85.07$\%$ and 73.1$\%$)  of covering the ground truth depth values are not sufficient to output accurate depth map prediction. CasMVSNet~\cite{cascade} obtains the highest ratios (95.32$\%$ and 84.41$\%$), but the lengths of its depth ranges for all pixels are not spatially varying. Their depth range hypotheses are not able to adapt to conditions with different uncertainties. Moreover, their depth ranges are longer than ours, which means both lower depth-wise sampling rate and lower accuracy.  Our proposed DDR-Net achieves the comparable ratio on the 3rd stage compared to CasMVSNet (84.35$\%$) but has a shorter range to have optimized intervals. Note that only the final stage (the 3rd stage) determines the quality of the final output depth prediction. This justifies that our estimated uncertainty intervals are of high confidence and also have short sampling distances to produce the accurate depth map. 

\begin{table}[t]
\footnotesize
\begin{tabular*}{8.3cm}{l|ll|lll}  
\hline  
Methods &REM &Loss Strategy &Acc. &Comp. &Overall\\ 
\hline  
Baseline~\cite{cascade}  &  &  &0.346 &0.351 &0.348\\  
Model A  & \checkmark &  &0.333 &0.341 &0.337\\
Model B  & \checkmark & \checkmark &0.339 &0.320 &0.329\\ 
\hline  
\end{tabular*} 
\caption{Ablation study on REM and loss strategy.}  
\label{Tab:AblationStudy}
\vspace{-0.7cm}
\end{table} 

The first and second rows in Table~\ref{Tab:AblationStudy} shows the results with or without the range estimation module (REM) compared with the baseline model~\cite{cascade}. Notes that the range estimation module improves the reconstruction results significantly in both accuracy and completeness (Acc: from 0.346 to 0.333; Comp: from 0.351 to 0.341), showing the effectiveness of our range estimation module. 

\noindent {\bf Effectiveness of loss strategy.} To evaluate the effectiveness of our novel loss strategy, we evaluated our network with or without our loss strategy on the DTU dataset. The third and fourth rows in Table~\ref{Tab:DDRRatios} show the results with or without our loss strategy. Comparing to the model only with range estimation module (REM), the model with additional loss strategy can reduce the range of the third stage from 20.91mm to 19.24mm in average (7.9$\%$ shorter). On the contrary, the ratio of covering the ground truth depth improves from 0.8391 to 0.8435. This demonstrates that our novel loss strategy can not only increase the confidence of the estimated depth hypotheses, but also increase the depth sampling rates to obtain depth maps with higher accuracy.
Also, as shown in Table~\ref{Tab:AblationStudy}, the model with our novel loss strategy, which is able to estimate high confidence depth ranges, improves both completeness and overall score significantly (Comp: from 0.341 to 0.320; Overall: from 0.337 to 0.329). These results show our novel loss strategy can further improve the reconstruction results.

\vspace{-1.0em}
\section{Conclusion}
\label{Sec:Conclusion}
\vspace{-0.5em}

In this paper, we present a DDR-Net which enables to dynamically estimate the depth range hypotheses. Comparing with previous methods which determine a depth range of each pixel identically, our proposed range estimation module (REM) can estimate ranges from global areas and obtain dynamic depth hypotheses with shorter running time and less memory consumption. Moreover, we proposed a novel loss strategy to refine REM and boost the reconstruction results. Experimental results show that our DDR-Net improves accuracy and completeness of point cloud reconstruction on several standard datasets and achieves state-of-the-art performance on the DTU benchmark.


{\small

\bibliographystyle{ieee_fullname}
\bibliography{egbib}
}
\clearpage

\section{Appendix}

\subsection{Overview}

In this supplementary details, we first evaluated the effectiveness of our proposed range estimation module (REM) with additional experiments in Section~\ref{Sec:REMAdditionalExperiments}. In Section~\ref{Sec:LossStrategyAdditionalExperiments}, more quantitative and qualitative evaluations on our proposed novel loss strategy are provided. In Section~\ref{Sec:PointCloudReconstruction}, more final point cloud reconstruction results of the DTU dataset~\cite{dtu} are presented. Finally, some limitations and future work are discussed in Section~\ref{Sec:FutureWork}.

\subsection{Additional Experiments on Range Estimation Module (REM)}
\label{Sec:REMAdditionalExperiments}

In this section, we introduce the full network architecture of our proposed range estimation module (REM)
and compare our proposed DDR-Net with other state-of-the-art methods to demonstrate the effectiveness of our module. All following experiments were conducted on the DTU testing dataset. To have a fair comparison, we set the input image resolution to $1152\times864$ and utilize original testing parameters of each method to complete following experiments.

As shown in Table~\ref{Tab:REM-NetArchitecture}, our proposed REM is composed of several simple 2D CNNs and a sigmoid activation function. The sigmoid function restricts the values of uncertainty map from 0 to 1. Note that the REM is light-weighted and can be applied to any multi-view stereo framework in a coarse-to-fine manner. Also, our REM is quite effective for the depth range hypotheses estimation since its strong ability to gather probability information from global areas.

\begin{figure*}
    \centering
    \subfigure[Scan 77 view 3]{
    \begin{minipage}[t]{0.33\linewidth}
    \centering
    \includegraphics[width=1.0\textwidth]{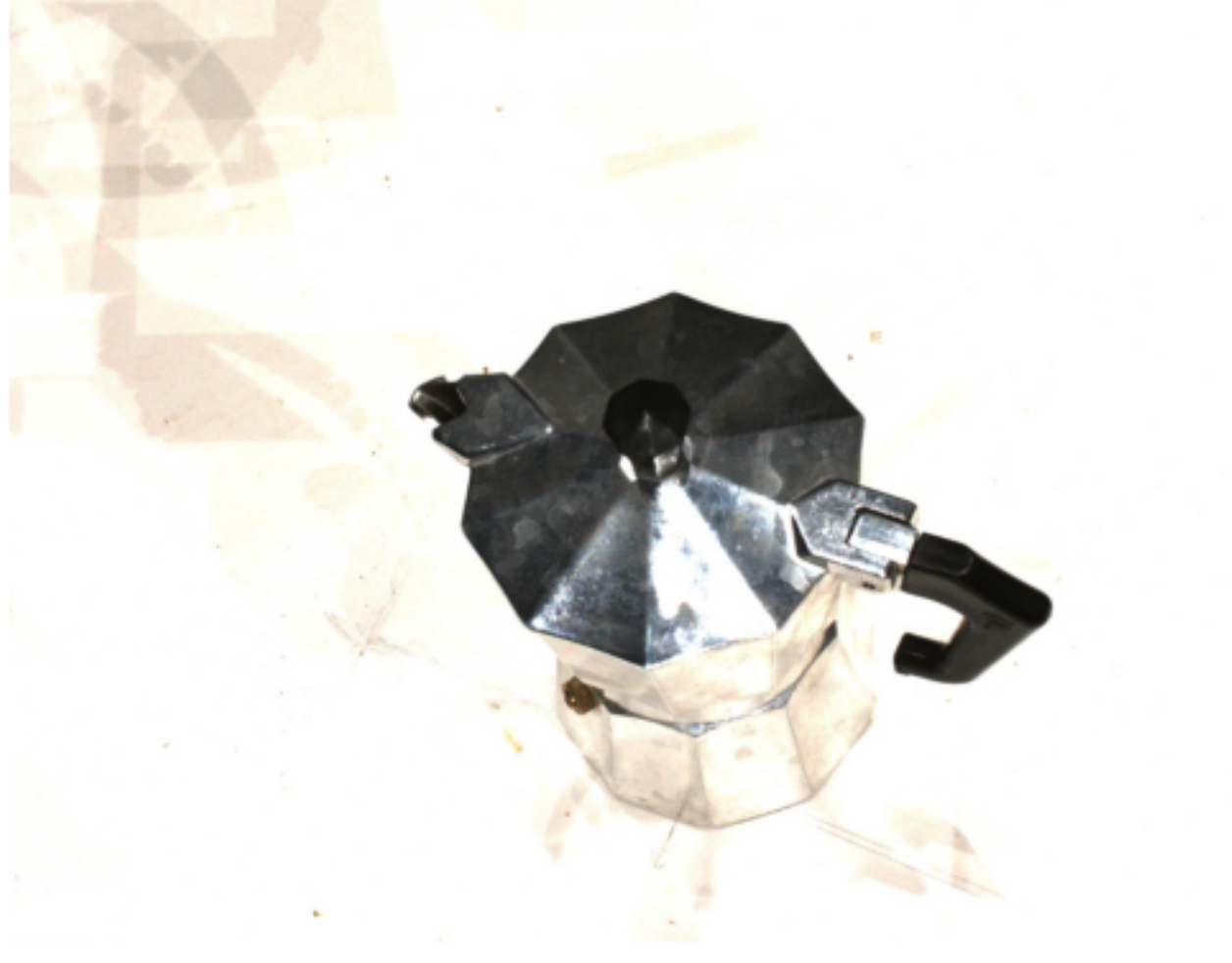}
    \end{minipage}%
    }%
    \subfigure[Depth prediction]{
    \begin{minipage}[t]{0.33\linewidth}
    \centering
    \includegraphics[width=1.0\textwidth]{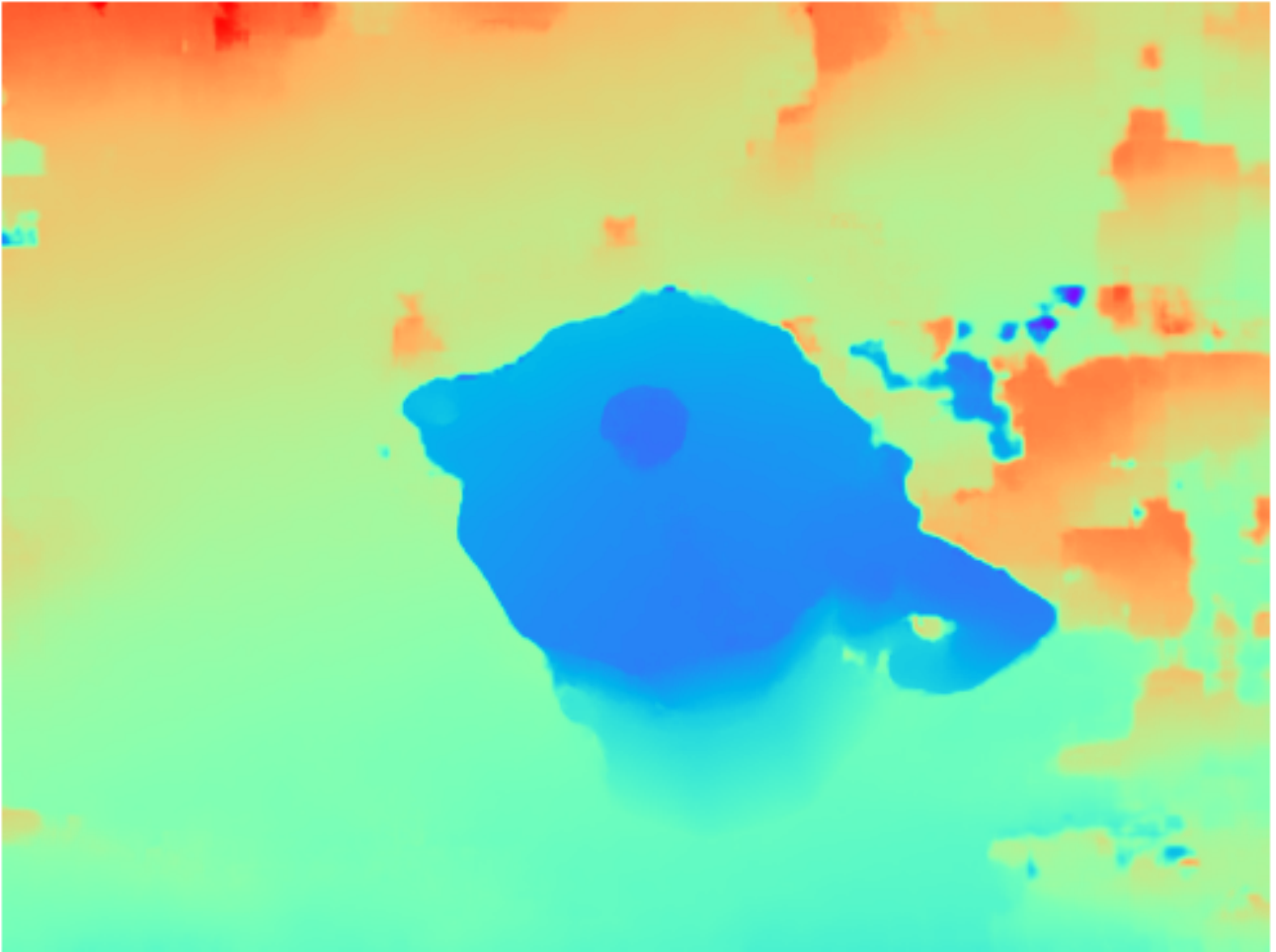}
    \end{minipage}%
    }%
    \subfigure[Ground Truth]{
    \begin{minipage}[t]{0.33\linewidth}
    \centering
    \includegraphics[width=1.0\textwidth]{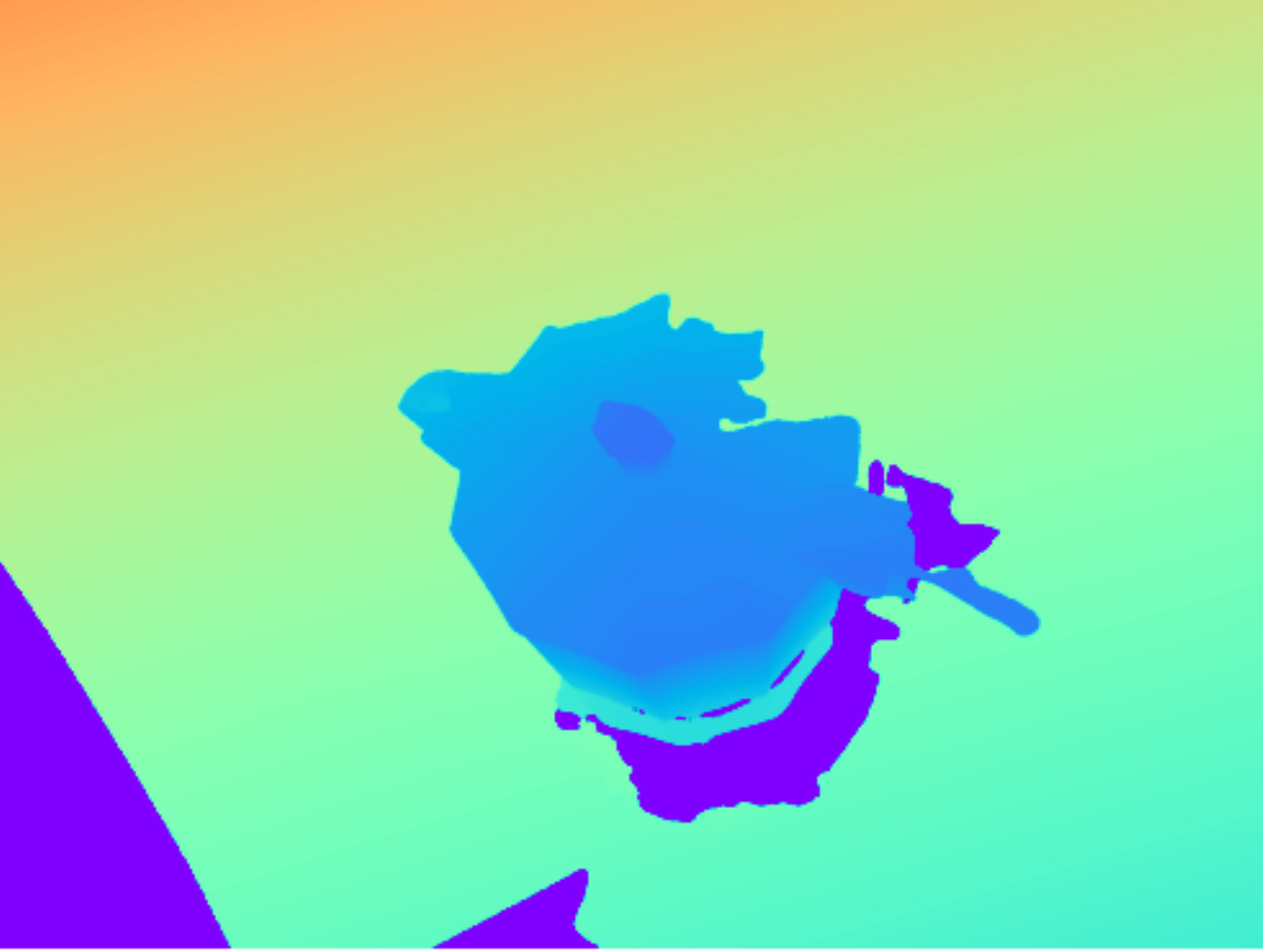}
    \end{minipage}%
    }%
    
    \subfigure[Scan 114 view 8]{
    \begin{minipage}[t]{0.33\linewidth}
    \centering
    \includegraphics[width=1.0\textwidth]{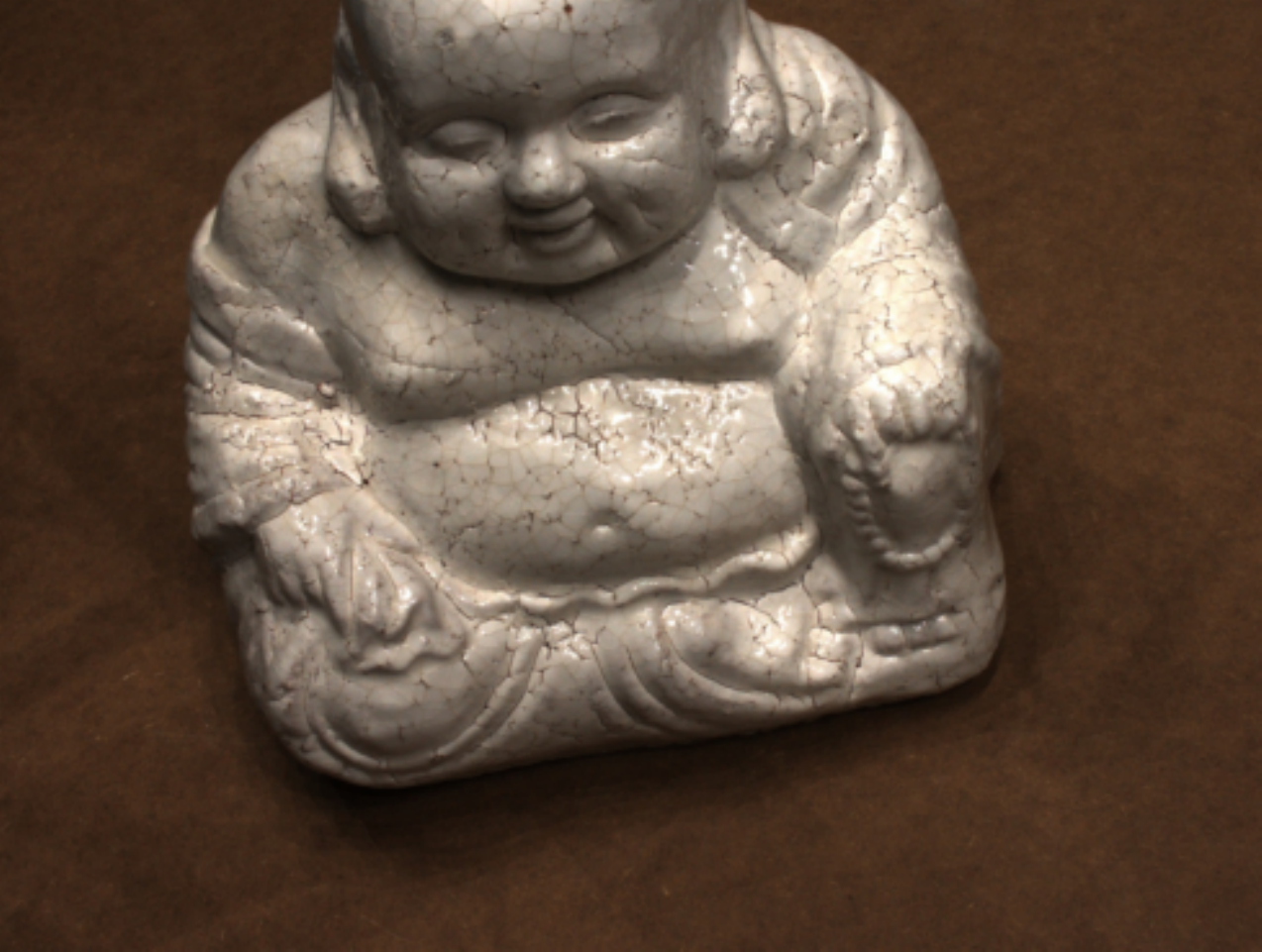}
    \end{minipage}%
    }%
    \subfigure[Depth prediction]{
    \begin{minipage}[t]{0.33\linewidth}
    \centering
    \includegraphics[width=1.0\textwidth]{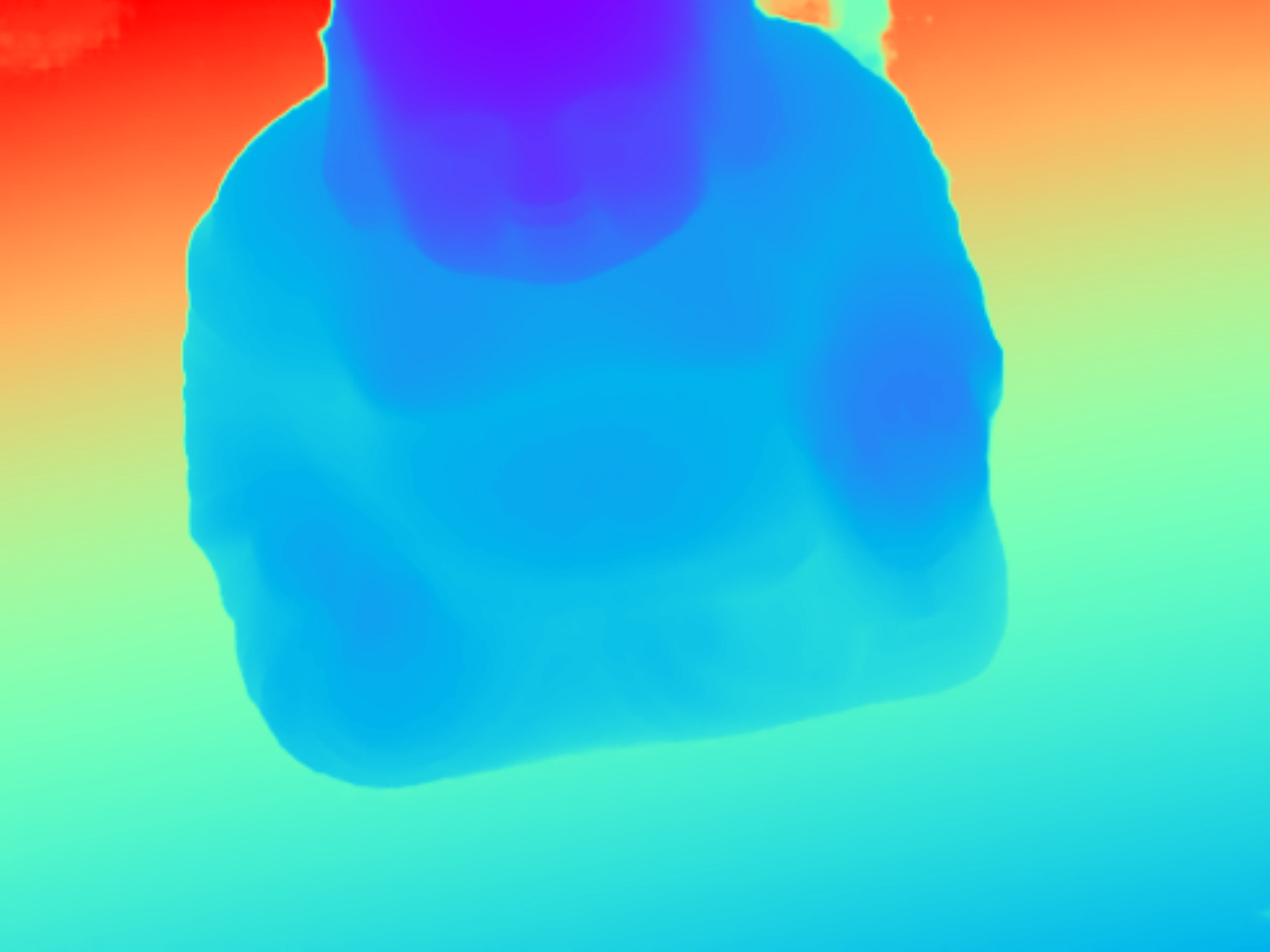}
    \end{minipage}%
    }%
    \subfigure[Ground Truth]{
    \begin{minipage}[t]{0.33\linewidth}
    \centering
    \includegraphics[width=1.0\textwidth]{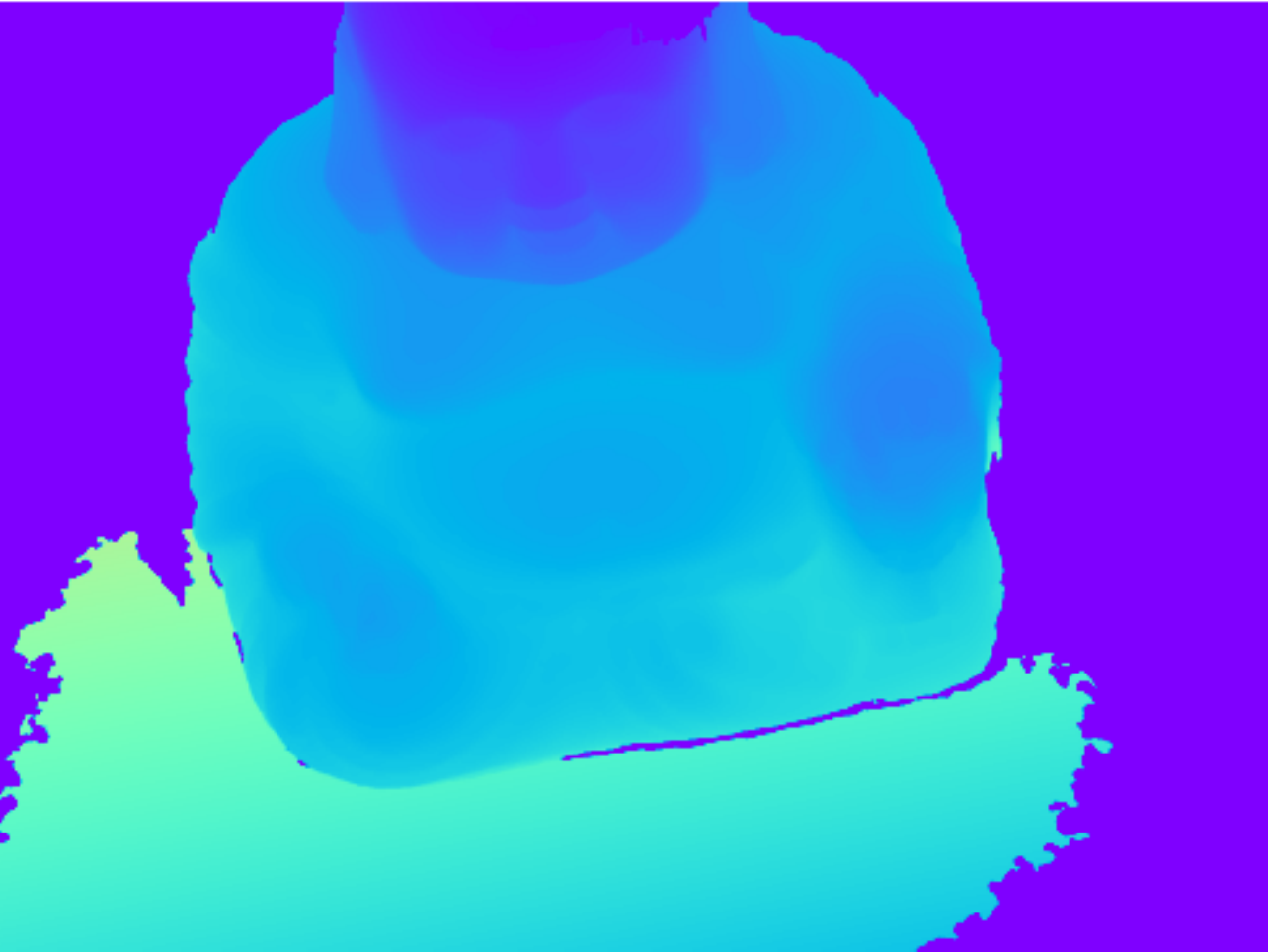}
    \end{minipage}%
    }%
    
    \subfigure[All scans]{
    \begin{minipage}[t]{0.33\linewidth}
    \centering
    \includegraphics[width=1.0\textwidth]{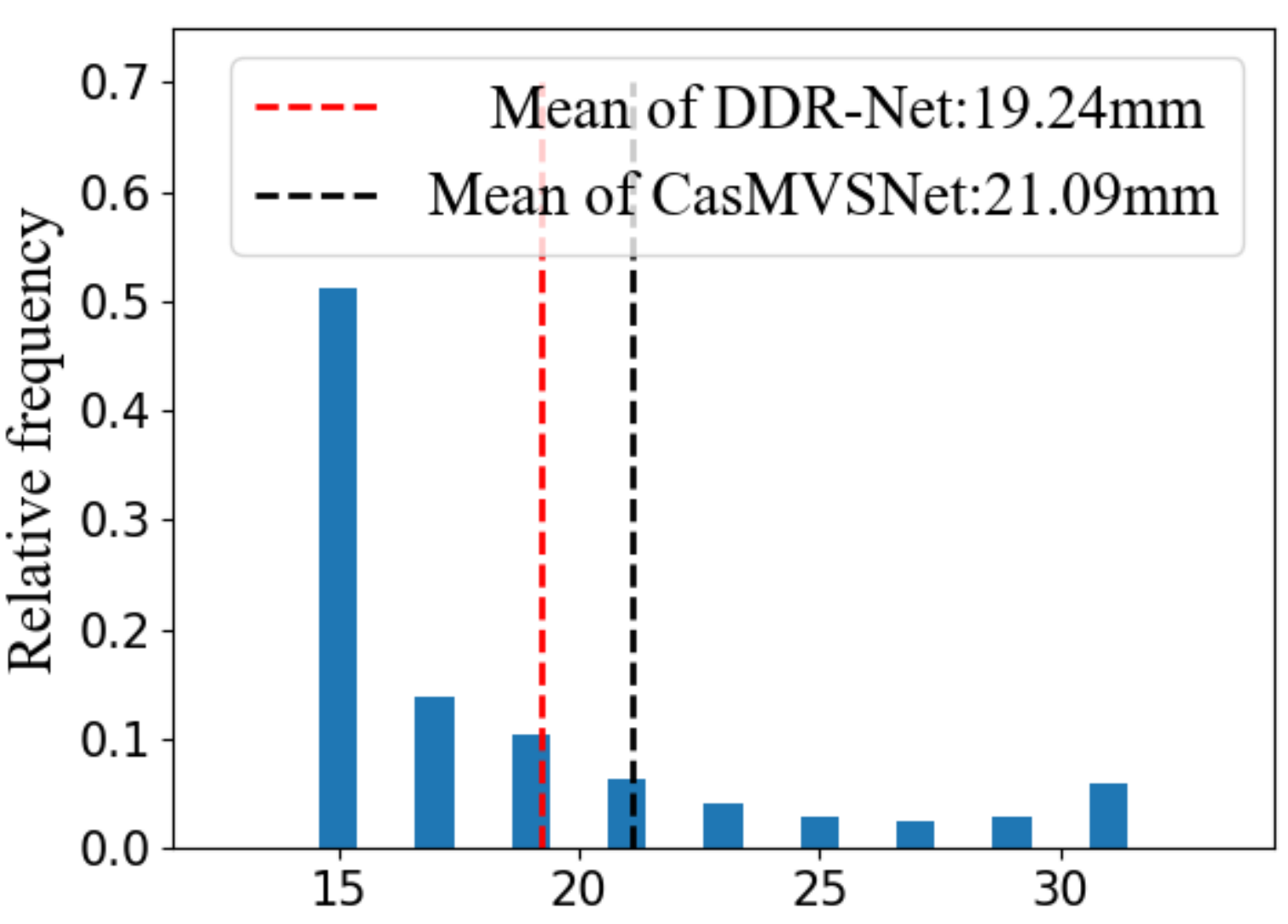}
    \end{minipage}%
    }%
    \subfigure[Scan 77]{
    \begin{minipage}[t]{0.33\linewidth}
    \centering
    \includegraphics[width=1.0\textwidth]{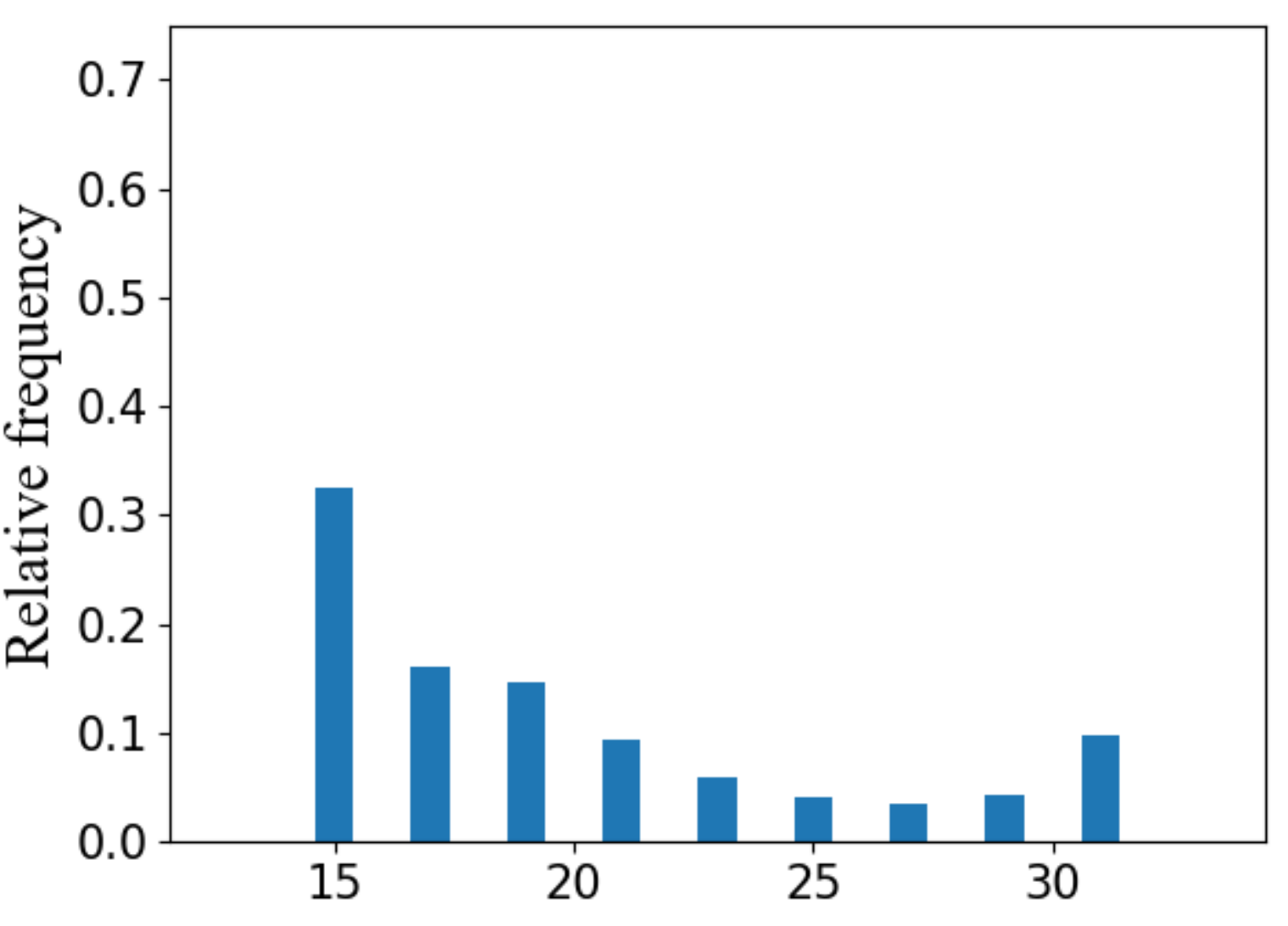}
    \end{minipage}%
    }%
    \subfigure[Scan 114]{
    \begin{minipage}[t]{0.33\linewidth}
    \centering
    \includegraphics[width=1.0\textwidth]{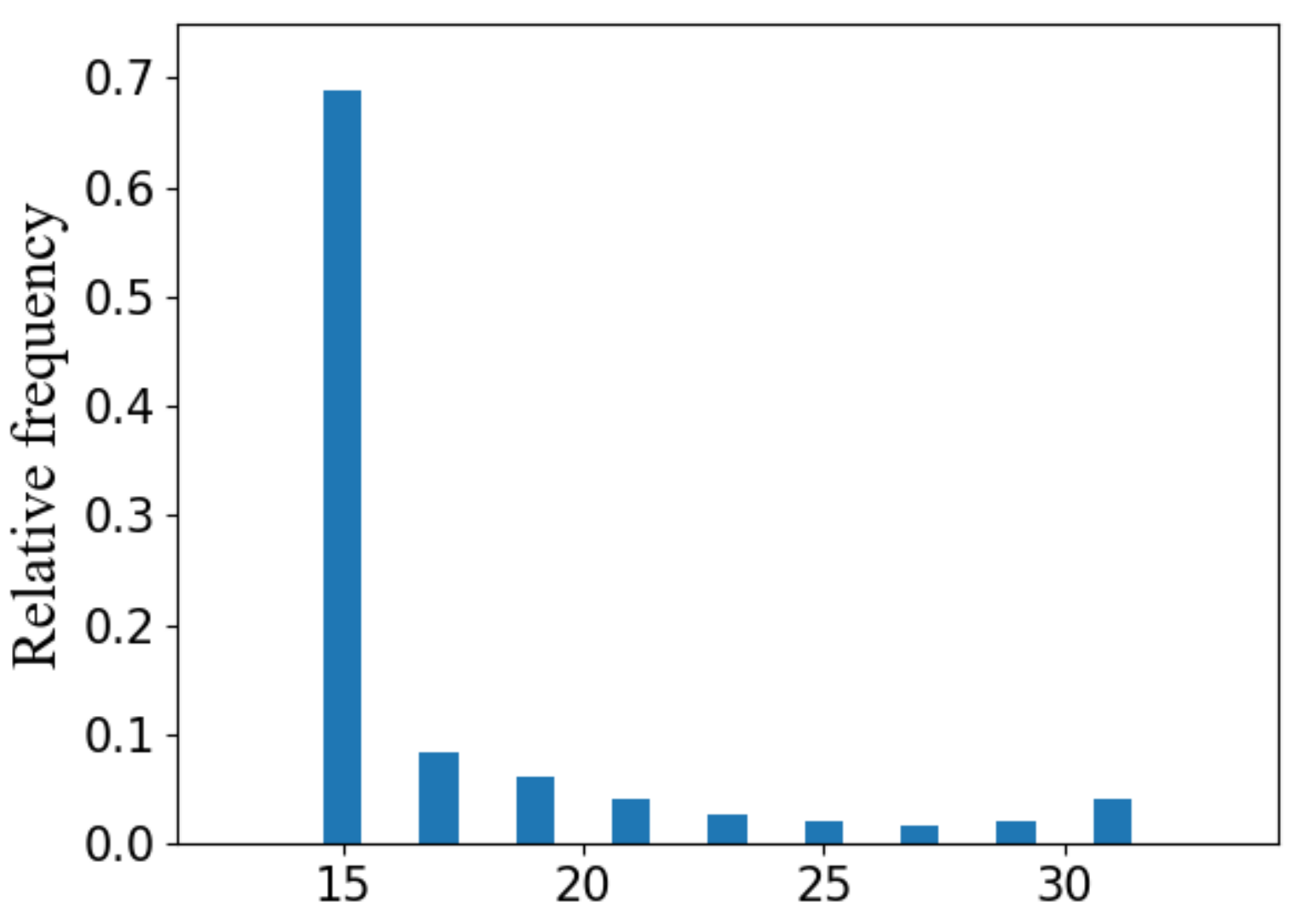}
    \end{minipage}%
    }%
    \caption{Histograms of the dynamic depth ranges on the 3rd stage in our proposed DDR-Net of all scans, scan 77 and scan 114 in the DTU testing dataset. The $x$-axis denotes the interval length (mm).  We mark the mean values of the lengths of CasMVSNet~\cite{cascade} (black dash line) and our DDR-Net (red dash line). On the first and second rows, we show RGB image crops, our depth predictions and ground truth depth maps of scan 77 and scan 114. }
    \label{Fig:DepthRange}
\end{figure*}

We now present the histograms of the estimated dynamic depth ranges of the 3rd stage of all scans, scan 77 and scan 114 in Figure~\ref{Fig:DepthRange}.  We also mark the mean lengths of our DDR-Net and CasMVSNet~\cite{cascade} in the first histograms. The reference RGB images, our depth predictions and ground truth depth maps of scan 114 and scan 77 are also provided. Note that the distributions of depth ranges of all scans meet our expectations: nearly half of all the depth ranges are of low uncertainties and some portions of the ranges are considered as high uncertainties. That is reflected by the fact that only boundary of each object, occluded areas and texture-less regions require longer depth range to cover their ground truth. Then we show two specific scans to illustrate the adaptation ability of our REM. Scan 77 is a scan of a white cup and scan 114 is about a monk sculpture. It is observed that scan 77 consists of large texture-less regions and suffers from optical reflection while the monk sculpture has rich textures.  From the results, we can observe that more pixels are considered as high uncertainties in scan 77 and more pixels are respected as high confidences in scan 114. This verifies that our range estimation module (REM) is able to adapt to different conditions even in texture-less areas and regions that are suffered from optical reflection.

\definecolor{orange}{RGB}{0,191,255}
\begin{table}[t]
\small\setlength\tabcolsep{4pt}
\centering
\begin{tabular}{c|c|c|c|c}  
\hline  
Layer &Stride &Kernel &Channel &Input \\ 
\hline  
conv0 & $1\times1$ & $3\times3$ & $48/32\rightarrow16$ &probability volume\\
\hline
conv1  & $1\times1$ & $3\times3$ & $16\rightarrow32$	&  conv0\\
\hline
conv2  & $1\times1$ & $3\times3$ & $32\rightarrow32$	&  conv1\\
\hline
conv3  & $1\times1$ & $3\times3$ & $32\rightarrow16$	&  conv2\\
\hline
conv4  & $1\times1$ & $3\times3$ & $16\rightarrow1$	&  conv3\\
\hline
\cellcolor{orange}sigmoid & — &  — &  — &  conv4\\
\hline
\end{tabular} 
\vspace{1em}
\caption{The detail architecture of our proposed range estimation module (REM). Each convolution layer consists of a 2D convolution layer, a batch normalization layer and a ReLU layer. The sigmoid layer (the colored cell) is utilized to limit the output values of the uncertainty map to the range [0, 1]. }  
\label{Tab:REM-NetArchitecture}
\end{table} 

\begin{figure*}[htbp]
\begin{center}

\subfigure[RGB image]{
\begin{minipage}[t]{0.245\linewidth}
\centering
\includegraphics[height=1.1in,width=1.65in]{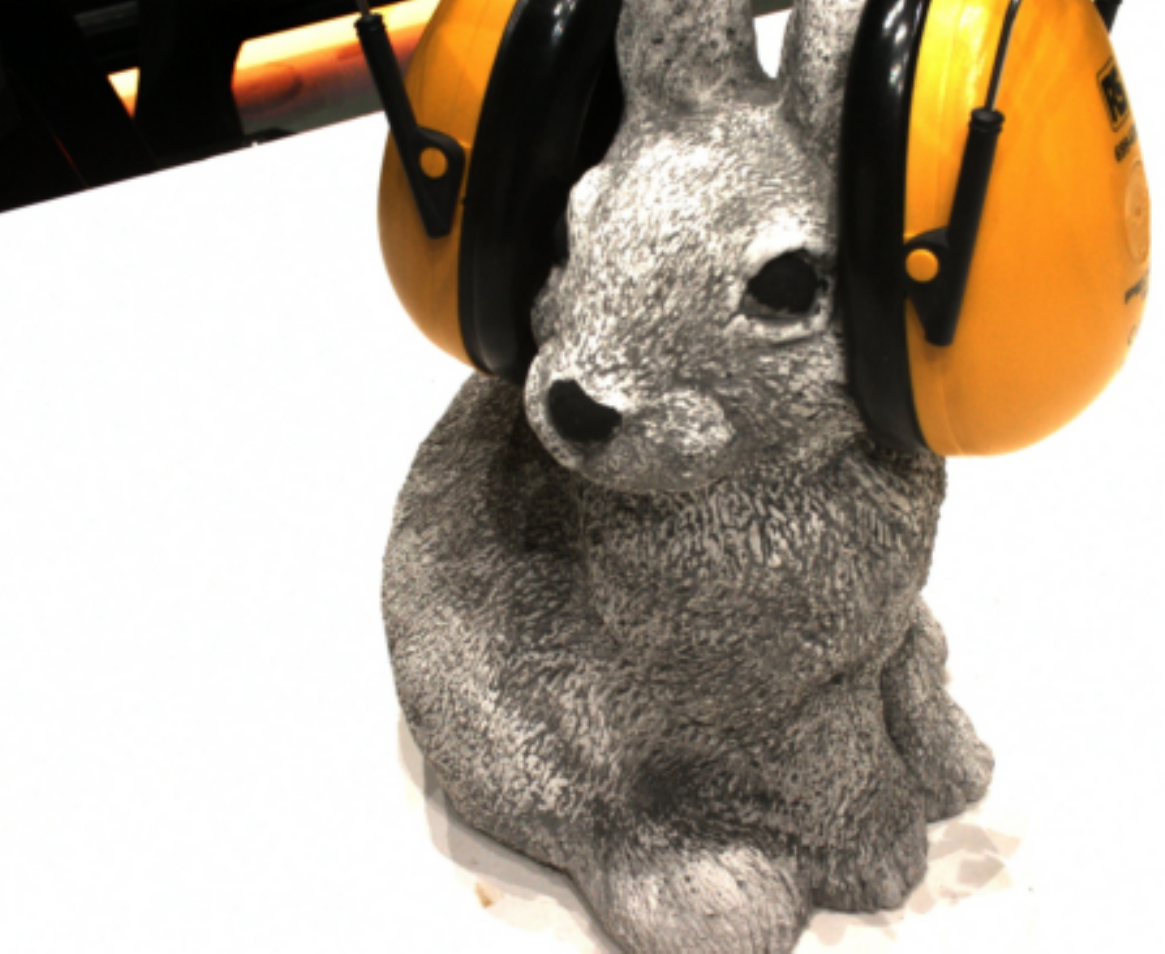}
\end{minipage}%
}%
\subfigure[Ground-truth depth map]{
\begin{minipage}[t]{0.245\linewidth}
\centering
\includegraphics[height=1.1in,width=1.65in]{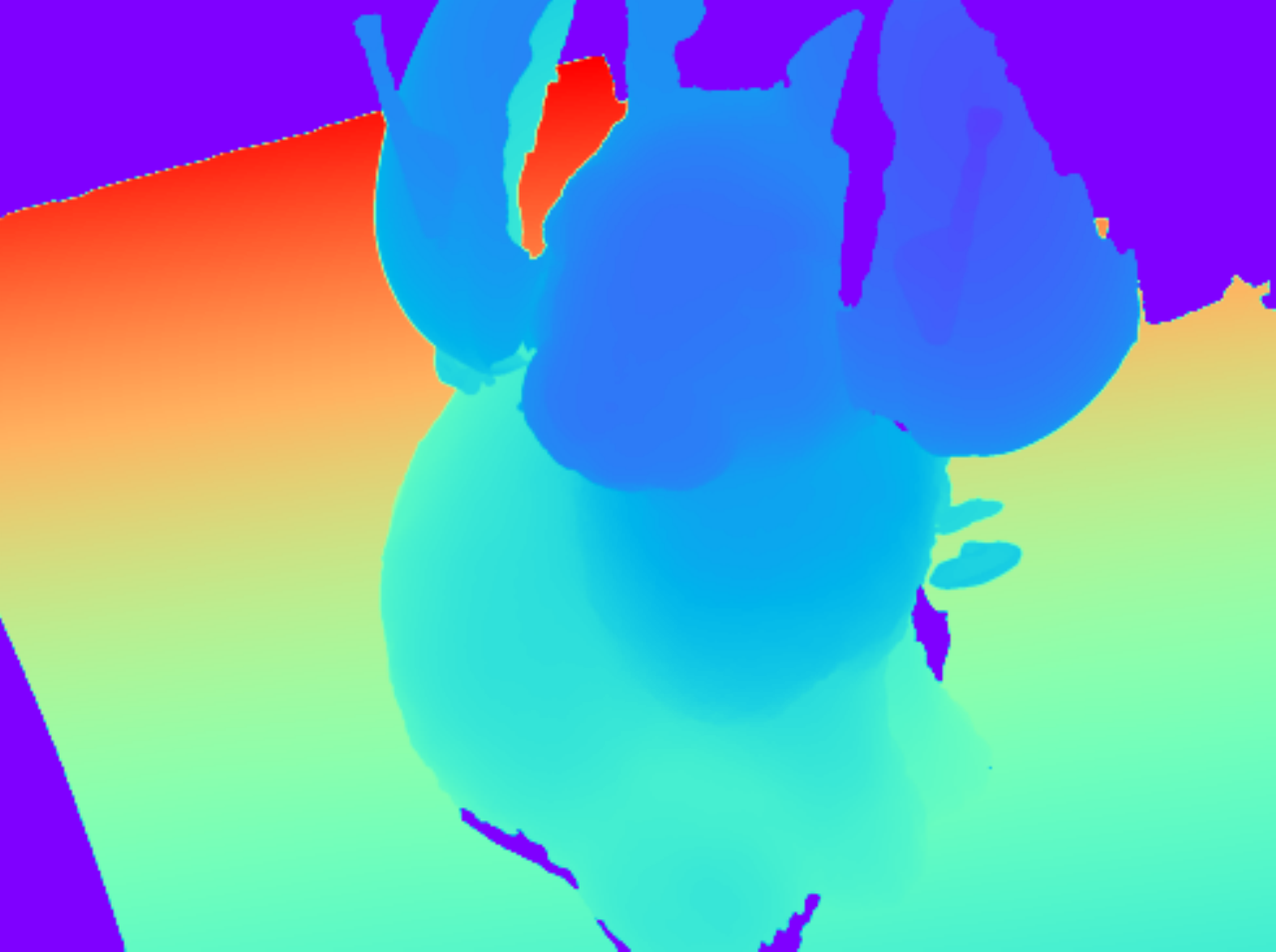}
\end{minipage}%
}%
\subfigure[RGB image]{
\begin{minipage}[t]{0.245\linewidth}
\centering
\includegraphics[height=1.1in,width=1.65in]{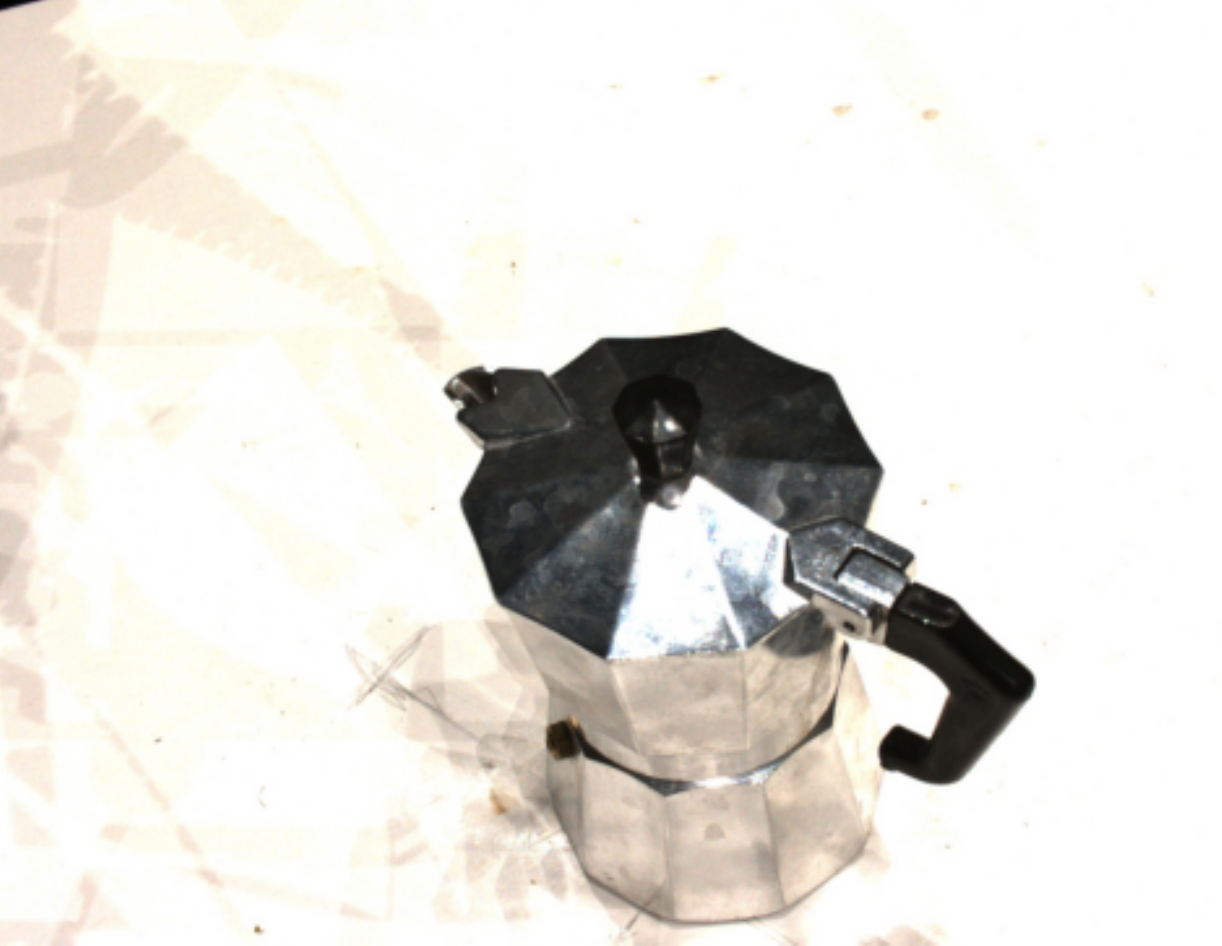}
\end{minipage}%
}%
\subfigure[Ground-truth depth map]{
\begin{minipage}[t]{0.245\linewidth}
\centering
\includegraphics[height=1.1in,width=1.65in]{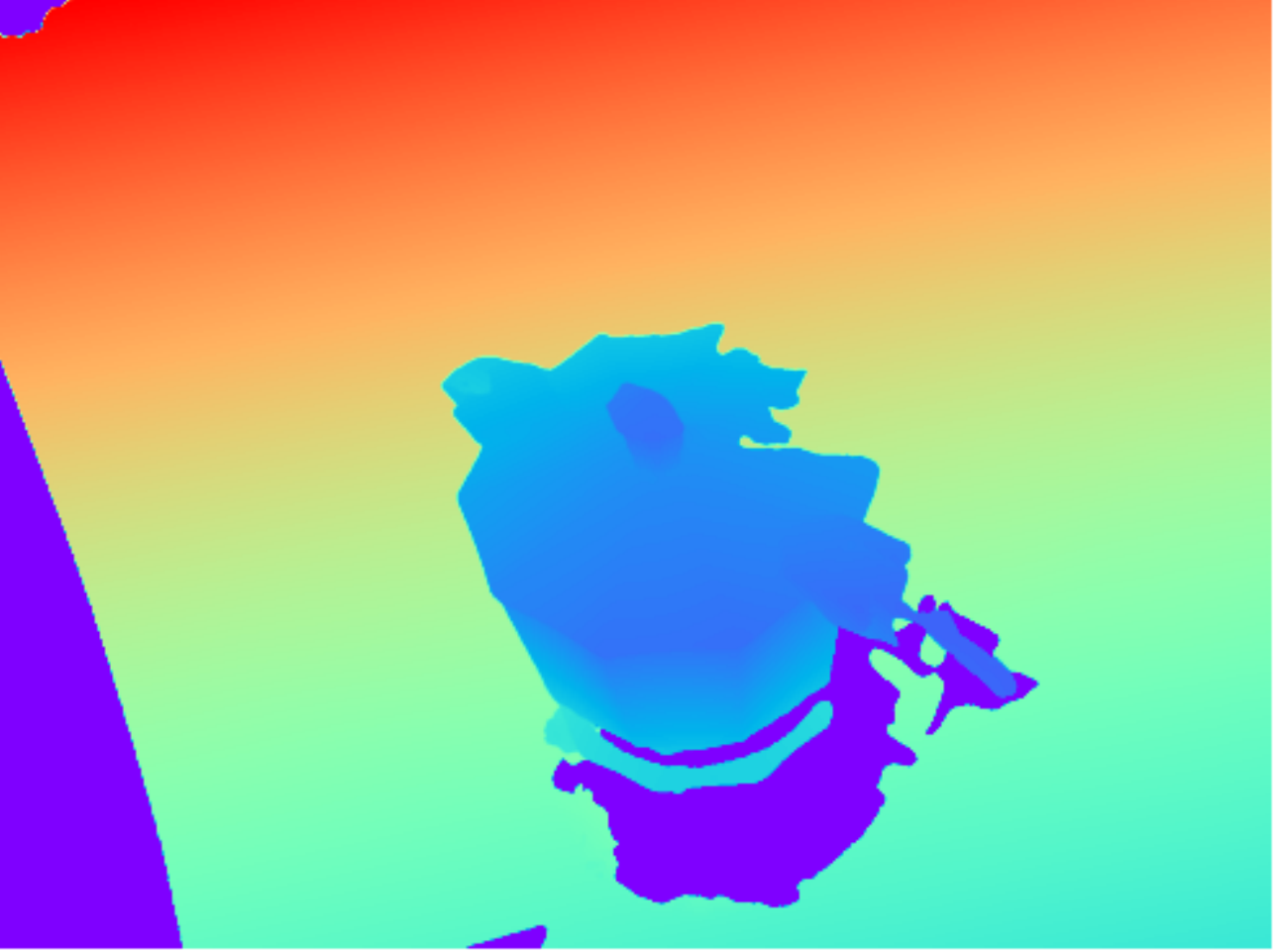}
\end{minipage}%
}%

\subfigure[UCSNet: depth map]{
\begin{minipage}[t]{0.245\linewidth}
\centering
\includegraphics[height=1.1in,width=1.65in]{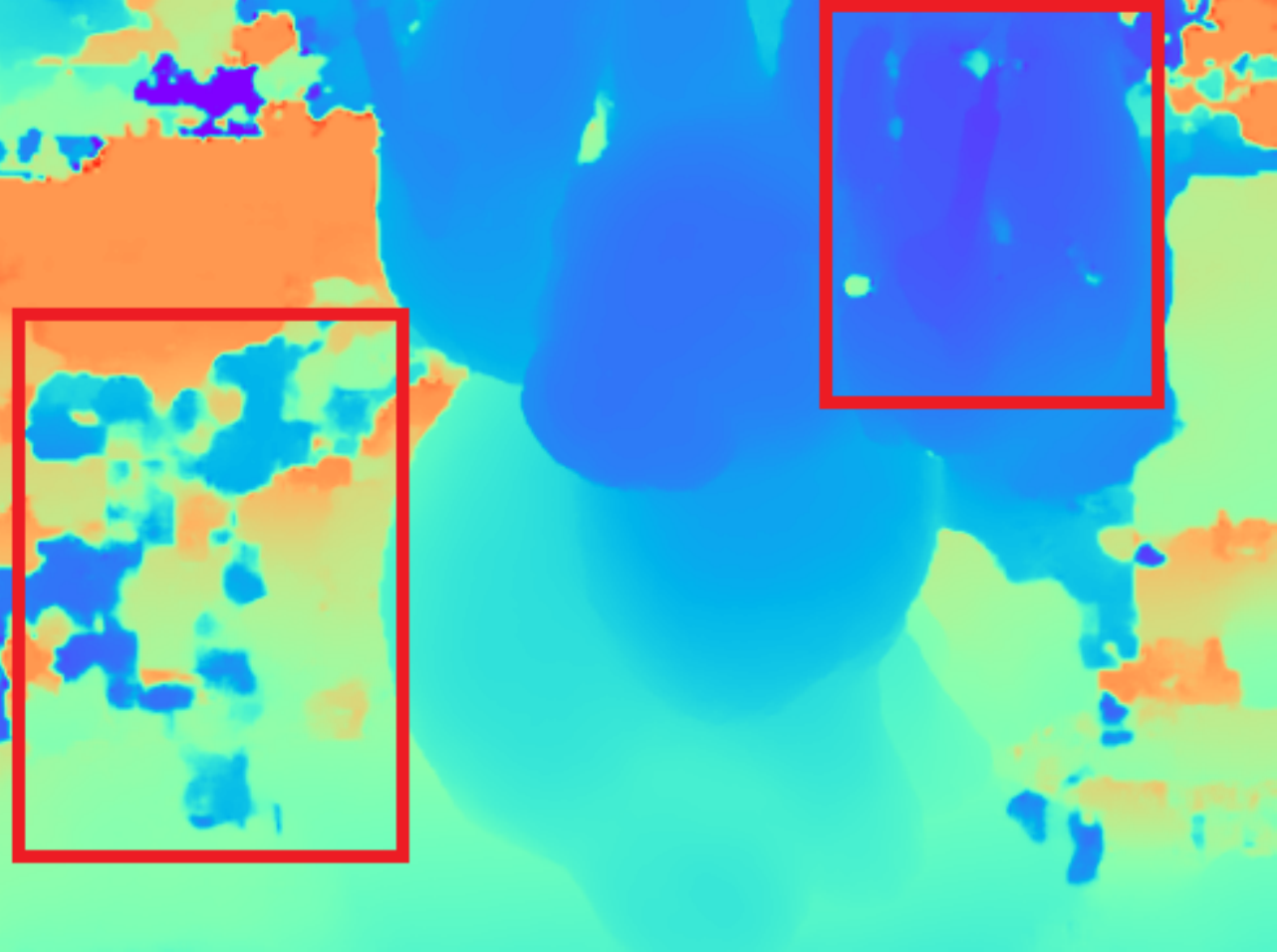}
\end{minipage}%
}%
\subfigure[DDR-Net: depth map]{
\begin{minipage}[t]{0.245\linewidth}
\centering
\includegraphics[height=1.1in,width=1.65in]{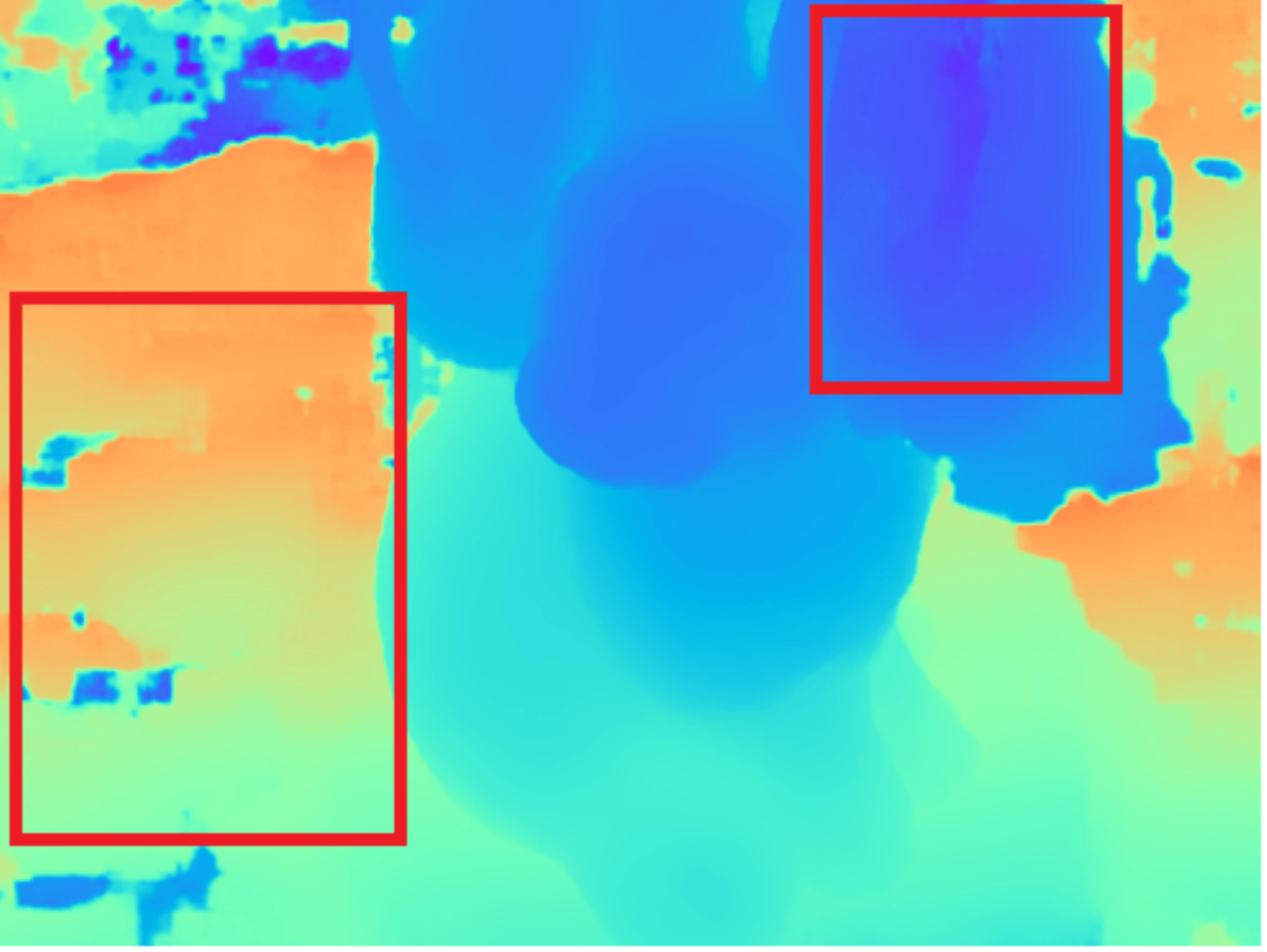}
\end{minipage}%
}%
\subfigure[UCSNet: depth map]{
\begin{minipage}[t]{0.245\linewidth}
\centering
\includegraphics[height=1.1in,width=1.65in]{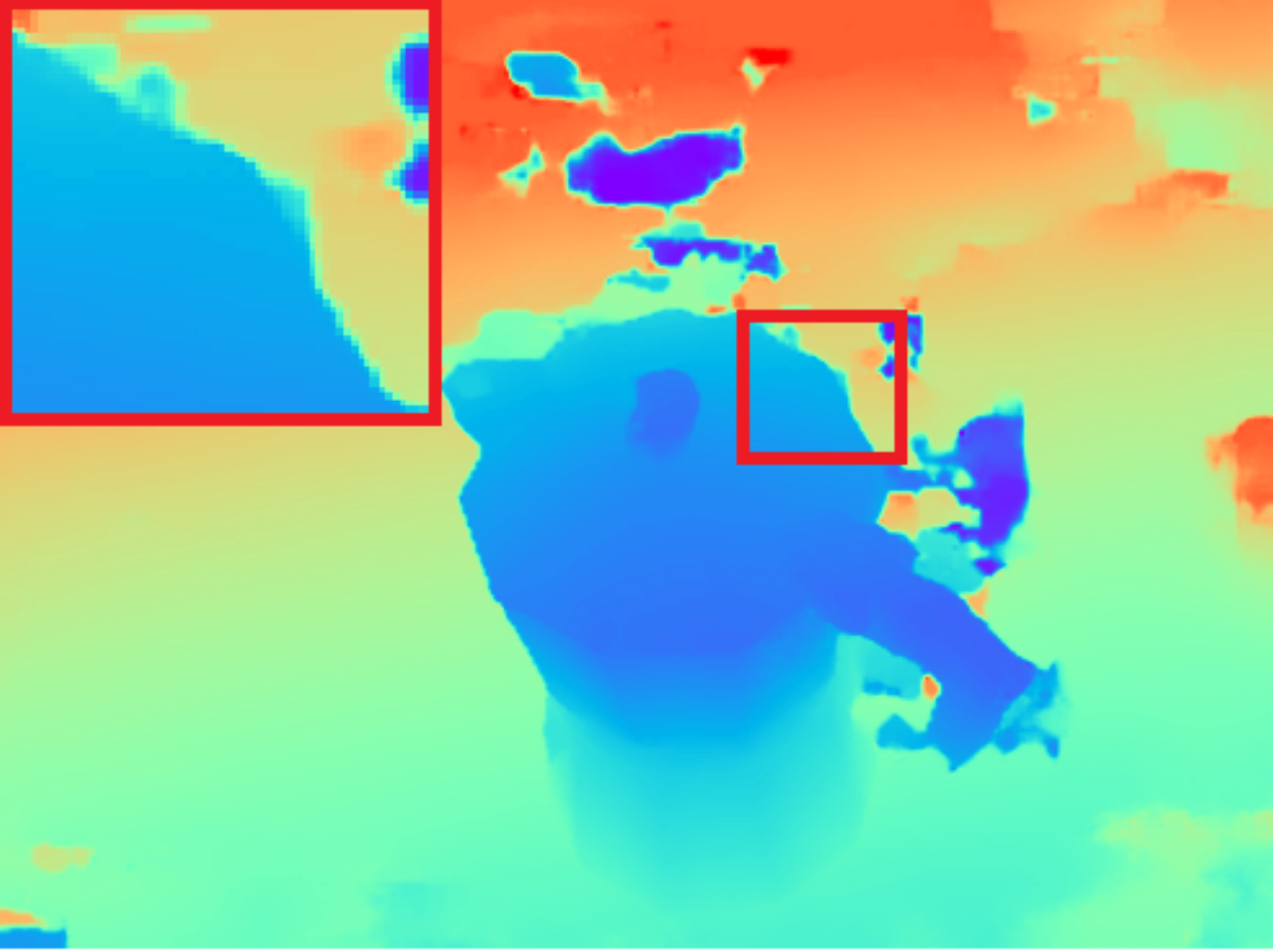}
\end{minipage}%
}%
\subfigure[DDR-Net: depth map]{
\begin{minipage}[t]{0.245\linewidth}
\centering
\includegraphics[height=1.1in,width=1.65in]{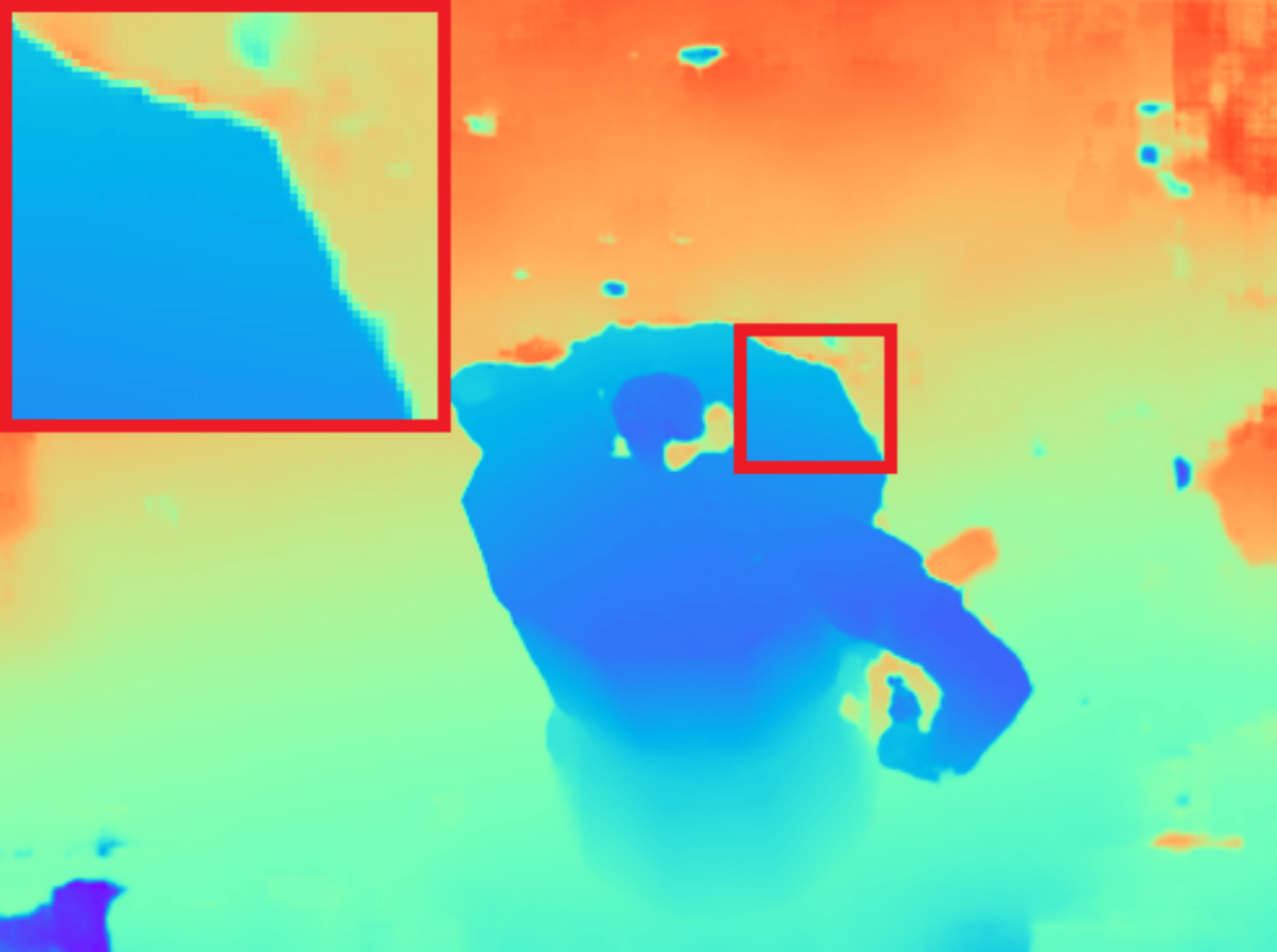}

\end{minipage}%
}%

\subfigure[Error map]{
\begin{minipage}[t]{0.245\linewidth}
\centering
\includegraphics[height=1.1in,width=1.65in]{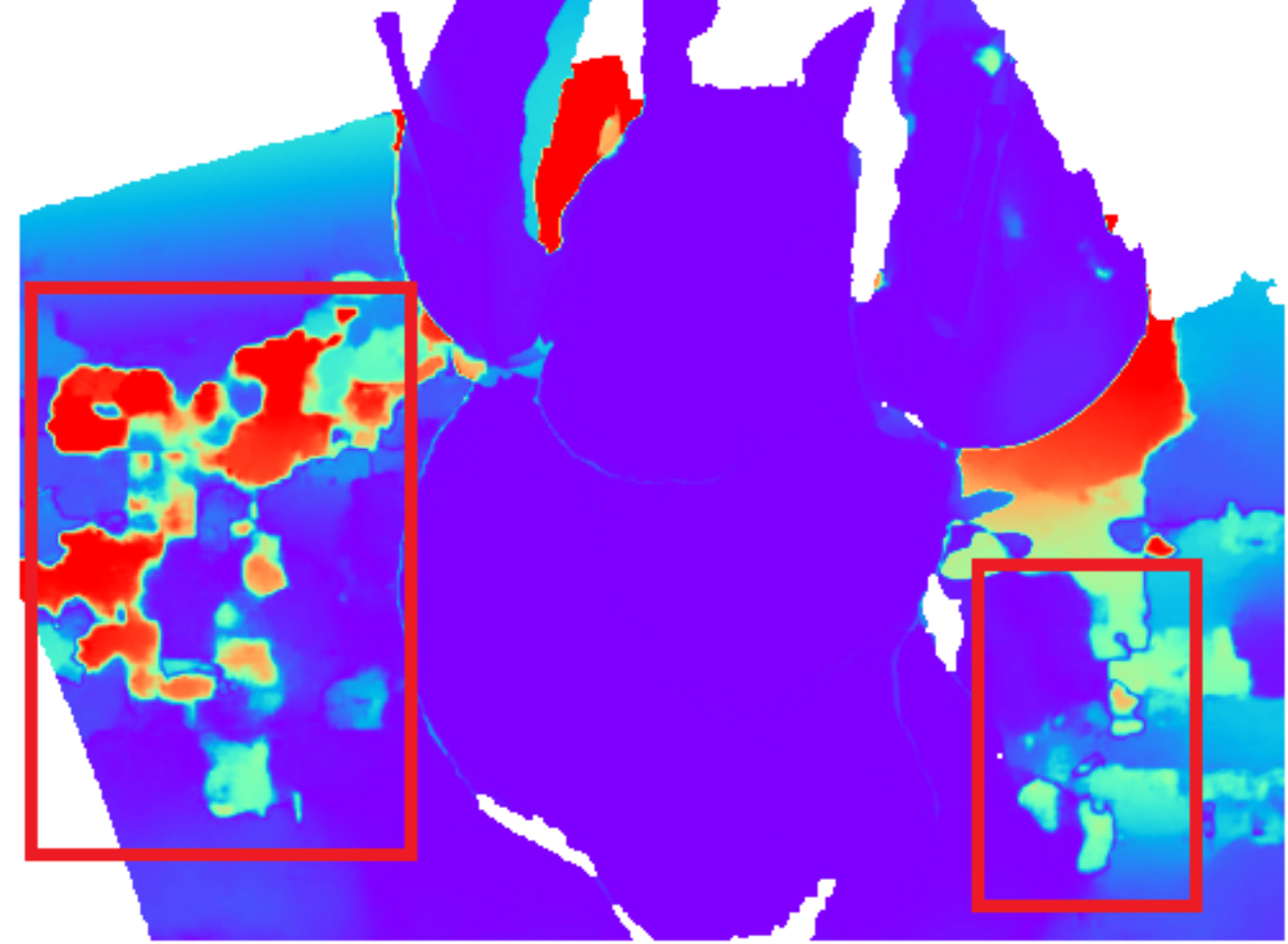}
\end{minipage}%
}%
\subfigure[Error map]{
\begin{minipage}[t]{0.245\linewidth}
\centering
\includegraphics[height=1.1in,width=1.65in]{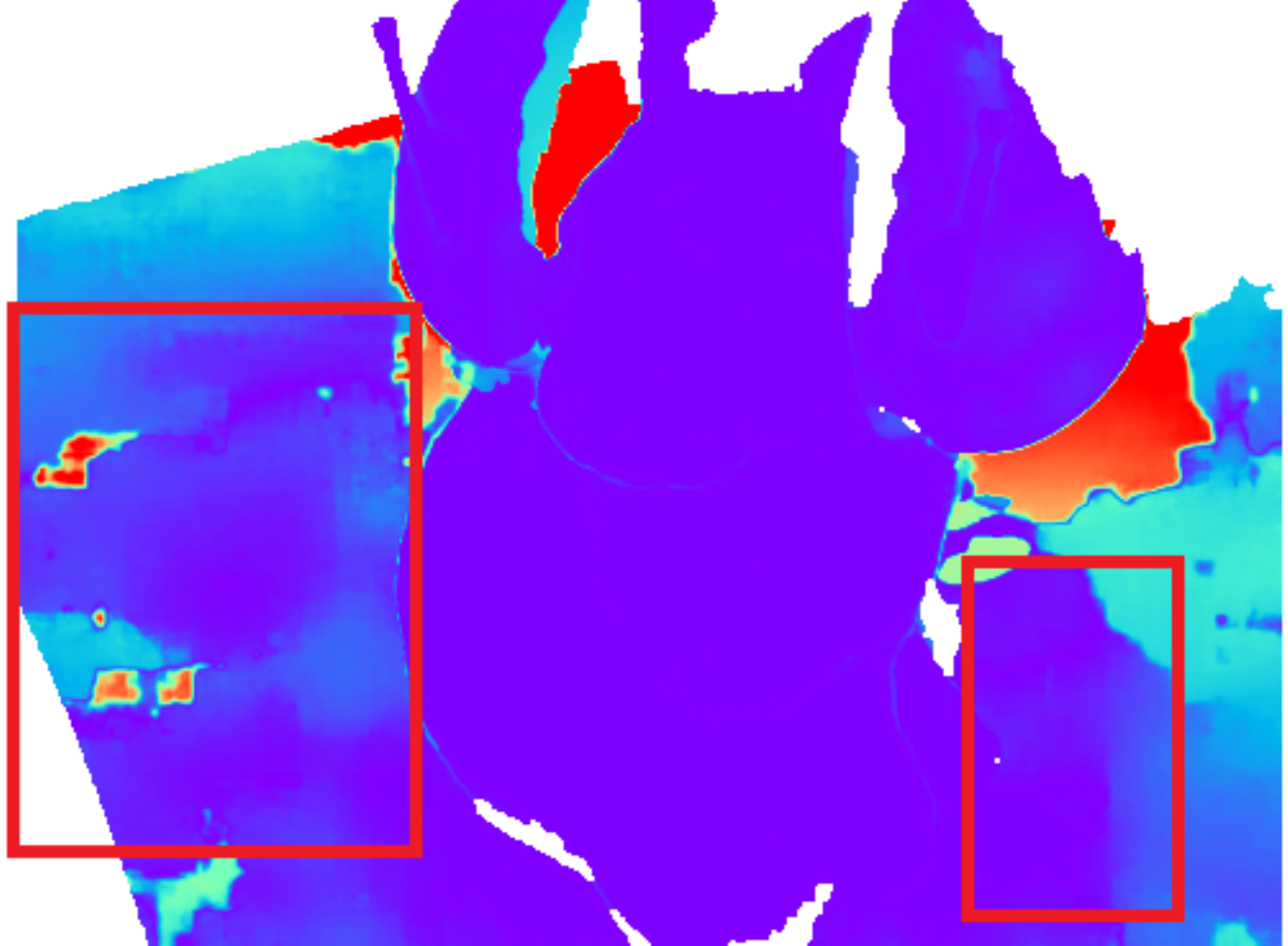}
\end{minipage}%
}%
\subfigure[Error map]{
\begin{minipage}[t]{0.245\linewidth}
\centering
\includegraphics[height=1.1in,width=1.65in]{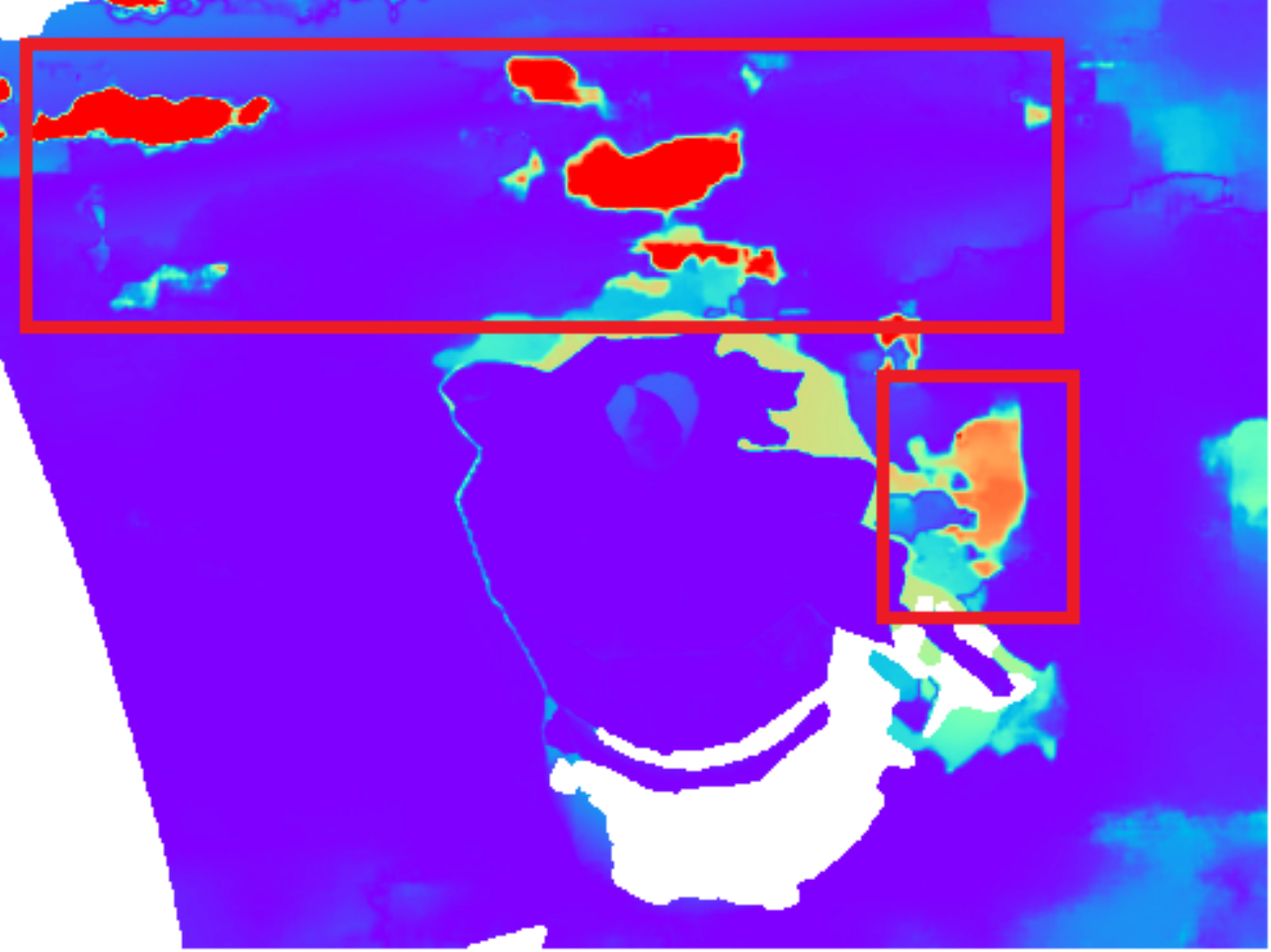}
\end{minipage}%
}%
\subfigure[Error map]{
\begin{minipage}[t]{0.245\linewidth}
\centering
\includegraphics[height=1.1in,width=1.65in]{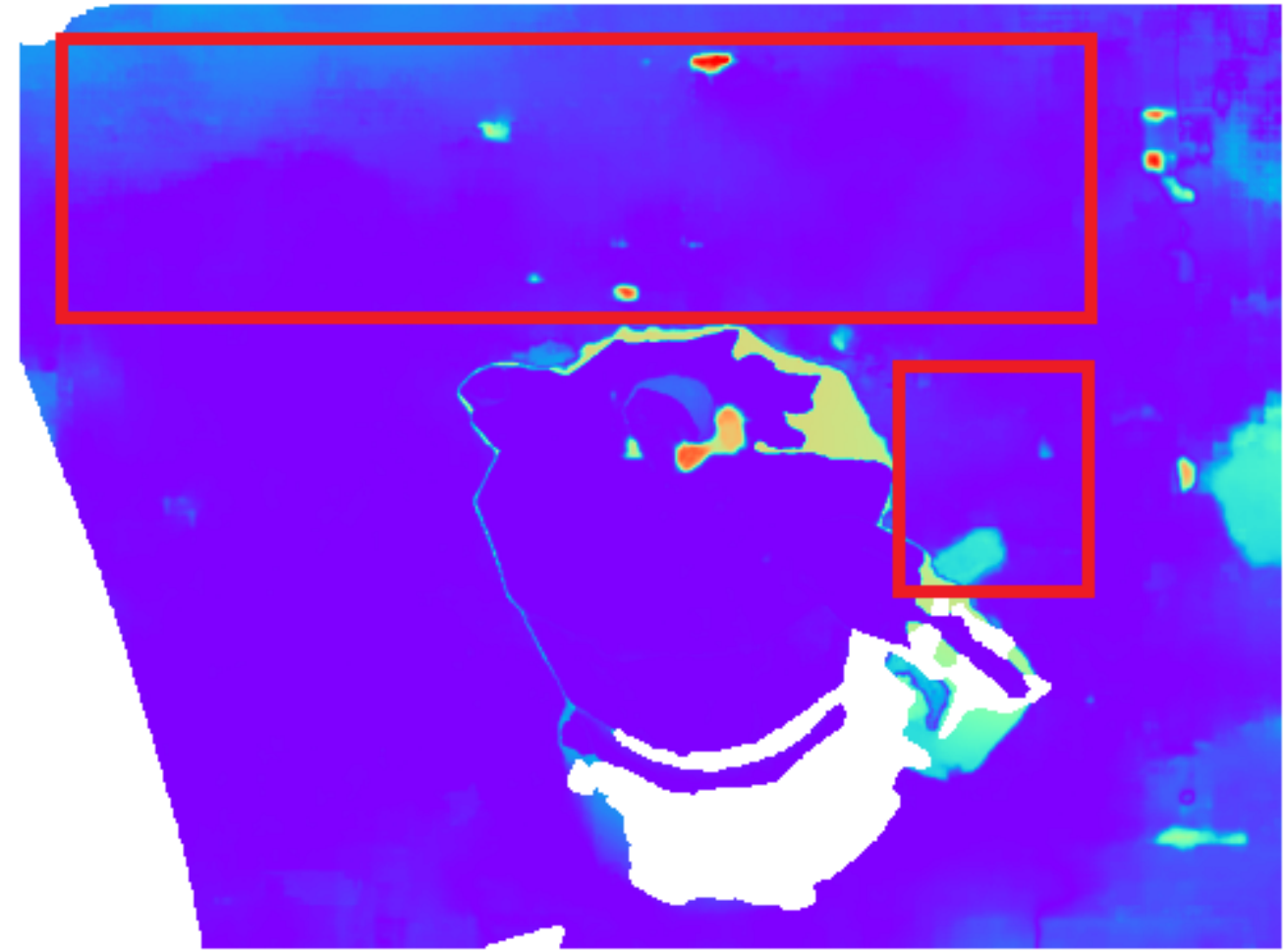}
\end{minipage}%
}%

\subfigure[Variance map of stage 1]{
\begin{minipage}[t]{0.245\linewidth}
\centering
\includegraphics[height=1.1in,width=1.65in]{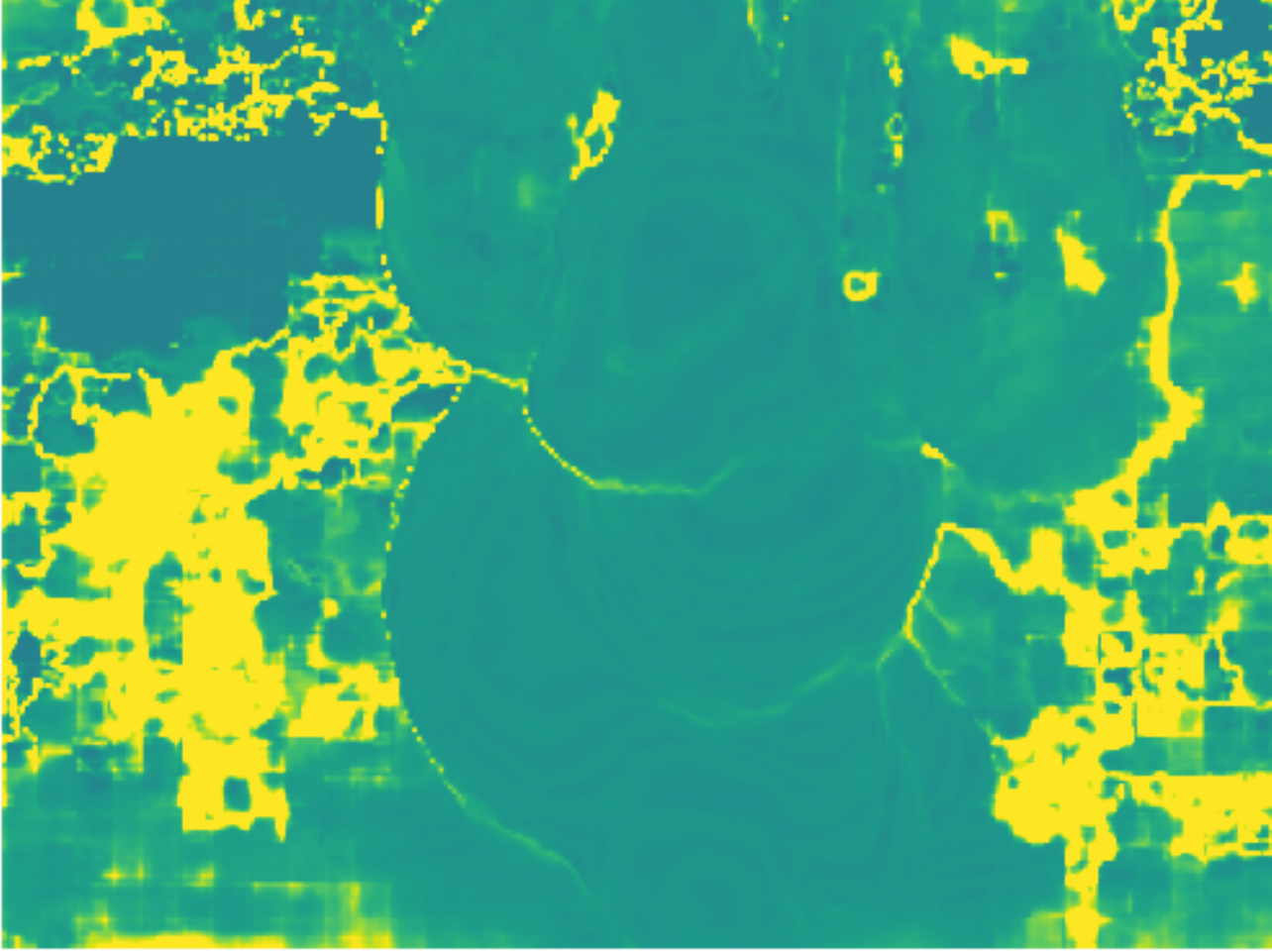}
\end{minipage}%
}%
\subfigure[Uncertainty map of stage 1]{
\begin{minipage}[t]{0.245\linewidth}
\centering
\includegraphics[height=1.1in,width=1.65in]{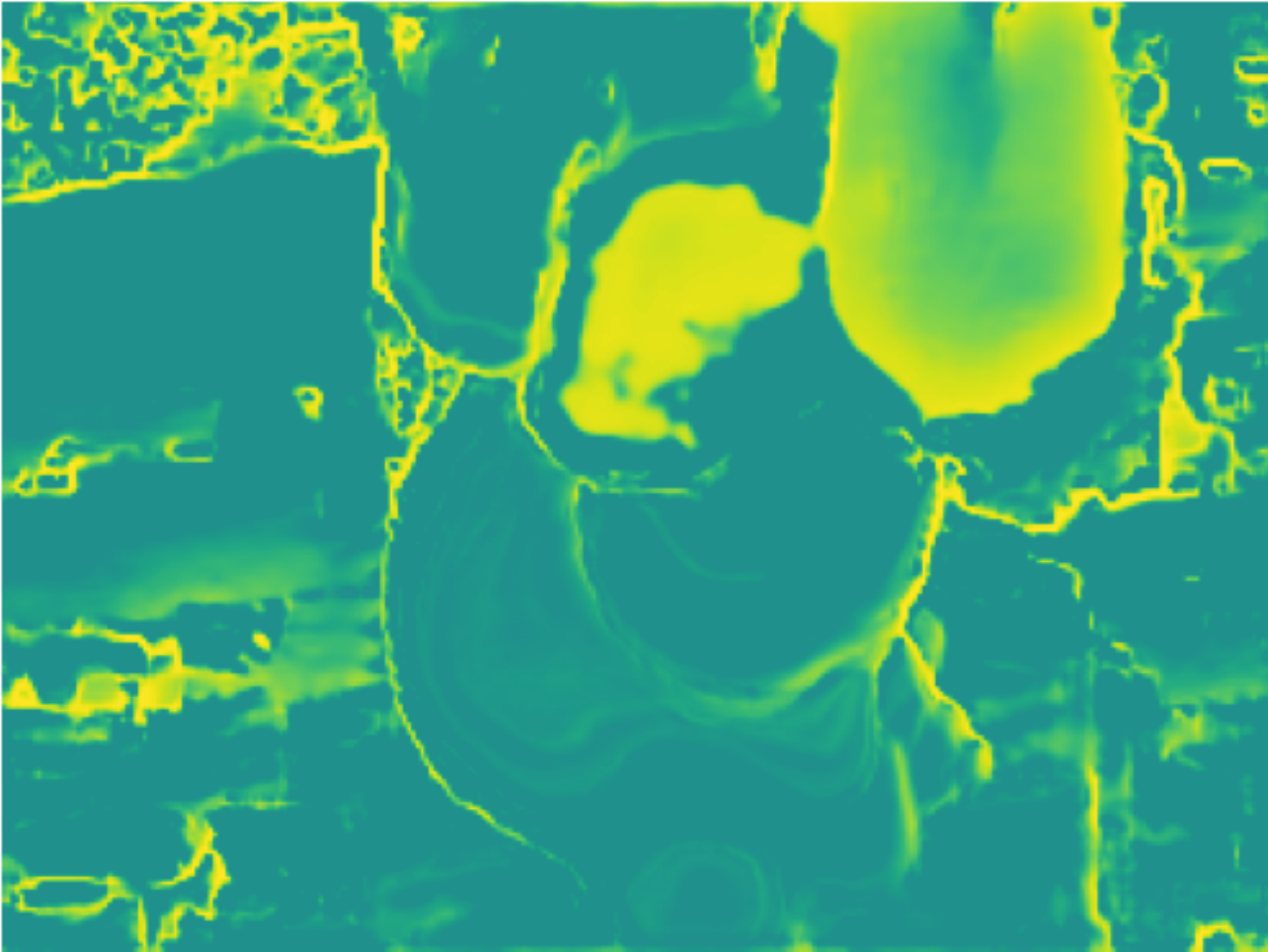}
\end{minipage}%
}%
\subfigure[Variance map of stage 1]{
\begin{minipage}[t]{0.245\linewidth}
\centering
\includegraphics[height=1.1in,width=1.65in]{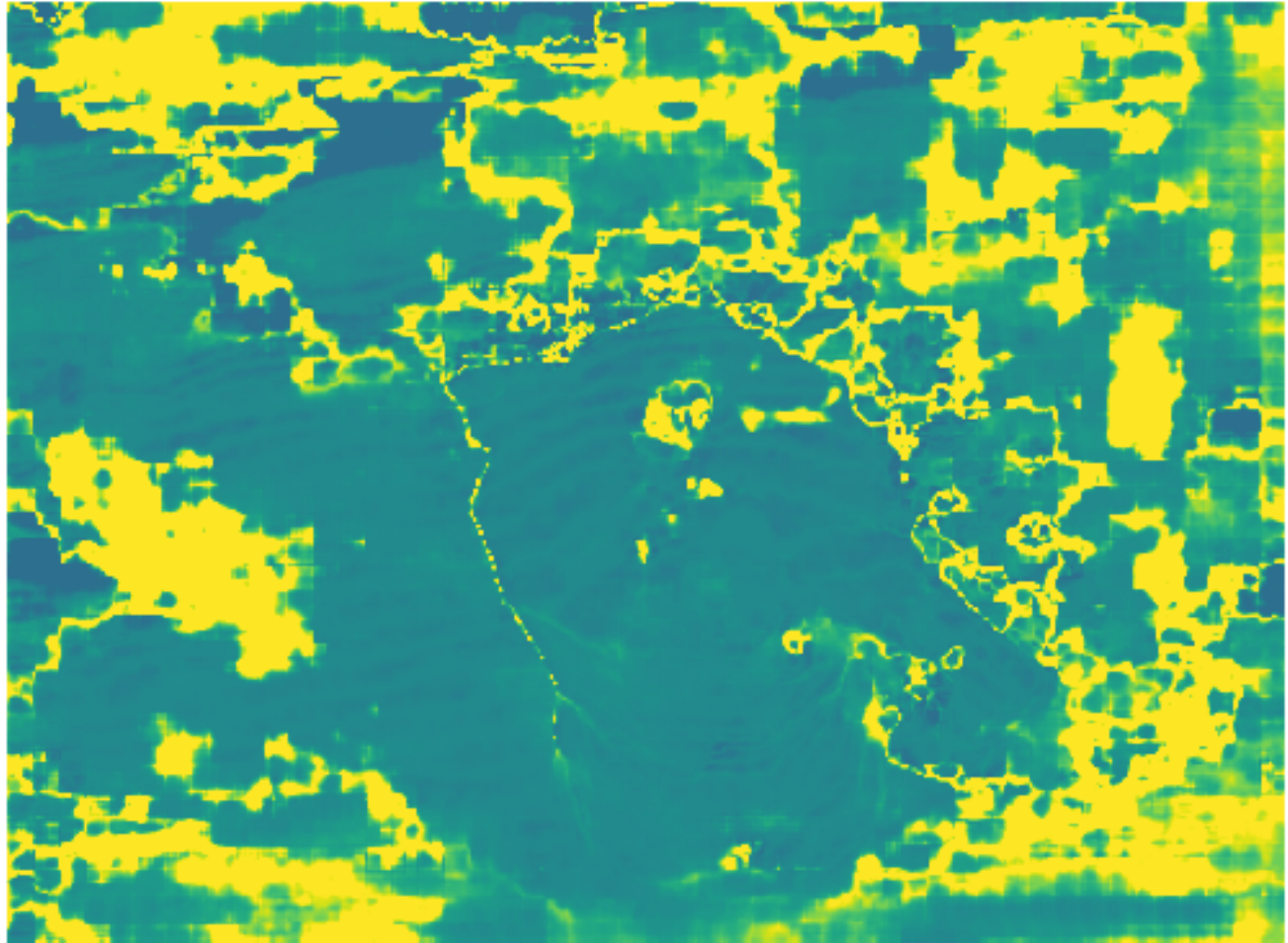}
\end{minipage}%
}%
\subfigure[Uncertainty map of stage 1]{
\begin{minipage}[t]{0.245\linewidth}
\centering
\includegraphics[height=1.1in,width=1.65in]{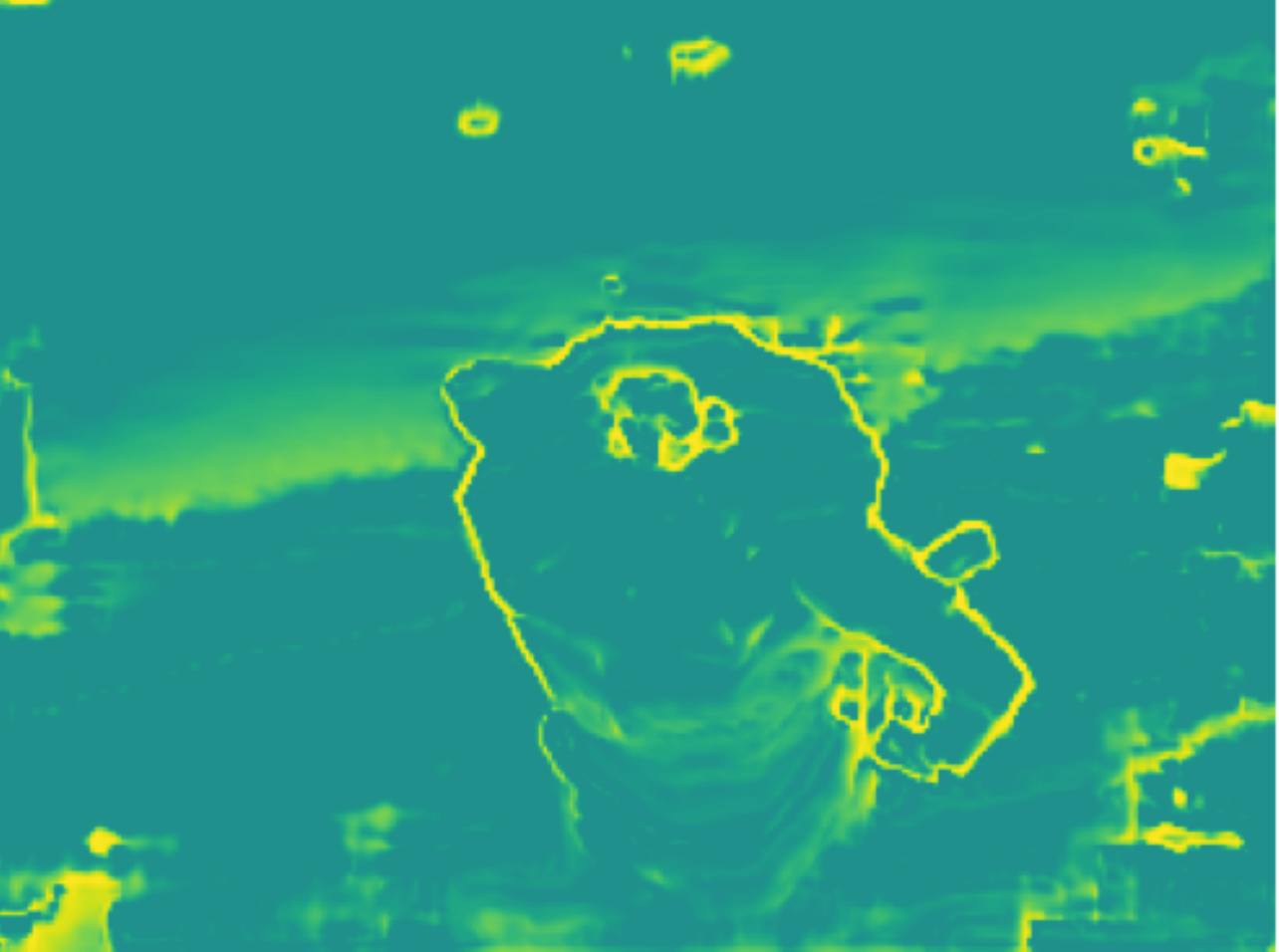}
\end{minipage}%
}%

\subfigure[Variance map of stage 2]{
\begin{minipage}[t]{0.245\linewidth}
\centering
\includegraphics[height=1.1in,width=1.65in]{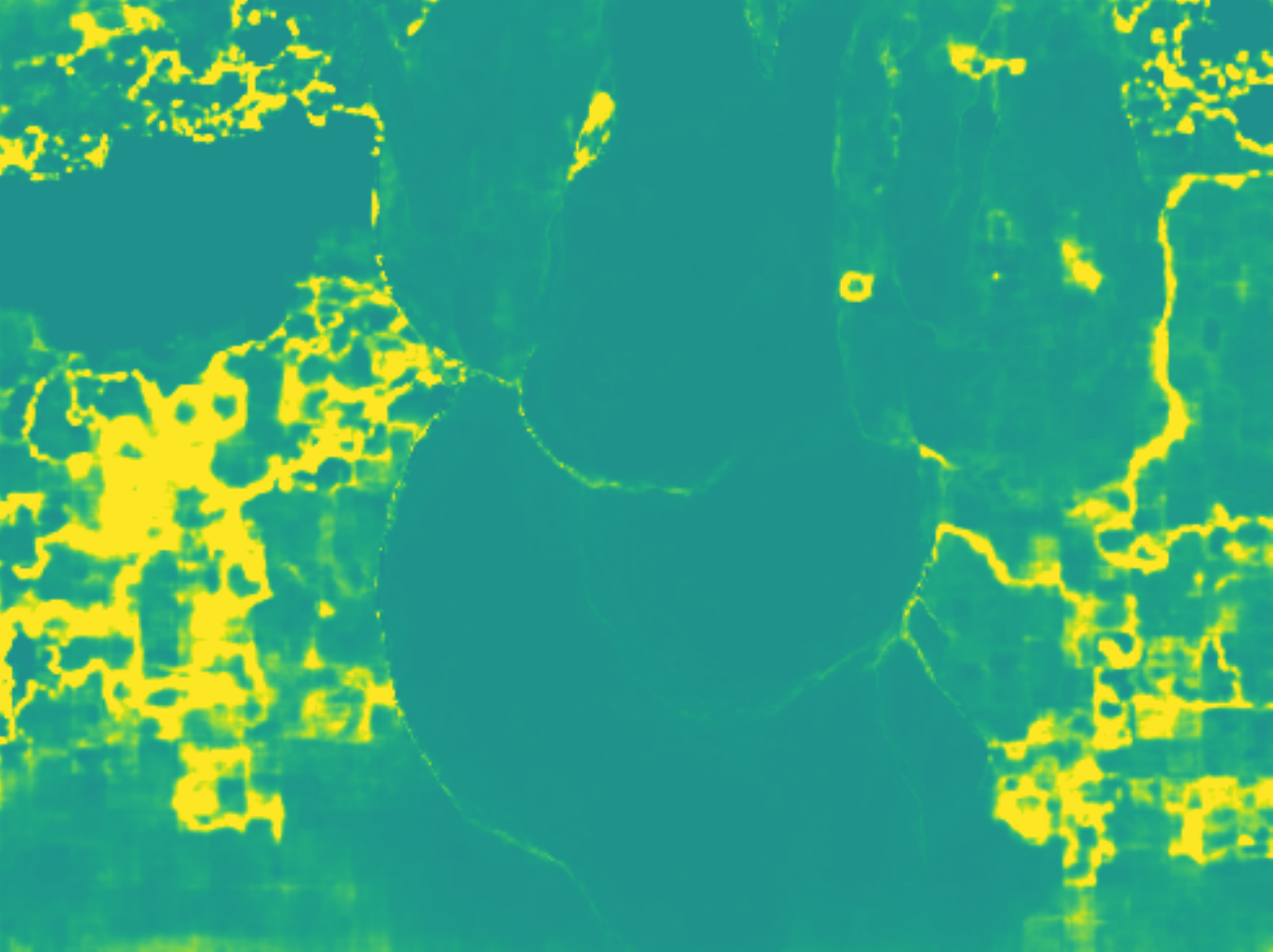}
\end{minipage}%
}%
\subfigure[Uncertainty map of stage 2]{
\begin{minipage}[t]{0.245\linewidth}
\centering
\includegraphics[height=1.1in,width=1.65in]{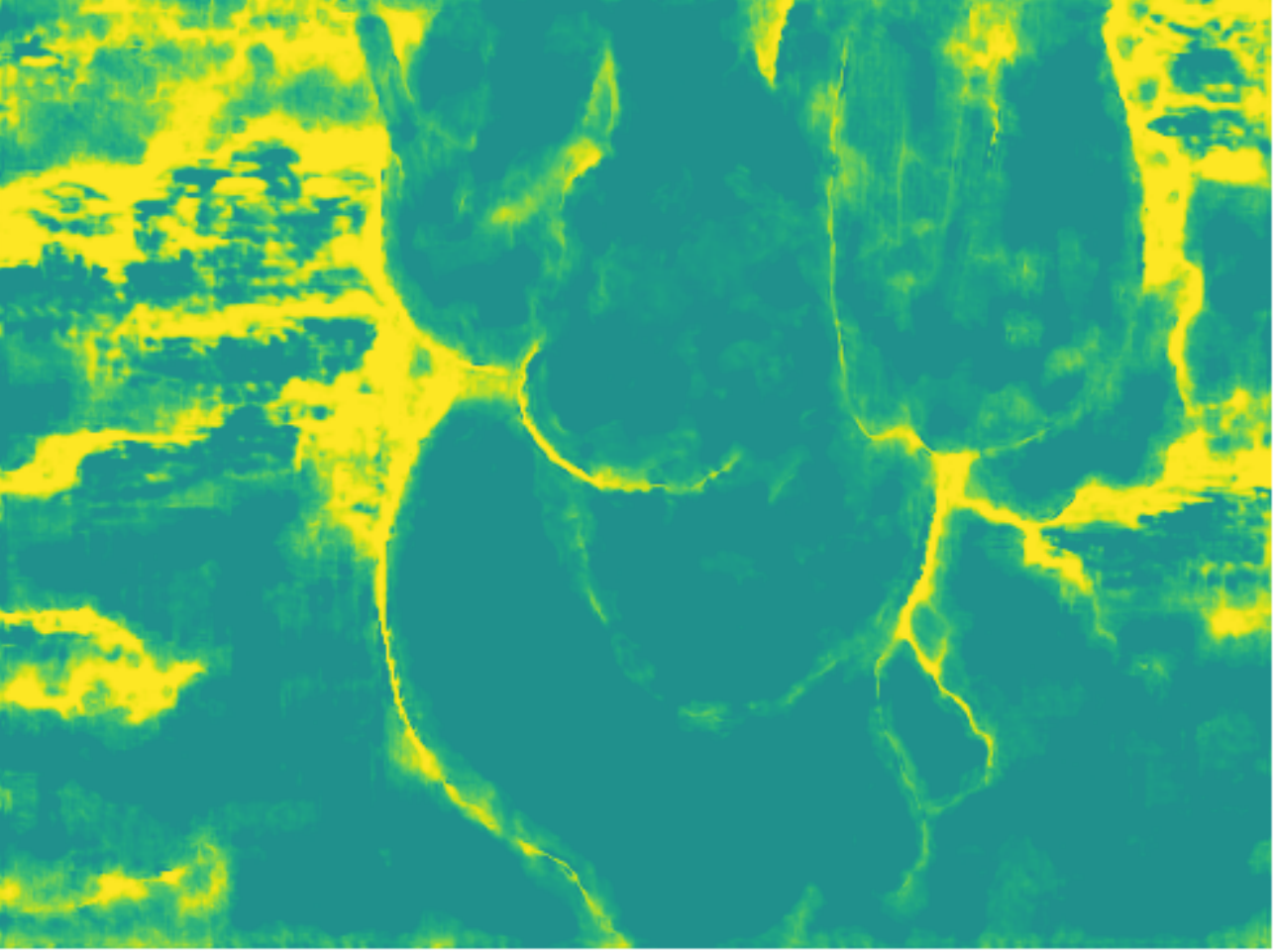}
\end{minipage}%
}%
\subfigure[Variance map of stage 2]{
\begin{minipage}[t]{0.245\linewidth}
\centering
\includegraphics[height=1.1in,width=1.65in]{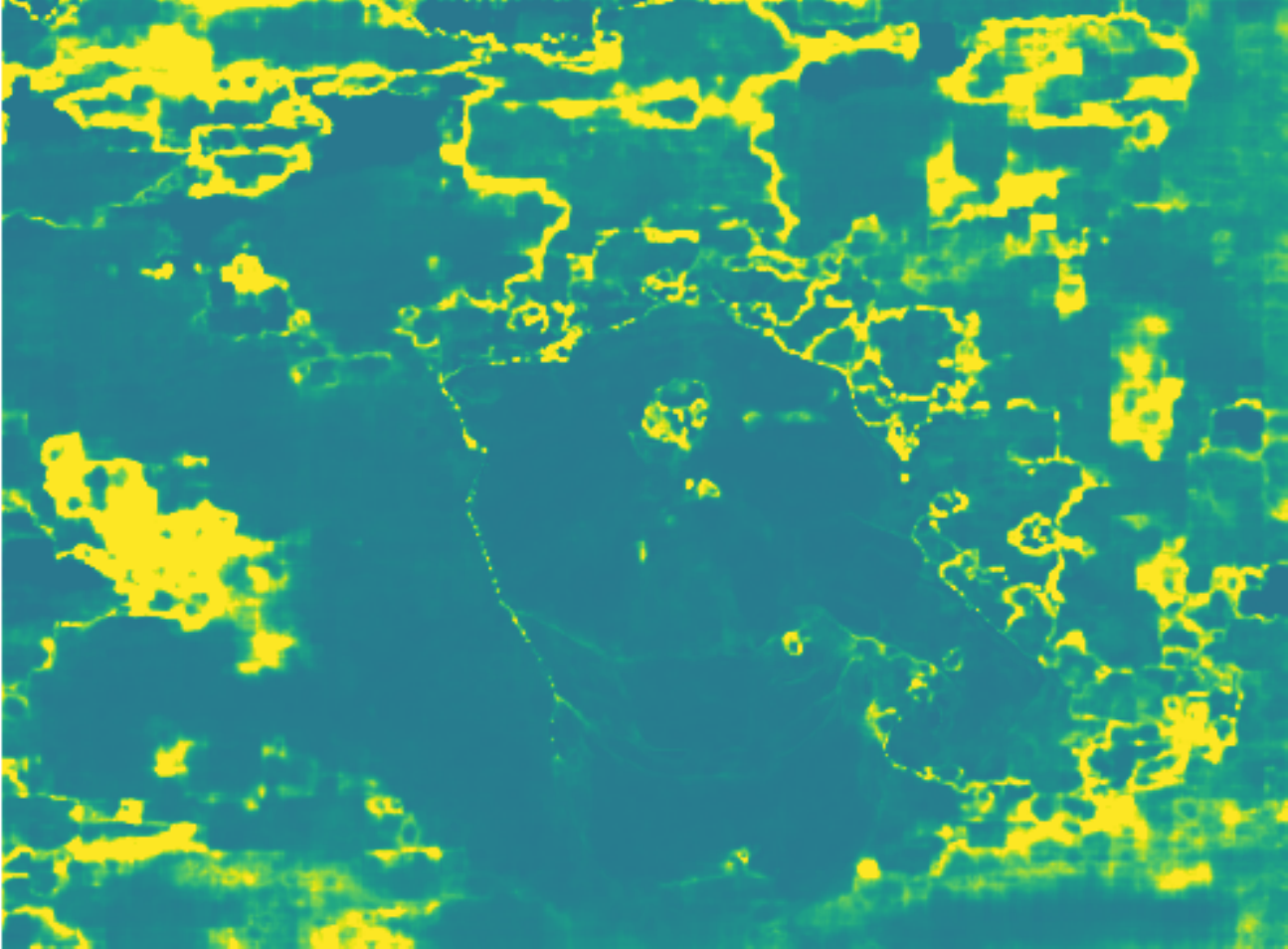}
\end{minipage}%
}%
\subfigure[Uncertainty map of stage 2]{
\begin{minipage}[t]{0.245\linewidth}
\centering
\includegraphics[height=1.1in,width=1.65in]{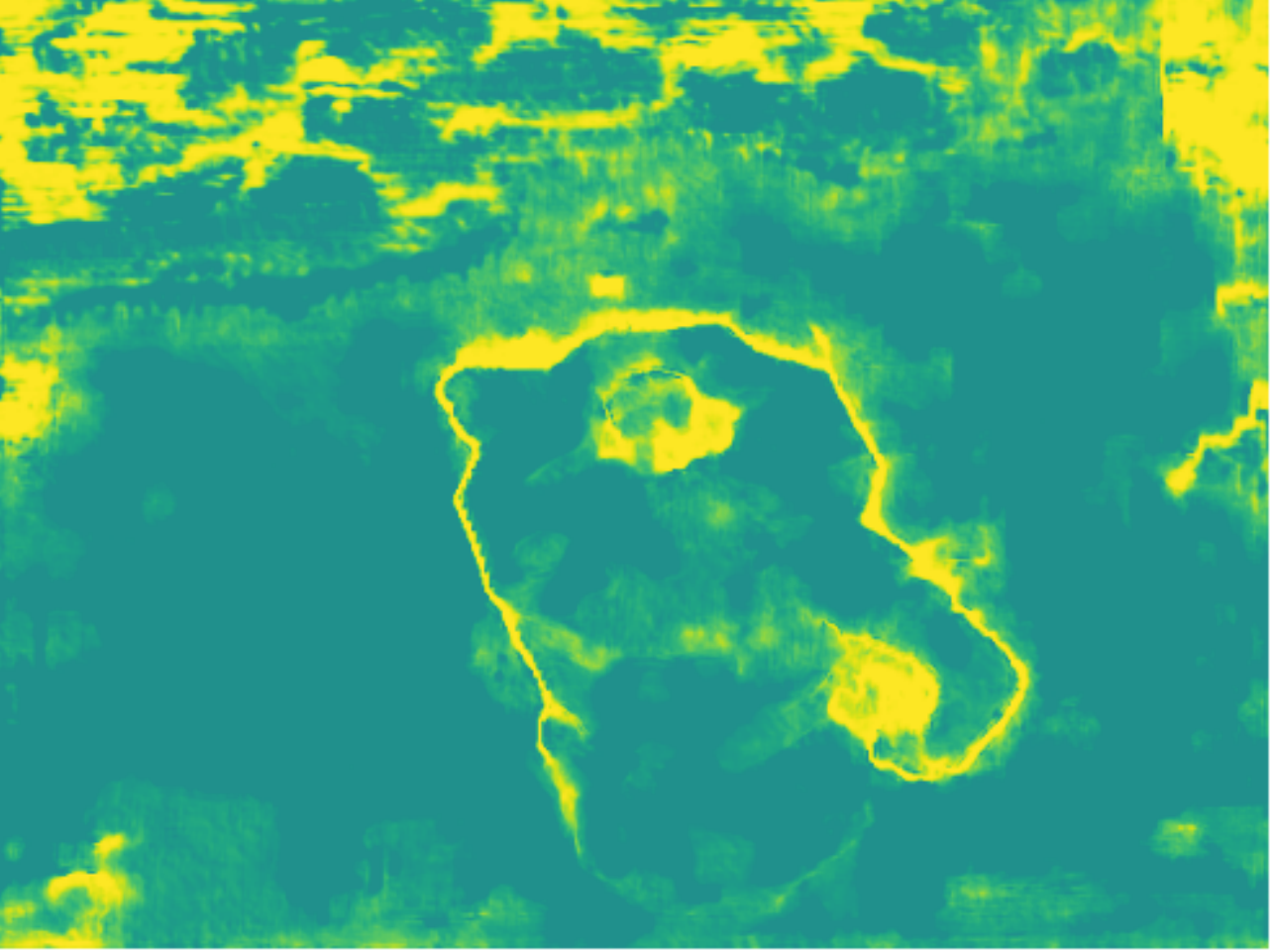}
\end{minipage}%
}%

\end{center}
\caption{Detailed depth predictions, error maps and range estimation of two specific scenes. In each scene, we show the RGB image crop, the ground depth map, the corresponding error maps obtained from UCSNet~\cite{ucs} and our DDR-Net. The variance maps obtained from UCSNet and uncertainty maps produced by DDR-Net are also provided to illustrate the effectiveness of our proposed range estimation module (REM).}
\label{Fig:ErrorVarianceMaps}
\end{figure*}

To further verify the effectiveness of our range estimation module (REM), we present the original RGB images, the detailed output depth maps, the error maps, the uncertainty maps obtained by our REM and the variance maps produced by UCSNet~\cite{ucs} of two scenes in Figure~\ref{Fig:ErrorVarianceMaps}. It is observed that our method (DDR-Net), which leverages 2D CNNs to learn dynamic depth ranges through gathering probability information of global areas, can estimate more reasonable dynamic depth ranges than UCSNet~\cite{ucs}. That is because our output uncertainty maps perform better on the boundary of objects and the texture-less region than the variance maps. Note that only pixels just close to the edges are considered with high uncertainty in the variance maps and these pixels are even not next to each other (margins are highlighted with thin and broken lines). This leads to the jagged edges of their final depth prediction (as shown in the boundary of the white cup). On the contrary, our uncertainty maps depict clear and complete boundaries of each object, which reflect the fact that our range estimation module takes the probability distribution of each pixel in the global regions into account. As a result, our final depth outputs are smoother not only in inner areas but also in the boundary of each object. Also, using variance to compute uncertainty is unable to produce satisfactory depth predictions in the texture-less regions (e.g., the white background). This is reflected by the fact that the information in the texture-less regions is insufficient to estimate depth ranges with taking only one single pixel into consideration. In contrast, our REM learns better uncertainties and produce smoother depth maps with higher accuracies on texture-less areas by utilizing probability distributions of global areas.

\begin{table}[t]
\centering\setlength\tabcolsep{8pt}
\begin{tabular}{c|c|c|c|c}  
\hline  
$\beta_1$ &$\beta_2$  &Acc. &Comp. &Overall\\ 
\hline  
3.0 &2.0 &0.348 &0.344 &0.346\\  
3.0 &1.0    &0.352 &0.338 &0.345\\
3.0 &0.5   &0.340 &0.324 &0.332\\
0.5 &0.0 & 0.344&0.326 &0.335\\
1.0 &0.0 & 0.342 & 0.326 & 0.334 \\
2.0 &0.0 &0.346 &0.322 &0.334 \\
3.0 &0.0 &0.339 &0.320 &0.329\\
\hline  
\end{tabular} 
\vspace{1em}
\caption{Detail accuracy, completeness and overall scores with different settings of weights of refined loss.} 
\label{Tab:HyperParameters}
\end{table} 

\begin{figure*}
    \centering
    \subfigure[Stage 1]{
    \begin{minipage}[t]{0.33\linewidth}
    \centering
    \includegraphics[height=1.6in,width=2.2in]{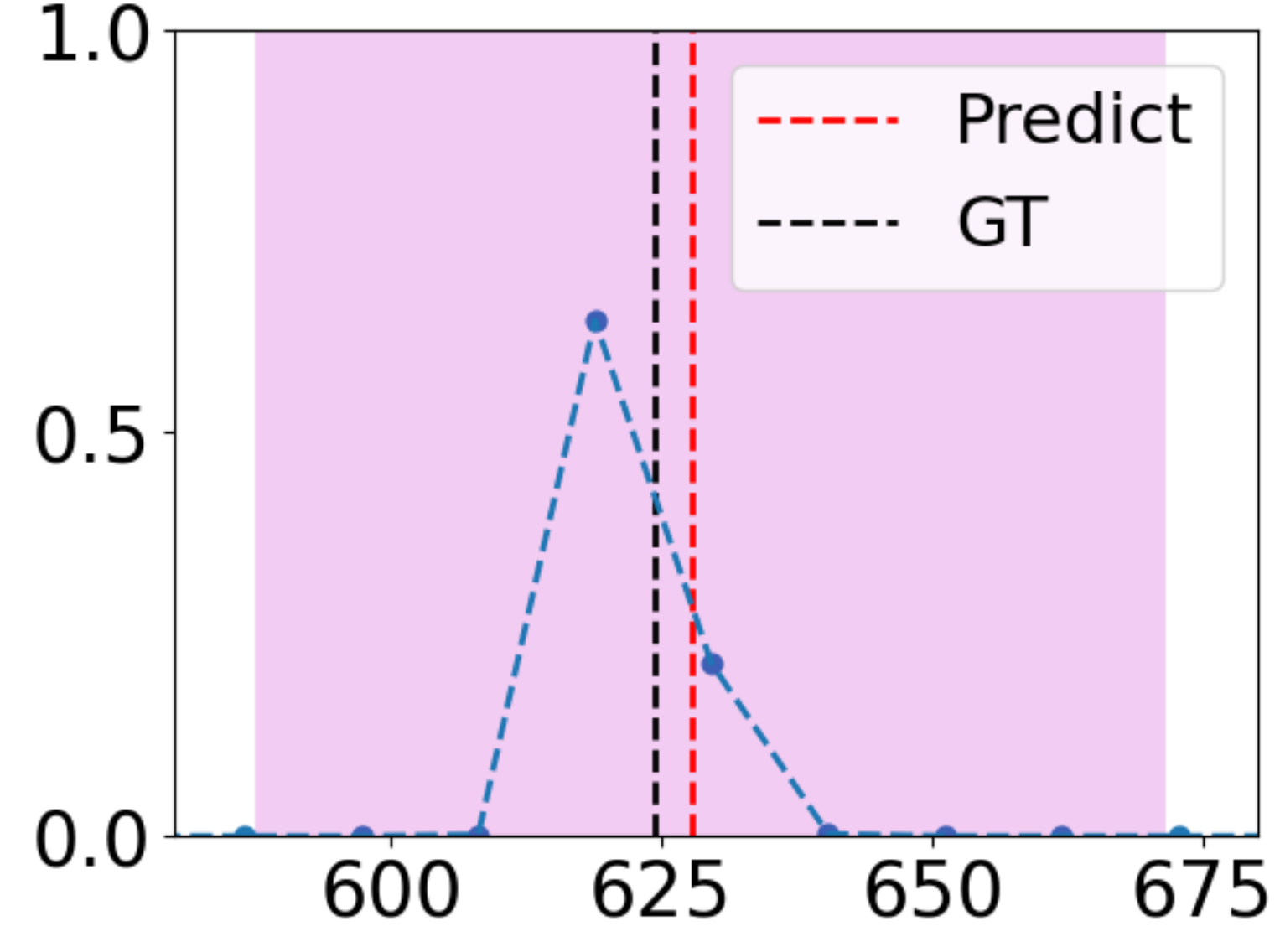}
    \end{minipage}%
    }%
    \subfigure[Stage 2]{
    \begin{minipage}[t]{0.33\linewidth}
    \centering
    \includegraphics[height=1.6in,width=2.2in]{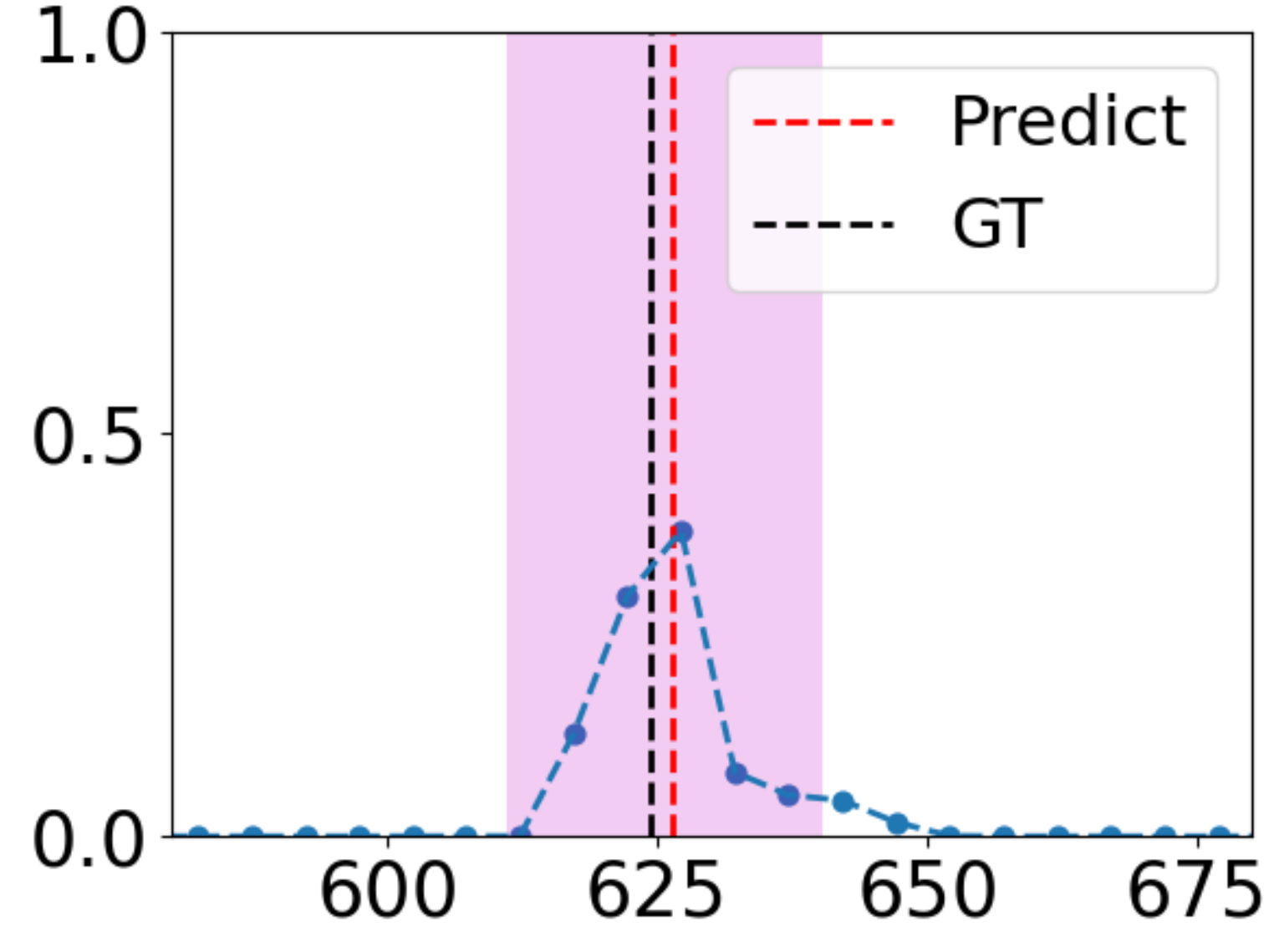}
    \end{minipage}%
    }%
    \subfigure[Stage 3]{
    \begin{minipage}[t]{0.33\linewidth}
    \centering
    \includegraphics[height=1.6in,width=2.2in]{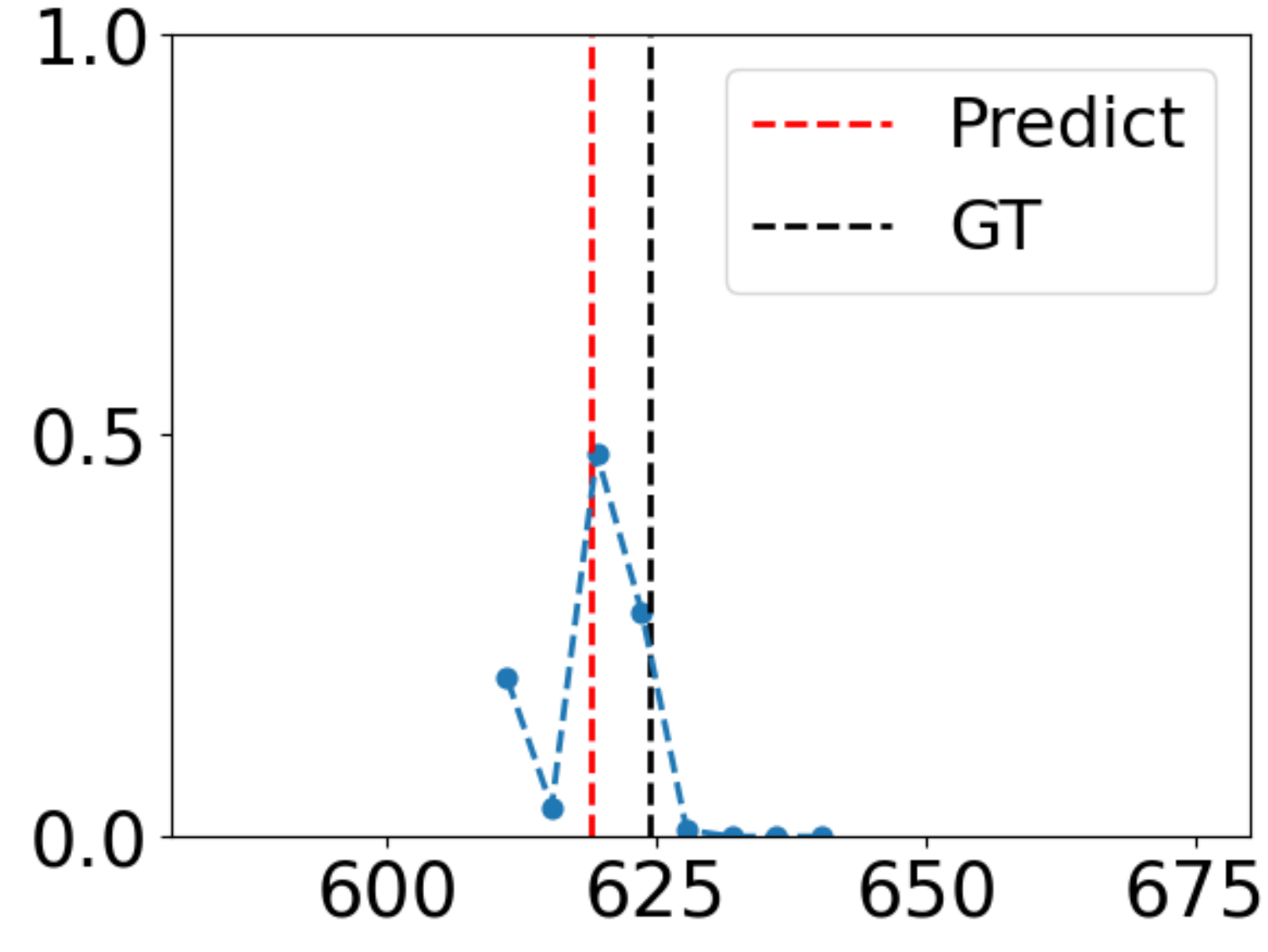}
    \end{minipage}%
    }%
    
    \caption{Illustration of a failure case with our proposed novel loss strategy in the stage 2. The clamped probability distribution of the stage 2 is different from the probability distribution in the stage 3.  }
    \label{Fig:FailureCaseProbDistribution}
\end{figure*}

\subsection{Additional Experiments on Loss Strategy}
\label{Sec:LossStrategyAdditionalExperiments}

In this section, we discuss the details of our proposed novel loss strategy. As mentioned before, our proposed novel loss strategy can leverage the estimated depth range hypotheses to output the refined depth maps in advance. In our experiments, the most important hyperparameter for loss strategy is the weight of the refined loss $\beta_k$. In order to find the best hyperparameter settings, we conducted a series of experiments with different $\beta_k$ settings. As shown in Table~\ref{Tab:HyperParameters}, the best setting of $\beta_k$ was $\beta_1$= 3 and $\beta_2$= 0 in our experiments. We argue that the novel loss strategy in the 1st stage is quite enough for accurate depth range estimation and there might be some bad cases failing to estimate appropriate refined depth maps in the stage 2. Our assumption for loss strategy is that the probability distributions in three stages are similar even the depth sampling rates progressively increase. But we discovered some failure cases between the stage 2 and the stage 3. As shown in Figure~\ref{Fig:FailureCaseProbDistribution}, comparing with initial depth range hypotheses in the stage 1, the new dynamic depth range hypotheses in the stage 2 are shortened appropriately that the clamped probability distribution of the stage 1 is basically the same as the distribution in the stage 2. However, the depth range hypotheses in the stage 3 might be too short that the probability distribution which is clamped by new estimated range hypotheses is different from the distribution of the stage 3. If we utilize our novel loss strategy to obtain the refined depth map in advance in this situation, we may get wrong refined depth prediction. Therefore, the setting of $\beta_1$= 3 and $\beta_2$ =0 is good choice for achieving good performance.

\subsection{Point Clouds Reconstruction}
\label{Sec:PointCloudReconstruction}

\begin{figure*}
    \centering
    \includegraphics[width=1\textwidth]{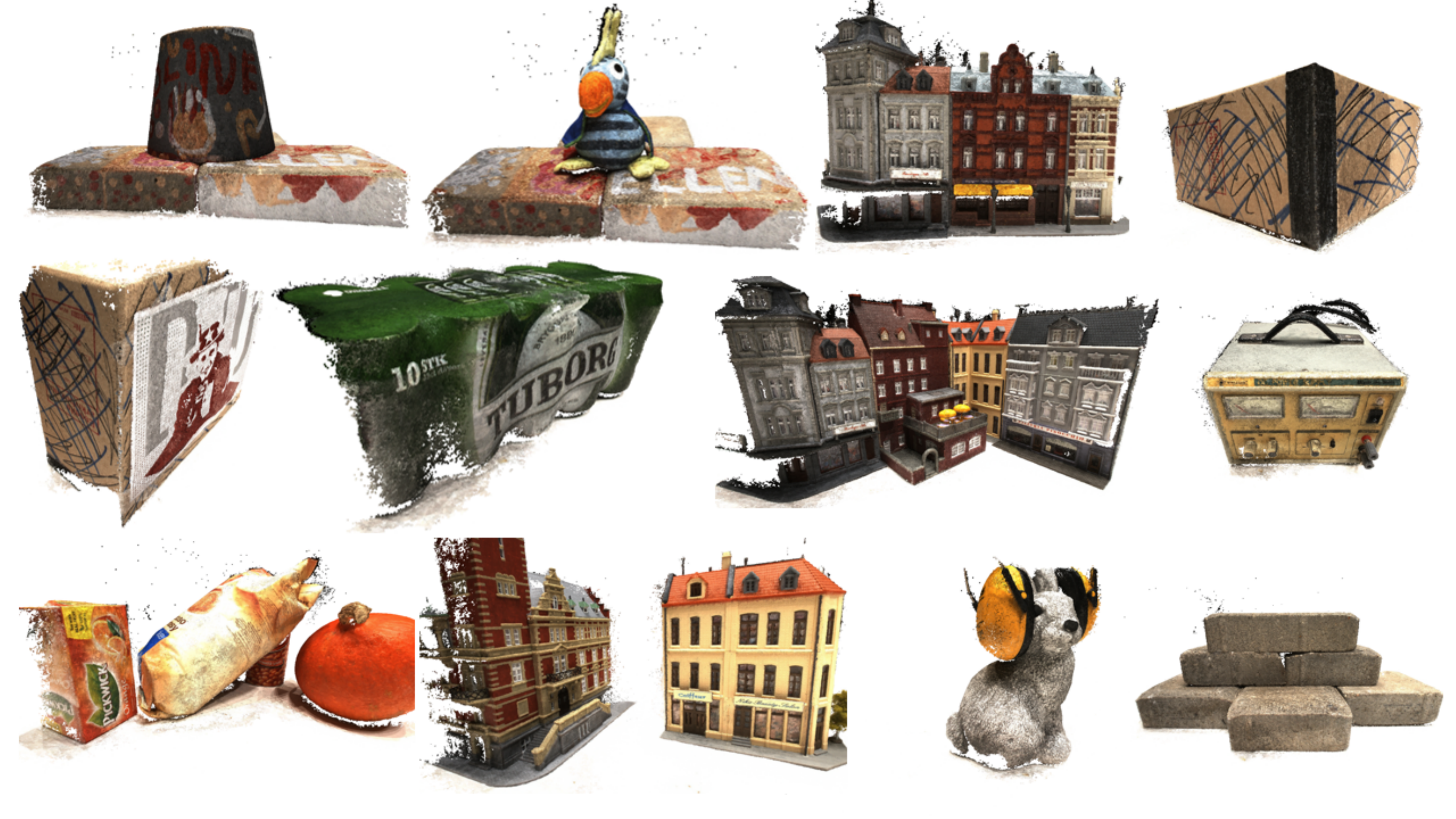}
    \includegraphics[width=1\textwidth]{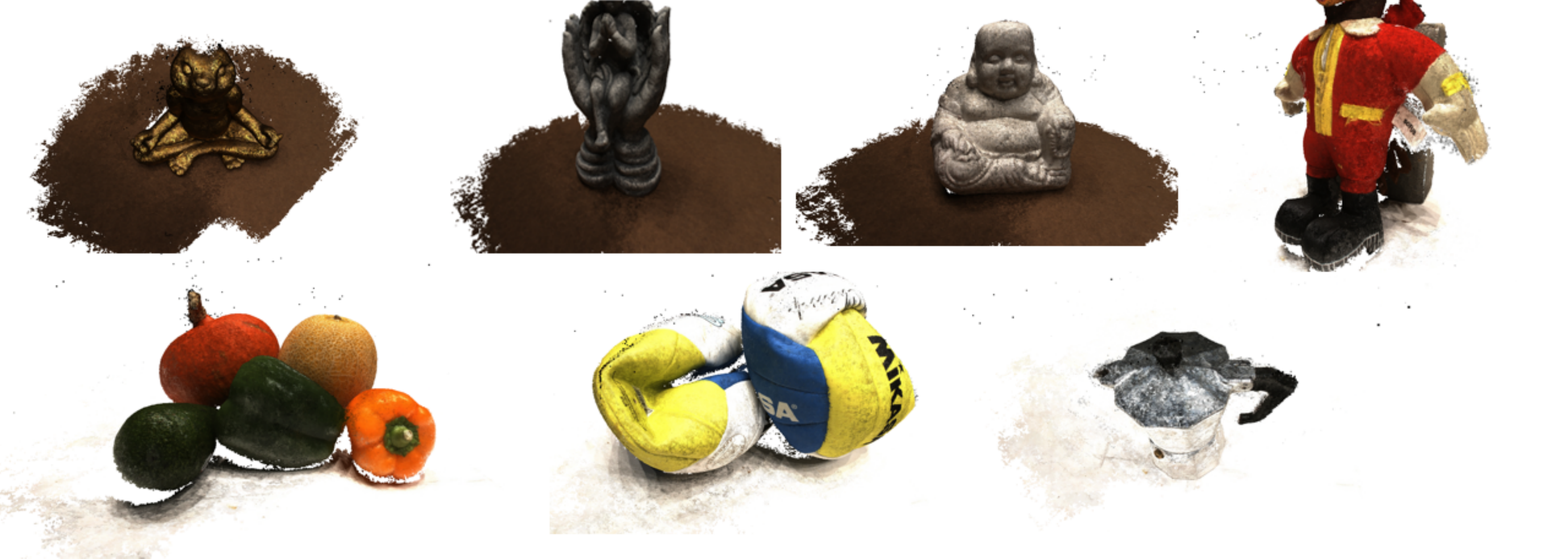}
    \caption{Details of some reconstructed point clouds on the DTU dataset.}
    \label{Fig:DTUPointClouds}
\end{figure*}

In this section, we present additional point clouds reconstruction results to those presented in main paper, as shown in Figure~\ref{Fig:DTUPointClouds}. From these results, we can observe that our proposed DDR-Net can reconstruct various scenes well.

\subsection{Future Work}
\label{Sec:FutureWork}

Our proposed DDR-Net dynamically infers the depth range hypotheses with a large reception field for depth prediction. We have evaluated the effectiveness of estimating dynamic depth range hypotheses. Although DDR-Net is able to incorporate information from neighborhood pixels to estimate better depth range hypotheses, some limitations still exist. During sampling with new depth ranges, the depth sampling intervals and the number of depth planes are identical which means the whole depth range will be divided equally into static number of samples. 

We argue obtaining adaptive depth sampling intervals and depth planes is essential for more accurate depth values prediction. The depth range areas with high probabilities should use small depth sampling intervals for more depth samples. And the samples of areas with low probabilities should be reduced which is beneficial for accurate depth estimation. At the same time, the adaptive depth planes are able to reduce unnecessary depth samples to drop GPU memory usage and runtime significantly. These ideas probably need a novel cost volume regularization strategy to learn dynamic depth plane settings for each pixel. 
\end{document}